\pdfoutput=1
\documentclass[10pt, a4paper, copyright, google]{google}

\uselogo{}
\usepackage{multirow}
\usepackage{makecell}
\usepackage{tabularx}
\usepackage{array}
\usepackage{adjustbox}
\usepackage{placeins}
\usepackage{wrapfig}
\usepackage{listings}
\usepackage{ulem}
\usepackage[most]{tcolorbox}
\usepackage[numbers,sort&compress]{natbib}
\usepackage[capitalize]{cleveref}
\crefname{section}{Sec.}{Secs.}
\Crefname{section}{Section}{Sections}
\Crefname{table}{Table}{Tables}
\crefname{table}{Tab.}{Tabs.}

\definecolor{linkblue}{rgb}{0.21,0.49,0.74}
\hypersetup{colorlinks=true,allcolors=linkblue}

\makeatletter
\DeclareRobustCommand\onedot{\futurelet\@let@token\@onedot}
\def\@onedot{\ifx\@let@token.\else.\null\fi\xspace}
\RequirePackage{xspace}

 \def\vs{\emph{vs}\onedot}

\makeatother

\definecolor{oursbg}{HTML}{EEF4FF}
\definecolor{ourshead}{HTML}{3C5488}
\definecolor{findingbg}{HTML}{F8FAFC}
\definecolor{findingframe}{HTML}{3C5488}
\definecolor{cambrianbg}{HTML}{EDF3F5}
\definecolor{cambrianframe}{HTML}{2F5561}
\definecolor{upgreen}{HTML}{1E7A68}
\definecolor{downred}{HTML}{D94F38}
\definecolor{secondblue}{HTML}{3C5488}

\newcommand{\best}[1]{\textbf{\boldmath #1}}

\newcommand{\questionbox}[1]{%
  \begin{tcolorbox}[
    enhanced,
    colback=white,
    colframe=cambrianframe!60,
    boxrule=0.5pt,
    arc=2pt,
    outer arc=2pt,
    left=7pt,
    right=7pt,
    top=3pt,
    bottom=3pt,
    before skip=4pt,
    after skip=4pt
  ]
  \hyphenpenalty=10000 \noindent\textbf{Question:}\hspace{0.35em}\emph{#1}
  \end{tcolorbox}%
}

\newcommand{\takeaway}[3]{%
  \begin{tcolorbox}[
    enhanced,
    colback=cambrianbg,
    colframe=cambrianframe,
    boxrule=0pt,
    arc=2pt,
    outer arc=2pt,
    left=7pt,
    right=7pt,
    top=3pt,
    bottom=3pt,
    borderline west={2.5pt}{0pt}{cambrianframe},
    before skip=4pt,
    after skip=4pt
  ]
  \noindent\textbf{Takeaway #1 (Figure~\ref{#2}):}\hspace{0.35em}#3
  \end{tcolorbox}%
}

\newcommand{\lead}[1]{\noindent\textbf{#1}}
\newcommand{\up}[1]{\textcolor{upgreen}{+#1}}
\newcommand{\dn}[1]{\textcolor{downred}{$-$#1}}

\lstdefinestyle{xhpython}{
  language=Python,
  basicstyle=\ttfamily\scriptsize\linespread{1.05}\selectfont,
  keywordstyle=\color{ourshead}\bfseries,
  commentstyle=\color{gray!85!black}\itshape,
  stringstyle=\color{upgreen!85!black},
  showstringspaces=false,
  breaklines=true,
  columns=flexible,
  keepspaces=true,
  aboveskip=2pt,
  belowskip=2pt,
}

\correspondingauthor{ ($\dagger$): mai.145@osu.edu, ruixifan@google.com}

\title{Toward Agentic Robotics:\\A Large-Scale Study of Harness Design for Robot-Use Agents}

\author[1,2]{Yuxiang Lai}
\author[3]{Jike Zhong}
\author[1]{Yu-Jhe Li}
\author[1]{Lan Wang}
\author[1,6]{Chengzhi Liu}
\author[1]{Youbao Tang}
\author[4]{Ming Li}
\author[8]{Xi Xiao}
\author[2,5]{Xiaofeng Yang}
\author[2]{Liang Zhao}
\author[7$\dagger$]{Zheda Mai}
\author[1$\dagger$]{Ruixi Fan}

\affil[1]{\thepa{}{}}
\affil[2]{Emory University}
\affil[3]{University of Southern California}
\affil[4]{University of Florida}
\affil[5]{University of Chicago}
\affil[6]{University of California, Santa Barbara}
\affil[7]{Ohio State University}
\affil[8]{University of Alabama at Birmingham}

\begin{abstract}
Recent vision-language models (VLMs) can act as robot-use agents that understand diverse scenes and control robots through policy code and tools. However, they predict 3D gripper poses poorly even with extended reasoning, struggling with precise manipulation. Vision-language-action (VLA) and world-action (WAM) models are the opposite: they execute fast, smooth control on trained tasks, yet fail under domain shifts or when tasks require long-horizon memory and reasoning. Agentic robots therefore combine VLM reasoning and perception with precise VLA/WAM control through a \emph{policy harness}. Yet harness design spans thousands of configurations, shifting success by tens of points with the same models. Moreover, testing even one design takes thousands of rollouts at high API and GPU cost, so prior works often commit to a single design chosen from few alternatives. To guide harness design, we evaluate over 470,000 rollouts across eight simulation benchmarks, six VLMs, and 15 VLA/WAM checkpoints. Seven controlled studies show effective harnesses must adapt to both VLM and VLA/WAM capabilities. Our findings specify what structure the VLM's policy code needs, where VLA manipulation needs help from tools and policy code, and when to involve the VLM at test time. The reference recipe \textbf{X-Harness} shows that these findings compose without weight updates, lifting base policies up to $+42.8$ points across memory, reasoning, and domain shifts.
\end{abstract}

\makeatletter
\renewcommand{\absfont}{\bfseries\linespread{1.12}\fontsize{10}{12}\selectfont}
\renewcommand{\maketitle}{%
  \bgroup\setlength{\parindent}{0pt}%
  \begin{adjustwidth}{0pt}{24pt}
    \begin{flushleft}
      {\raggedright \titlefont \@title\par}%
      \vskip 5pt
      {\raggedright \@author\par}%
      \vskip 6pt
    \end{flushleft}
  \end{adjustwidth}
  \egroup
  \noindent\begin{minipage}{\textwidth}
    \captionsetup{type=figure}%
    \centering
    \includegraphics[width=0.86\textwidth]{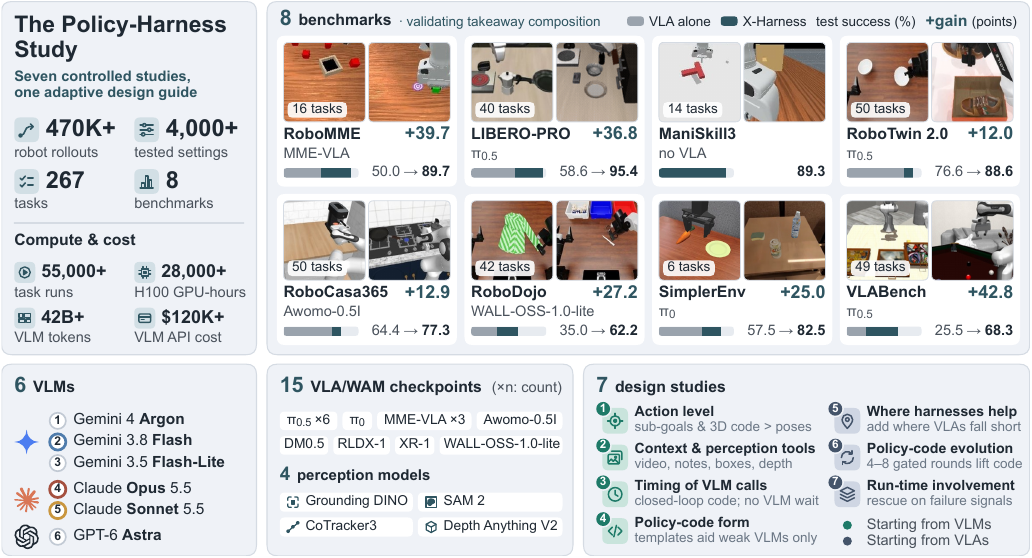}
    \vspace{-5pt}
    \caption{\textbf{Overview of the policy-harness study.} We study the policy harness, which connects vision-language models (VLMs) to robots, on eight simulation benchmarks across seven design aspects. X-Harness is the weight-update-free reference recipe testing whether the seven takeaways (\S\ref{sec:vlm}--\ref{sec:vla}) compose. Tiles report test success (\%) without and with X-Harness across our single-VLM Offline and multi-VLM Semi-online recipes (\S\ref{sec:recipe}).}
    \label{fig:overview}
  \end{minipage}\par
  \vskip 7pt
  \noindent\begin{minipage}{\textwidth}
    {\abscontent}\par
    \vskip 6pt
    \noindent\raisebox{-0.20em}{\includegraphics[height=1.05em]{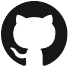}}%
    \hspace{0.45em}\textbf{Code \& Data:}\hspace{0.4em}\href{https://github.com/google-research/x-as-policies}{\texttt{https://github.com/google-research/x-as-policies}}%
  \end{minipage}\par
  \thispagestyle{firststyle}%
  \clearpage
}
\makeatother

\begin{document}
\maketitle
\fancyhead[C]{\footerfont Toward Agentic Robotics: A Large-Scale Study of Harness Design for Robot-Use Agents}

\section{Introduction}
\label{sec:intro}

Recent vision-language models (VLMs) have emerged as \emph{robot-use agents}~\cite{isola2026robotuse} that understand diverse scenes, recall past events, and---building on programmatic robot control~\cite{liang2023codeaspolicies, huang2023voxposer}---control robots through policy code and tools~\cite{capx2026, hexashell2026}. However, they struggle with precise manipulation: they predict 3D gripper poses poorly even with extended reasoning and motion planners (\S\ref{sec:action}), and calling them in the loop stalls the robot. Vision-language-action (VLA)~\cite{black2024pi0, intelligence2025pi05, kim2024openvla} and world-action (WAM)~\cite{awomo2026} models are the opposite: they execute fast, smooth control on trained tasks, yet fail under domain shifts or when tasks require long-horizon memory and reasoning. For instance, $\pi_{0.5}$~\cite{intelligence2025pi05} exceeds $95\%$ on standard LIBERO~\cite{liu2023libero} but drops to about a quarter once LIBERO-PRO~\cite{liberopro2025} shifts object positions.

\emph{Agentic robots}~\cite{qin2026towardsagenticrobots} therefore combine VLM reasoning and perception with precise VLA or WAM control, alongside motion primitives (functions that move the gripper). The interface that connects them is the \emph{policy harness} (\S\ref{sec:harness}): it specifies what context the VLM receives, whether it acts in-loop or writes \emph{policy code} (free-form Python or symbolic templates), and when it is called. Yet harness design spans thousands of configurations, shifting success by tens of points with the same VLM and action policy \mbox{(\S\ref{sec:vlm}--\ref{sec:vla})}. Moreover, testing even one design takes thousands of rollouts at high API and GPU cost, so prior works often commit to a single harness~\cite{liang2023codeaspolicies, huang2024rekep, hexashell2026, harnessvla2026} or restrict their ablations to a single regime and benchmark~\cite{wang2026justrua, hu2026orchestrating}. As a result, how to build effective harnesses for agentic robots remains unclear.

To answer this question, we evaluate over 4,000 settings and 470,000 rollouts across eight simulation benchmarks (267 tasks), six VLMs, and 15 VLA/WAM checkpoints (Figure~\ref{fig:overview}). Controlled simulation enables one-factor-at-a-time ablations on matched seeds, complemented by supplementary physical validation on a bimanual robot (Appendix~\ref{supp:real_robot}). Starting from VLMs (\S\ref{sec:vlm}), we study (1)~\emph{action level}: 3D poses versus VLA \emph{sub-goals}, 2D visual prompts, and 3D target code; (2)~\emph{context and perception}: video history, task notes, pre-coding detections, and depth quality; (3)~\emph{call timing}: in-loop VLM calls versus closed-loop policy code; and (4)~\emph{code structure}: free-form Python versus symbolic templates. Starting from VLAs (\S\ref{sec:vla}), we study (5)~\emph{where harnesses help} across tasks and base policies; (6)~\emph{pre-test evolution}: revising policy code or a shared tool bank from training failures under regression checks; and (7)~\emph{test-time VLM involvement}: \emph{Offline} does not call the VLM, \emph{Semi-online} calls the VLM on failure signals, and \emph{Full-online} calls the VLM per sub-goal.

Across these seven studies, we find that effective harnesses must adapt to both VLM coding strength and VLA/WAM capability rather than following one fixed design. To provide open baselines and practical guidelines for researchers stepping into agentic robotics, our contributions are:
\begin{itemize}[leftmargin=*, topsep=2pt, itemsep=1pt]
    \item \textbf{Large-scale benchmark study and open baselines} (\S\ref{sec:setup}), evaluating over 4,000 configurations across eight benchmarks under one protocol, releasing our evaluation suite, policy code, and 470,000+ rollout logs.
    \item \textbf{Seven empirical takeaways as a design guideline} (\S\ref{sec:vlm}--\ref{sec:vla}; Table~\ref{tab:recipe}), specifying perception grounding, code structure, VLA-vs.-primitive routing, code evolution, and test-time VLM scheduling.
    \item \textbf{X-Harness, a reference recipe} (\S\ref{sec:recipe}) showing that these guidelines compose without weight updates to lift base policies under domain shifts, on memory tasks, and in precise manipulation.
\end{itemize}

\section{Related Work}
\label{sec:related}

\lead{VLMs as robot-use agents.} Programmatic robot control began with Code as Policies~\cite{liang2023codeaspolicies} and ProgPrompt~\cite{singh2023progprompt}, which synthesize Python over perception APIs and motion primitives, while VoxPoser~\cite{huang2023voxposer} and ReKep~\cite{huang2024rekep} formulate manipulation through 3D value maps or relational keypoint constraints. Recent \emph{robot-use agents}~\cite{isola2026robotuse} generalize these ideas through multi-turn visual feedback loops using either direct perception and motion tool calls~\cite{gpt6astra2026, li2026roboharness} or code execution environments~\cite{capx2026, hexashell2026, openrua2026}. Under matched perception tools, primitives, and test episodes, we systematically compare open-loop plans, free-form code, five symbolic templates, and visual-prompted sub-goals (\S\ref{sec:vlm}; Table~\ref{supp:tab:prior}).

\lead{VLAs and hierarchical VLM--VLA systems.} Generalist visuomotor policies---including VLAs~\cite{brohan2023rt2, kim2024openvla, black2024pi0, intelligence2025pi05, bjorck2025gr00t} and world-action models (WAMs)~\cite{awomo2026}---excel at reactive low-level control on in-distribution demonstrations but degrade under distribution shifts or long-horizon memory demands. Hierarchical systems therefore layer VLM planners over low-level VLAs for sub-goal decomposition, episodic memory, or failure recovery, either by fine-tuning the hierarchy~\cite{shi2025hirobot, memer2026} or by steering frozen VLAs without weight updates~\cite{harnessvla2026, astraguide2026}. Rather than proposing a single fixed orchestrator, we characterize where harnesses help across 15 frozen VLA/WAM checkpoints and how to balance runtime latency and token overhead against recovery accuracy (\S\ref{sec:vla}).

\lead{Empirical harness studies and policy-code evolution.} Prior empirical studies focus on a single regime---ablating VLM-only robot-use agents on one benchmark~\cite{wang2026justrua} or comparing VLM sub-goal planners over VLAs~\cite{hu2026orchestrating}---while recent self-improving frameworks refine policy code or reusable skill libraries from rollout experience~\cite{embodiedevo2026, embodiedrsi2026, aspire2026, rpg2026}. We unify both perspectives across eight benchmarks (267 tasks), studying how pre-test code and tool evolution, regression checks, and failure-triggered recovery compose in \S\ref{sec:recipe}.

\section{Study Design}
\label{sec:setup}

\subsection{The Policy Harness}
\label{sec:harness}

\begin{wrapfigure}{r}{0.52\textwidth}
    \vspace{-10pt}
    \centering
    \includegraphics[width=0.51\textwidth]{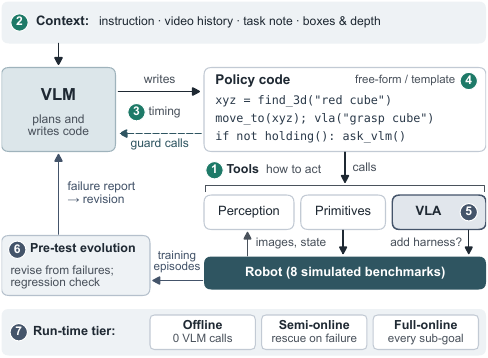}
    \caption{\textbf{Overview of the policy harness and our seven design studies.} Badges mark each studied aspect, starting from VLMs (green, \S\ref{sec:vlm}) and VLAs (gray, \S\ref{sec:vla}); the dashed arrow is a \emph{guard} that calls the VLM on failure.}
    \label{fig:harness}
    \vspace{-8pt}
\end{wrapfigure}

The \emph{policy harness} comprises five components (Figure~\ref{fig:harness}).

\lead{Context.} The VLM receives the instruction, camera frames or video, tool outputs, and an optional task note specifying conventions (e.g., which side counts as ``right'').

\lead{Tools.} Policy code acts by calling three tool types. Perception tools detect, segment, and track objects and return depth. Motion primitives open or close the gripper and move it to a target pose (position and rotation) along a planned path. The VLA also serves as a tool, taking a language sub-goal (``pick up the red cube'') and a step limit.

\lead{Policy code.} By default, the VLM writes policy code on training tasks and reuses it in all test episodes. The code is free-form Python or fills a fixed template such as behavior trees or PDDL. It is closed-loop if it reads live images and robot state, and open-loop if it runs a fixed sequence. Optional guards call the VLM when it stalls or fails, returning corrective actions or replacing it in a takeover.

\lead{Pre-test evolution.} Before testing, code evolves on training episodes (ten by default). In each round, the VLM revises it from failure reports (with optional training diagnostics unused at test time), and a selection gate keeps a revision only if training success rises. One-shot evolution revises code once from one episode.

\lead{Run-time tier.} Each test-time VLM call incurs latency and API tokens. \emph{Offline} runs the frozen policy code with zero VLM calls. \emph{Semi-online} calls the VLM only on failure signals, for a one-shot takeover or multi-turn rescue. \emph{Full-online} calls the VLM for every sub-goal.

\subsection{Benchmarks, Models, and Protocol}
\label{sec:protocol}

\lead{Benchmarks.} We evaluate eight benchmarks spanning 267 tasks (Figure~\ref{fig:overview}; Table~\ref{supp:tab:bench}): RoboMME~\cite{robomme2025} (memory), LIBERO~\cite{liu2023libero}, ManiSkill3~\cite{tao2024maniskill3} (primitives only), RoboTwin~2.0~\cite{mu2025robotwin2}, RoboCasa365~\cite{nasiriany2026robocasa365}, RoboDojo~\cite{robodojo2025}, SimplerEnv~\cite{li2024simplerenv}, and VLABench~\cite{zhang2024vlabench}. All 40 LIBERO tasks are tested under standard and four LIBERO-PRO~\cite{liberopro2025} shifts (object, language, \emph{position}, \emph{task}); \emph{LIBERO(-PRO)} pools the latter two with standard LIBERO.

\lead{Models.} Three frozen cloud VLMs, from weaker to stronger, cover most studies: \emph{Flash} (Gemini~3.8 Flash), \emph{Sonnet} (Claude \mbox{Sonnet~5.5}), and \emph{Opus} (Claude Opus~5.5). For cross-family coverage we add Gemini~4 Argon, GPT-6 Astra (\S\ref{sec:improve}, \S\ref{sec:recipe}), and Gemini~3.5 Flash-Lite, supplemented by two quantized local Qwen3.6 models for zero-API edge validation (\S\ref{sec:edge_gen}; Appendix~\ref{supp:models}). We also evaluate 15 frozen base policies (``VLAs'' below; Table~\ref{supp:tab:vla}): six $\pi_{0.5}$ fine-tunes, three MME-VLA variants~\cite{robomme2025}, $\pi_0$~\cite{black2024pi0}, \mbox{RLDX-1}~\cite{rldx2026}, Xiaomi-Robotics-1 (\mbox{XR-1})~\cite{xiaomirobotics2026xr1}, DM0.5~\cite{dm05_2026}, WALL-OSS-1.0-lite~\cite{walloss2026, walloss1lite2026}, and the Awomo-0.5I WAM~\cite{awomo2026}.

\lead{Sensing.} At test time, no benchmark uses ground-truth simulator object poses or masks: RGB boxes, masks, and tracks come from Grounding~DINO~\cite{liu2023groundingdino}, SAM~2~\cite{ravi2024sam2}, and CoTracker3~\cite{karaev2024cotracker3}, with robot-link depth masked out. On seven benchmarks, the VLA takes RGB while 3D tools read camera depth (with Depth Anything~V2~\cite{yang2024depthanythingv2} filling depth holes on RoboCasa365 in our scaling runs). RoboDojo is \emph{Pure-RGB} (no depth sensor): single-VLM studies use 2D detections only, whereas our scaling runs (\S\ref{sec:runtime}, \S\ref{sec:recipe}) estimate metric depth from RGB via Depth Anything~V2 and known robot-and-table geometry (Appendix~\ref{supp:protocol}).

\lead{Protocol.} Each \emph{cell} (setting) pairs one task with one VLM, VLA, and harness design across at least 20 test episodes, spanning over 470,000 rollouts in total (489,380 logged across 4,025 cells; Table~\ref{supp:tab:scale}, Figure~\ref{supp:fig:settings}). We report test success (\%) and point changes; \emph{VLA alone} runs the VLA on the same episodes with only the instruction. Key comparisons use a two-level paired bootstrap with Holm correction ($\alpha=0.05$; denoted \emph{significant}). Evolution, candidate selection, and rescue tuning use only training seeds (1000+; LIBERO states 30--39; RoboDojo $\geq 100$) before blind evaluation on held-out test seeds (0--19; LIBERO 10--29; RoboDojo~0).

\Needspace*{10\baselineskip}
\section{Starting from VLMs}
\label{sec:vlm}

\subsection{Action Level of the VLM}
\label{sec:action}

\questionbox{Can VLMs control the gripper directly, or should they delegate low-level motion to VLAs?}

\begin{figure}[!htb]
    \centering
    \includegraphics[width=\textwidth]{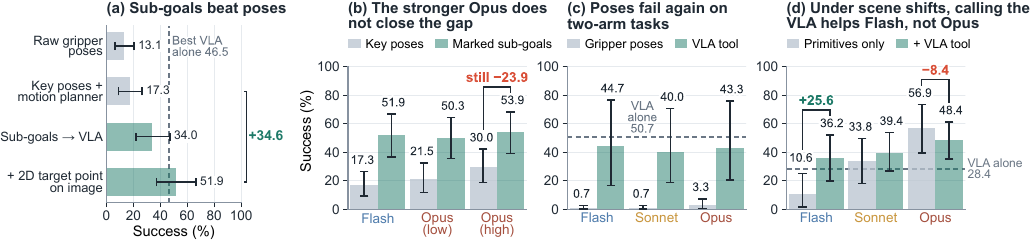}
    \caption{\textbf{Study 1 / Takeaway 1: Direct VLM pose predictions fail; VLMs should delegate via VLA sub-goals or query 3D tools in code.} \textbf{(a)} On RoboMME (Flash), language sub-goals beat gripper or key poses, and marking a 2D target point lifts the VLA past its standalone baseline (dashed). \textbf{(b)} Scaling to Opus with high reasoning effort does not close the key-pose gap. \textbf{(c)} On two-arm RoboDojo, direct gripper poses fail across VLMs, while calling $\pi_{0.5}$ restores execution. \textbf{(d)} Under LIBERO-PRO shifts, calling the VLA helps weaker VLMs, whereas Opus excels with 3D-tool primitives. Error bars: 95\% CIs.}
    \label{fig:action}
\end{figure}

\lead{Sub-goals beat direct pose prediction.} Having a VLM output 3D gripper poses directly fails in most episodes---even one \emph{key pose} at a time---and scaling to Opus with high reasoning effort does not close the gap (Figure~\ref{fig:action}a--c; Table~\ref{supp:tab:action}). Delegating to a VLA via language sub-goals succeeds far more often, especially when the VLM marks a 2D target point on the image (\emph{target-marked sub-goals}), lifting Flash past MME-VLA alone independently of prompt wording (Appendix~\ref{supp:fig:action}).

\lead{Code with 3D perception tools is a third option.} Instead of guessing poses, policy code can query detectors and depth for 3D primitive targets. Under LIBERO-PRO shifts where the VLA struggles, VLA calls still help Flash and Sonnet, whereas Opus performs better with 3D primitives (Figure~\ref{fig:action}d). Sub-goals stay best for contact-rich motion, noisy depth (\S\ref{sec:context}), and weaker coders.

\takeaway{1}{fig:action}{VLMs understand what to do, not how to move: even frontier models fail when predicting 3D poses directly. Instead, VLMs should delegate motion to VLAs via \textbf{language sub-goals}, especially by marking a 2D target point on the image. Under domain shifts, calling the VLA still helps weaker VLMs, whereas frontier models can write code that queries 3D perception tools to drive primitives.}

\Needspace*{8\baselineskip}
\subsection{Context and Perception Tools}
\label{sec:context}

\questionbox{Which context should VLMs and their tools receive to recall past events and read the scene?}

\begin{figure}[!htb]
    \centering
    \includegraphics[width=\textwidth]{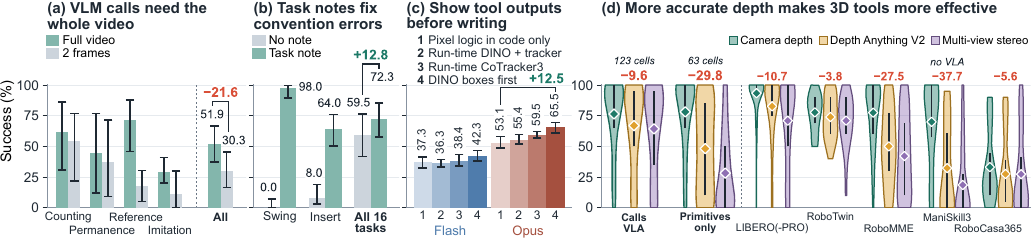}
    \caption{\textbf{Study 2 / Takeaway 2: Video history, task notes, pre-coding vision tools, and accurate depth prevent perception and convention failures.} \textbf{(a)} Passing only two frames instead of full video misses intermediate events on RoboMME (Flash). \textbf{(b)} Brief task notes resolve spatial convention ambiguities. \textbf{(c)} Extracting initial Grounding~DINO boxes before coding (4) beats raw pixel logic (1) or run-time-only tools (2--3). \textbf{(d)} Across five benchmarks under three depth sources (diamonds: means; red: monocular drop), accurate camera depth lifts 3D primitives strongly, whereas VLA-calling code stays resilient. Error bars: 95\% CIs.}
    \label{fig:context}
\end{figure}

\lead{The VLM needs full video history and task conventions.} RoboMME tasks depend on prior events, such as where an object was hidden, that a two-frame summary misses (Figure~\ref{fig:context}a). In RoboMME's Swing task, the robot's ``right'' lies on the image's left; stating this in a short task note resolves the ambiguity (Figure~\ref{fig:context}b).

\lead{Querying vision tools before coding beats coding from raw images.} Coding directly from raw images pushes the VLM toward brittle pixel heuristics such as color thresholds (Figure~\ref{fig:context}c). Calling 2D detectors only at run time yields limited gain, whereas extracting initial Grounding~DINO boxes before coding markedly lifts Opus.

\lead{More accurate depth amplifies 3D perception tools.} Over $80\%$ of evolved policy code projects 2D masks into 3D coordinates (Appendix~\ref{supp:fig:context}). Because primitives-only code plans directly from those coordinates, camera depth sharply outperforms monocular Depth Anything~V2 across five benchmarks, whereas VLA-calling code stays far more resilient by absorbing small geometric offsets through RGB control (Figure~\ref{fig:context}d; Table~\ref{supp:tab:depth_strata}). When no depth sensor is available, as on Pure-RGB RoboDojo, calibrating monocular depth against known robot and table geometry recovers metric 3D tools (Appendix~\ref{supp:protocol}).

\takeaway{2}{fig:context}{Many apparent control failures stem from \textbf{missing context or perception}, not VLM reasoning. Give VLMs full video history, task notes, and vision-tool outputs \emph{before} coding---run-time detectors alone add little. Moreover, \textbf{VLA calls buffer depth noise}, whereas primitives-only code needs accurate depth.}

\Needspace*{8\baselineskip}
\subsection{Timing of VLM Calls}
\label{sec:online}

\questionbox{Should the VLM be called many times per episode, once per episode, or once per task?}

\begin{figure}[!htb]
    \centering
    \includegraphics[width=\textwidth]{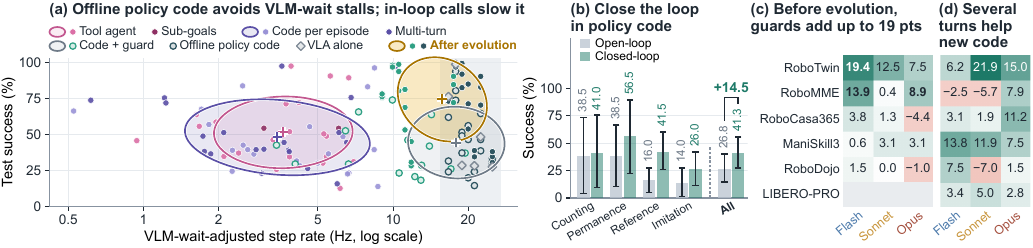}
    \caption{\textbf{Study 3 / Takeaway 3: Pre-written closed-loop code runs at native control rate, whereas in-episode VLM calls stall the robot.} \textbf{(a)} Test success vs.\ VLM-wait-adjusted step rate (gray band: simulator rate; ovals: group means $+$ and spreads). In-episode VLM calls pull step rates far below native control rates, whereas evolved \emph{Offline policy code} (gold) runs at full speed with high success. \textbf{(b)} Closed-loop code checking live feedback far beats open-loop fixed plans on RoboMME (Flash). \textbf{(c)} Before evolution, a corrective \emph{guard} aids weaker coders. \textbf{(d)} Multi-turn per-episode refinement beats a single draft (bold: significant; gray: not run; 95\% CIs).}
    \label{fig:online}
\end{figure}

\lead{Frequent VLM calls stall the robot.} Measured by \emph{VLM-wait-adjusted step rate} (control steps per wall-clock second including VLM wait; Figure~\ref{fig:online}a), in-loop sub-goals, tool agents, and per-episode coding all stall the robot while awaiting remote replies, as do prior open-loop plans, ReKep constraints, and PDDL predicates (Appendix~\ref{supp:fig:online}; Table~\ref{supp:tab:prior}). Running pre-written policy code Offline eliminates VLM wait altogether, finishing RoboMME episodes in 31--35\,s versus 43--327\,s for online designs while outperforming in-loop sub-goal planners at zero test-time VLM calls (Figure~\ref{fig:validation}c).

\lead{Check action feedback in code; call the VLM only when needed.} Once a task is decomposed into policy code, routine step-by-step execution needs no VLM reasoning: \emph{closed-loop} code checks live feedback (images, gripper state, and object positions) to branch or retry immediately, far beating \emph{open-loop} fixed plans (Figure~\ref{fig:online}b). The VLM is called only when a \emph{guard} detects a stall or failure requiring higher-level recovery (lifting Flash before evolution; Figure~\ref{fig:online}c), while multi-turn refinement similarly improves per-episode coding (Figure~\ref{fig:online}d).

\takeaway{3}{fig:online}{Routine action feedback belongs in \textbf{closed-loop policy code}, not in step-by-step VLM reasoning that stalls the robot. Such policy code checks live feedback to issue next actions immediately, calling VLMs only when a guard detects failures.}

\Needspace*{8\baselineskip}
\subsection{Policy-Code Form: Templates vs.\ Free-form}
\label{sec:forms}

\questionbox{Do fixed templates make VLM-written policy code succeed more often than free-form Python?}

\begin{figure}[!htb]
    \centering
    \includegraphics[width=\textwidth]{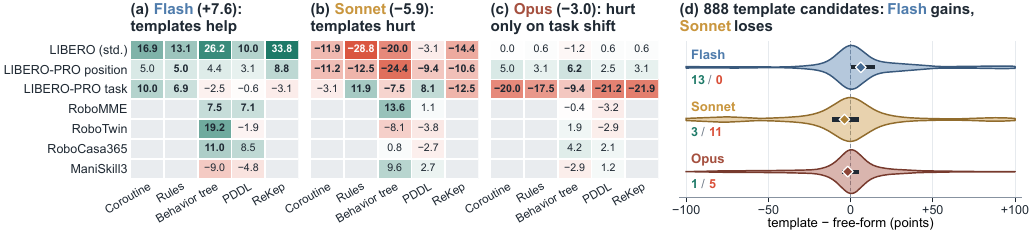}
    \caption{\textbf{Study 4 / Takeaway 4: Policy-code form should match VLM capability: structured templates guide weaker coders, whereas stronger VLMs do better with free-form Python.} \textbf{(a--c)} Point change in unevolved Offline success when replacing free-form Python with five templates across seven benchmark splits (green/red: gains/losses; bold: significant; gray: not run). Templates lift Flash (\textbf{a}), hurt Sonnet (\textbf{b}), and restrict Opus on task-shifted LIBERO-PRO (\textbf{c}). \textbf{(d)} Across 888 candidates (diamonds: means), Flash logs only significant gains, whereas Sonnet and Opus mostly lose.}
    \label{fig:forms}
\end{figure}

We compare unevolved free-form Python with five structured templates adapted to Offline execution (Figure~\ref{fig:forms}; Appendix~\ref{supp:fig:forms}): coroutines, condition--action rules~\cite{nilsson1994teleoreactive}, behavior trees~\cite{colledanchise2018bt}, PDDL~\cite{mcdermott1998pddl}, and ReKep~\cite{huang2024rekep}.

\lead{Templates guide weaker models; free-form code empowers stronger ones.} Templates aid unevolved Flash, whereas free-form Python gives Sonnet and Opus higher accuracy and flexibility under task shifts (Figure~\ref{fig:forms}a--c). Strong coders naturally structure free-form Python into compact closed-loop state machines without rigid template constraints (Appendix~\ref{supp:fig:forms}; an evolved program example appears in Figure~\ref{supp:fig:program}).

\takeaway{4}{fig:forms}{Policy-code form should \textbf{match VLM capability} rather than fixing one format: structured templates guide weaker coders, whereas free-form Python lets stronger VLMs express flexible perception and control logic without rigid constraints.}
\Needspace*{10\baselineskip}
\section{Starting from VLAs}
\label{sec:vla}

Following \S\ref{sec:vlm}, policy code is free-form and closed-loop and calls primitives and the VLA (none on ManiSkill3).

\subsection{Where Harnesses Benefit VLAs}
\label{sec:where}

\questionbox{On which tasks do harnesses boost \mbox{fine-tuned} VLAs, and when should they simply delegate?}

\begin{figure}[!htb]
    \centering
    \includegraphics[width=\textwidth]{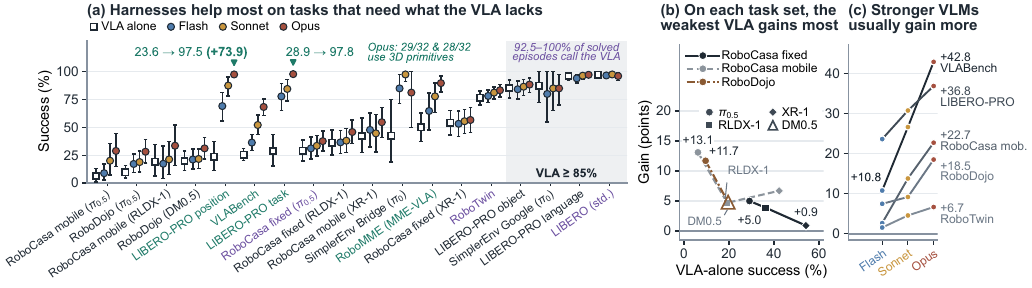}
    \caption{\textbf{Study 5 / Takeaway 5: Harnesses help most where the base VLA falls short and preserve performance where the VLA is strong.} \textbf{(a)} Across 18 benchmark--VLA pairs ($\pi_{0.5}$ unless noted) sorted by VLA-alone success (open squares), evolved Offline code (dots) lifts memory, reasoning, and shifted tasks (green, routing to 3D primitives under shifts) while delegating to strong in-distribution VLAs (violet, gray band). \textbf{(b)} Within each benchmark, weaker VLAs gain more. \textbf{(c)} Stronger coding VLMs yield larger gains over $\pi_{0.5}$. Error bars: 95\% CIs.}
    \label{fig:where}
\end{figure}

\lead{Gains depend on task demands and model capabilities.} On in-distribution suites such as standard LIBERO, RoboTwin, and fixed-base RoboCasa365, fine-tuned VLAs act reliably, so evolved code delegates directly to the VLA in nearly all solved episodes to preserve peak performance (Figure~\ref{fig:where}a; Tables~\ref{supp:tab:where} and~\ref{supp:tab:vla_use}). Conversely, harnesses bring large gains wherever tasks require what the base policy lacks---explicit memory on RoboMME, spatial or goal generalization via 3D primitives on LIBERO-PRO shifts, and multi-stage reasoning on VLABench---with weaker VLAs and stronger VLMs yielding the steepest gains (Figure~\ref{fig:where}b,c; Table~\ref{supp:tab:where_bc}).

\takeaway{5}{fig:where}{Effective harnesses act as \textbf{capability-aware routers}, not fixed wrappers. On trained skills, evolved code delegates directly to the VLA to keep peak accuracy. On memory, reasoning, or shifted tasks, it guides the VLA with sub-goals and logic, or switches to 3D primitives.}

\Needspace*{8\baselineskip}
\subsection{Policy-Code Evolution and Revision Selection}
\label{sec:improve}

\questionbox{How should policy code be evolved before testing, and which revisions should be kept?}

\begin{figure}[!htb]
    \centering
    \includegraphics[width=\textwidth]{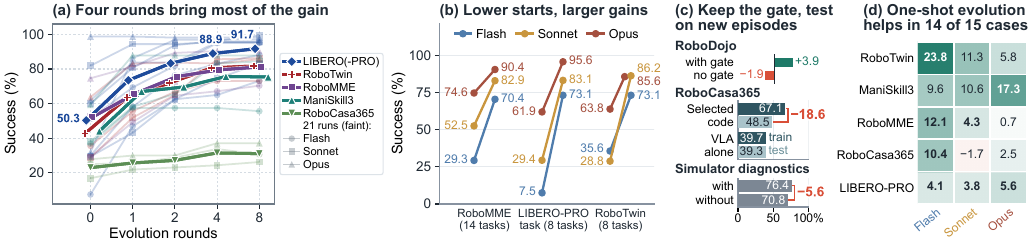}
    \caption{\textbf{Study 6 / Takeaway 6: A few rounds of pre-test evolution, guarded by a regression check, lift weak and strong VLMs alike.} \textbf{(a)} Offline test success across rounds (0: initial code; faint: 21 runs; bold: benchmark means) rises steeply by round~4 and plateaus by round~8. \textbf{(b)} Weaker coders (Flash, Sonnet) gain more by round~8, narrowing the gap to Opus. \textbf{(c)} Dropping the regression check hurts test success on RoboDojo (top, GPT-6 Astra); training-only selection generalizes to blind RoboCasa365 tests (middle, Opus); and evolution stays strong without simulator diagnostics (bottom). \textbf{(d)} One-shot evolution improves nearly all settings (bold: significant).}
    \label{fig:improve}
\end{figure}

\lead{Four rounds bring most of the gain.} Even \emph{one-shot evolution} helps almost everywhere (Figure~\ref{fig:improve}d). Gains rise sharply by round~4 and plateau by round~8 (extending to rounds~16 and~32 shifts accuracy by roughly a point; Table~\ref{supp:tab:improve_ab}), with weaker VLMs gaining most from trial feedback (Figure~\ref{fig:improve}a,b).

\lead{Evolution works without simulator diagnostics.} Action logs and binary success alone preserve most evolution gains across benchmarks, with Opus retaining nearly full accuracy (Figure~\ref{fig:improve}c, bottom). Even at zero rollouts, reusable cross-task \emph{skill cards} and VLA-fallback initialization lift starting candidates (Appendix~\ref{supp:extended}).

\lead{Revisions need a regression check.} Our selection gate accepts a revision only when training success rises without major regressions, outperforming stricter alternatives (Appendix~\ref{supp:fig:improve}). Without this gate, unverified edits degrade test accuracy on RoboDojo, whereas training-only selection reliably transfers gains to blind RoboCasa365 tests (Figure~\ref{fig:improve}c, top and middle; Table~\ref{supp:tab:improve_cd}).

\lead{The tool bank evolves too.} Holding initial policy code frozen and evolving a shared \emph{tool bank} of perception and action helpers for four rounds matches eight rounds of full code evolution (Figure~\ref{supp:fig:ablations}c; Appendix~\ref{supp:extended}).

\takeaway{6}{fig:improve}{Initial policy code misses physical edge cases even for frontier VLMs. \textbf{Evolving code from trial feedback} over a few rounds substantially lifts weak and strong models alike. Revisions require regression checks, and evolving a shared tool bank matches policy-code evolution.}

\Needspace*{8\baselineskip}
\subsection{VLM Involvement at Run Time}
\label{sec:runtime}

\questionbox{After evolution, do VLM calls at run time still help, and when should they be triggered?}

\begin{figure}[!htb]
    \centering
    \includegraphics[width=\textwidth]{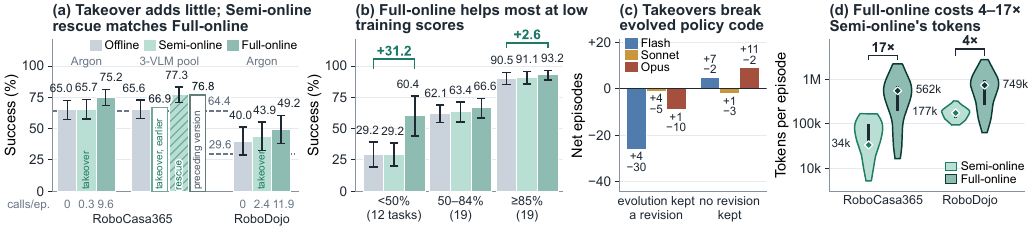}
    \caption{\textbf{Study 7 / Takeaway 7: After evolution, Semi-online multi-turn rescue matches Full-online at much lower cost, outperforming one-shot takeovers.} \textbf{(a)} With Awomo-0.5I (dashed: VLA alone), one-shot takeovers add little over Offline code, whereas Semi-online multi-turn rescue (hatched) matches Full-online with fewer calls. \textbf{(b)} On RoboCasa365, Full-online helps mainly on hard tasks ($<50\%$ training score) and adds little when Offline code is reliable ($\ge 85\%$). \textbf{(c)} One-shot takeovers break more evolved episodes than they rescue (labels: $+$/$-$ counts). \textbf{(d)} Full-online uses $4$--$17\times$ more tokens than Semi-online takeovers (violins: tasks with calls; diamonds: means; 95\% CIs).}
    \label{fig:runtime}
\end{figure}

\lead{Full-online recovers failures at high cost.} Calling the VLM at every sub-goal beats Offline execution (Figure~\ref{fig:runtime}a), but queries the VLM about ten times per episode at several times the token cost (Figure~\ref{fig:runtime}d). Nearly all its gain is on hard tasks where Offline code fails most training episodes (Figure~\ref{fig:runtime}b).

\lead{One-shot takeovers overwrite evolved fixes.} Under Semi-online execution, evolved code runs until a guard signals failure. Replacing it via an untested \emph{one-shot takeover} adds little overall (Figure~\ref{fig:runtime}a) and breaks more evolved episodes than it rescues across all three VLMs, as weaker models over-trigger (Figure~\ref{fig:runtime}c; Table~\ref{supp:tab:runtime}).

\lead{Semi-online multi-turn rescue matches Full-online.} Rather than overwriting evolved code, \emph{multi-turn rescue} invokes the VLM step by step after a physical failure trigger---inspecting each image and action outcome before choosing the next recovery step. On long-horizon RoboCasa365 tasks (Figure~\ref{fig:runtime}a), this matches Full-online and far beats one-shot takeovers while cutting VLM token usage by $42\%$, because over half of test episodes finish cleanly Offline without triggering rescue (Appendix~\ref{supp:extended}).

\lead{Closed-loop code withstands disturbances.} Because evolved closed-loop code checks state feedback and retries failed sub-goals, it keeps a $>20$-point lead over the base VLA when objects are unexpectedly shifted or dropped mid-episode across four benchmarks. It likewise preserves its advantage under RoboTwin~2.0 clutter, lighting, and height shifts, while a lightweight Semi-online guard resolves remaining stalls (Appendix~\ref{supp:extended}).

\takeaway{7}{fig:runtime}{Semi-online execution---running evolved code autonomously and calling the VLM \textbf{only on failure signals}---best balances speed and success. Offline code handles routine retries at native speed, while multi-turn rescue adds adaptive recovery without Full-online's latency or one-shot overwrites.}

\FloatBarrier
\section{Do the Takeaways Compose?}
\label{sec:recipe}

\subsection{Benchmark Composition and Scaling}
\label{sec:recipe_main}

\begin{table}[!htbp]
    \centering
    \caption{\textbf{Summary of the seven studies as a conditional design guide.} Effect: alternative $\rightarrow$ our choice in success (\%) or wait-adjusted step rate (Hz); T4: template gain, Flash/Sonnet/Opus; T6: rounds 0/4/8; T7: Offline $\rightarrow$ rescue. $^{*}$Significant. Full numerical tables in Appendix~\ref{supp:figs}.}
    \label{tab:recipe}
    \footnotesize
    \setlength{\tabcolsep}{4.5pt}
    \renewcommand{\arraystretch}{1.0}
    \newcommand{\tk}[1]{\makebox[1.7em][l]{\textbf{#1}}}
    \begin{adjustbox}{max width=\textwidth}
    \begin{tabular}{@{}l l l l l@{}}
        \toprule
        \tk{}\textbf{Observed problem} & \textbf{Tested alternatives} & \textbf{Conditional harness choice} & \textbf{Measured effect} & \textbf{Fig.} \\
        \midrule
        \multicolumn{5}{@{}l}{\emph{Starting from VLMs} (\S\ref{sec:vlm})} \\
        \tk{T1}VLM poses fail on the robot & Poses; unmarked sub-goals & Sub-goals; 3D-target code if depth is accurate & $17.3 \rightarrow 51.9$ (Flash) & \ref{fig:action}a \\
        \tk{T2}VLMs misread the scene & Two frames; estimated depth & Full video, task note, tool outputs first, RGB-D & $30.3 \rightarrow 51.9$; depth $+16.4^{*}$ & \ref{fig:context} \\
        \tk{T3}VLM calls stall the robot & In-loop calls; per-episode code & Closed-loop policy code, no in-loop VLM call & $0.5$--$10 \rightarrow 16.7$--$25$\,Hz & \ref{fig:online}a \\
        \tk{T4}Weak VLMs' code often fails & Five symbolic templates & Templates for weak VLMs; free-form for strong & $+7.6$ / $-5.9$ / $-3.0$ & \ref{fig:forms} \\
        \midrule
        \multicolumn{5}{@{}l}{\emph{Starting from VLAs} (\S\ref{sec:vla})} \\
        \tk{T5}VLAs fail on shifted tasks & VLA alone; harness always & Delegate to strong VLAs; add logic on shifts & $23.6 \rightarrow 97.5$ (Opus) & \ref{fig:where}a \\
        \tk{T6}Initial policy code often fails & No evolution; ungated & With training episodes: 4--8 gated rounds & $50.3 \rightarrow 88.9 \rightarrow 91.7^{*}$ & \ref{fig:improve}a--c \\
        \tk{T7}Residual failures after evolution & Offline; takeovers; Full-online & Semi-online: multi-turn rescue only on triggers & $65.6 \rightarrow 77.3$ (pool) & \ref{fig:runtime} \\
        \bottomrule
    \end{tabular}
    \end{adjustbox}
\end{table}

\begin{table}[!htbp]
    \centering
    \caption{\textbf{The takeaways compose: one reference recipe on eight benchmarks.} Test success (\%) of VLAs alone and in X-Harness. \emph{Official}: leaderboard; \emph{Rerun}: same VLA, our protocol. Top: Offline code after eight evolution rounds; \emph{Gain}: Opus minus \emph{Rerun} (unrounded). Bottom: tiers of \S\ref{sec:runtime}; \emph{pool}: added tools, one training-selected candidate per task; \emph{Gain}: best tier minus \emph{Official}. Bold: best.}
    \label{tab:results}
    \footnotesize
    \setlength{\tabcolsep}{5.0pt}
    \renewcommand{\arraystretch}{1.0}
    \begin{adjustbox}{max width=\textwidth}
    \begin{tabular}{@{}l l l c c c c c c@{}}
        \toprule
        \multicolumn{3}{@{}l}{\emph{Single-VLM reference recipe across all eight benchmarks (267 tasks)}} & \multicolumn{2}{c}{\textbf{VLA alone}} & \multicolumn{3}{c}{\textbf{X-Harness (Offline), policy code by}} & \\
        \cmidrule(lr){4-5}\cmidrule(lr){6-8}
        \textbf{Benchmark (tasks)} & \textbf{What it tests} & \textbf{VLA} & Official & Rerun & Flash & Sonnet & Opus & \textbf{Gain} \\
        \midrule
        RoboMME (16) & memory & MME-VLA$^{\dagger}$ & 44.5 & 50.0 & 64.7 & 77.8 & \best{89.7} & \up{39.7} \\
        LIBERO-PRO (40 $\times$ 4 shifts) & four shifts, pooled & $\pi_{0.5}$ & -- & 58.6 & 82.2 & 89.3 & \best{95.4} & \up{36.8} \\
        LIBERO, standard (40) & in-distribution & $\pi_{0.5}$ & -- & 97.2 & 96.6 & \best{97.5} & 96.1 & \dn{1.1} \\
        RoboTwin 2.0 (50) & two arms & $\pi_{0.5}$ & 70.7 & 76.6 & 75.9 & 83.4 & \best{88.6} & \up{12.0} \\
        RoboCasa365, fixed base (37) & kitchen tasks & $\pi_{0.5}$ & \multirow{2}{*}{16.9$^{\ddagger}$} & 29.1 & 31.1 & 33.4 & \best{37.6} & \up{8.5} \\
        RoboCasa365, mobile base (13) & kitchen, mobile base & $\pi_{0.5}$ & & 6.2 & 8.8 & 20.0 & \best{28.8} & \up{22.7} \\
        RoboDojo (42) & five skill axes & $\pi_{0.5}$ & 6.9 & 9.6 & 17.1 & 18.8 & \best{28.1} & \up{18.5} \\
        RoboDojo (42) & five skill axes & DM0.5 & 19.3 & 19.6 & 21.1 & 21.6 & \best{31.1} & \up{11.4} \\
        VLABench (49) & language, common sense & $\pi_{0.5}$ & -- & 25.5 & 36.3 & 52.1 & \best{68.3} & \up{42.8} \\
        SimplerEnv (6) & two robot types & $\pi_{0}$ & -- & 57.5 & 83.3 & \best{93.3} & 82.5 & \up{25.0} \\
        ManiSkill3 (14) & contact skills & none$^{\S}$ & -- & -- & 69.6 & 81.8 & \best{89.3} & -- \\
        \midrule
        \multicolumn{5}{@{}l}{\emph{Scaling with stronger base policies and a three-VLM pool (Opus, GPT-6 Astra, Argon)}} & \textbf{Offline} & \textbf{Semi-online} & \textbf{Full-online} & \\
        RoboCasa365 (50), Argon & kitchen, both bases & Awomo-0.5I$^{\P}$ & 64.4 & -- & 65.0 & 65.7 & \best{75.2} & \up{10.8} \\
        RoboCasa365 (50), pool & kitchen, both bases & Awomo-0.5I$^{\P}$ & 64.4 & -- & 65.6 & \best{77.3}\rlap{$^{\|}$} & -- & \up{12.9} \\
        RoboDojo (42), Argon & five skill axes & Awomo-0.5I$^{\P}$ & 29.6 & -- & 40.0 & 43.9 & \best{49.2} & \up{19.6} \\
        RoboDojo (42), pool & five skill axes & Awomo-0.5I$^{\P}$ & 29.6 & -- & 45.1 & 51.4 & \best{53.2} & \up{23.6} \\
        RoboDojo (42), pool & five skill axes & WALL-OSS-1.0-lite$^{\P}$ & 35.0 & -- & 57.0 & \best{62.2}\rlap{$^{\|}$} & -- & \up{27.2} \\
        \bottomrule
    \end{tabular}
    \end{adjustbox}

    \vspace{2pt}
    \parbox[t]{\textwidth}{\footnotesize $^{\dagger}$Harness calls the sub-goal MME-VLA variant (Table~\ref{supp:tab:vla}, Appendix~\ref{supp:models}); \emph{Official}, \emph{Rerun}: strongest memory variant. $^{\ddagger}$All 50 tasks (\emph{Rerun}: $23.1$; full-body interface: $26.7$ alone, $36.0$ Opus Offline). $^{\S}$Primitives only. $^{\P}$\emph{Official} leaderboard reference for matching weights (Awomo-0.5I / WALL-OSS-1.0-lite). $^{\|}$Multi-turn rescue stage (replaces Full-online); other rows use one-shot takeovers. WALL-OSS-1.0-lite reports five-axis macro mean ($57.0$, $62.4$ pooled); Awomo-0.5I reports pooled success.}
\end{table}

\textbf{X-Harness} combines the seven takeaways in Table~\ref{tab:recipe}. Without updating any model weights, it evolves closed-loop Python policy code on 5--40 training episodes per task. At test time, six benchmarks run purely Offline while the two longest-horizon suites add trigger-gated Semi-online rescue (\S\ref{sec:runtime}).

\lead{Single-VLM recipe across eight benchmarks.} Varying only the coding VLM across all 267 tasks in Table~\ref{tab:results} (top) lifts every VLM over the base policy, with stronger coders consistently unlocking larger gains. Improvements concentrate where base policies lack explicit memory on RoboMME, multi-stage reasoning on VLABench, or spatial and goal generalization on LIBERO-PRO shifts, while preserving peak in-distribution accuracy on standard LIBERO (Appendix~\ref{supp:extended}).

\lead{Scaling with stronger base policies, candidate pools, and selective rescue.} Table~\ref{tab:results} (bottom) extends X-Harness to stronger base policies on RoboCasa365 and Pure-RGB RoboDojo. Evolved Offline code already surpasses official leaderboards on every suite, and combining shared verification tools with a three-VLM candidate \emph{pool} boosts Offline accuracy further (Table~\ref{supp:tab:pool}). Step-level logs confirm that pooled code still invokes Awomo-0.5I in nearly every solved RoboCasa365 episode, orchestrating rather than replacing strong visuomotor control (Table~\ref{supp:tab:vla_use_pool}). Adding trigger-gated Semi-online multi-turn rescue closes remaining long-horizon failures while leaving over half of test episodes completely Offline, and enabling rescue \emph{selectively} on harder training-identified tasks nearly matches all-task rescue at lower compute (Table~\ref{supp:tab:selective_rescue}; qualitative rollouts in Figures~\ref{supp:fig:qualitative_rd}--\ref{supp:fig:qualitative_rc}).

\subsection{Generalization to Unseen Seeds, Visual Shifts, and Edge Deployment}
\label{sec:edge_gen}

\begin{figure}[!htb]
    \centering
    \includegraphics[width=0.95\textwidth]{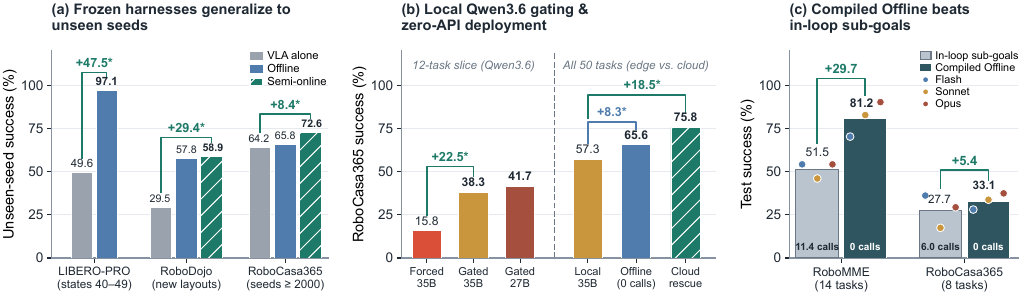}
    \vspace{-4pt}
    \caption{\textbf{Generalization to unseen seeds, zero-API local edge models, and in-loop sub-goal baselines.} \textbf{(a)} Frozen X-Harness policies reproduce large, statistically significant gains on brand-new unseen seeds and layouts (Table~\ref{supp:tab:fresh_seeds}; $^*$95\% paired bootstrap CI strictly above zero). \textbf{(b)} Zero-API quantized local Qwen3.6 deployment on RoboCasa365 (Table~\ref{supp:tab:local_qwen}): selective gating lifts local 35B over forced step-1 takeover (left), while compiled Offline policy code beats local 35B online rescue at zero hosting overhead across all 50 tasks (right). \textbf{(c)} Compiled Offline code outperforms multi-call in-loop sub-goal planners across three VLMs on RoboMME and RoboCasa365 (bars: mean; dots: per VLM).}
    \label{fig:validation}
\end{figure}

\lead{Unseen test seeds and visual shifts.} Evaluated frozen across 2,080 newly sampled unseen episodes per arm on LIBERO-PRO, Pure-RGB RoboDojo, and RoboCasa365, X-Harness sustains significant gains over the base VLA without seed leakage (Figure~\ref{fig:validation}a; Table~\ref{supp:tab:fresh_seeds}). Deployed unchanged under appearance and environmental shifts, frozen Offline code matches or exceeds $\pi_0$ across all official SimplerEnv conditions---even with monocular depth---and beats $\pi_{0.5}$ across all 50 domain-randomized RoboTwin~2.0 tasks (Tables~\ref{supp:tab:simpler_va} and~\ref{supp:tab:rand}).

\lead{Zero-API edge deployment with local open-weight models.} Our capability-matching rule transfers directly to offline edge hardware running quantized local Qwen3.6 models behind the frozen RoboCasa365 rescue interface (Figure~\ref{fig:validation}b; Appendix~\ref{supp:models}). Failure-triggered gating remains essential over step-1 takeover, yet across all 50 tasks compiled Offline code beats local 35B online rescue without local-VLM overhead---making it the strongest zero-API edge option today, while cloud rescue unlocks further headroom (Table~\ref{supp:tab:local_qwen}). Appendix~\ref{supp:real_robot} further verifies unchanged transfer to a physical bimanual YAM robot paired with adapted-$\pi_{0.5}$ servers.
\FloatBarrier
\section{Conclusion}
\label{sec:conclusion}

We present a large-scale empirical study of policy-harness design for robot-use agents, spanning over 4,000 configurations and 470,000 rollouts across eight benchmarks (267 tasks), six VLMs, and 15 frozen VLA and WAM checkpoints. Rather than committing to one fixed architecture, our seven controlled studies turn harness design into an actionable, capability-matched guide: effective harnesses ground perception before coding, match code structure to VLM coding strength, compile feedback into closed-loop code, apply harness logic where the base policy falls short while delegating cleanly where it excels, evolve policy code or shared tools under regression checks, and reserve test-time VLM calls for trigger-gated multi-turn rescue.

Composed without updating any model weights, these seven principles yield \mbox{\textbf{X-Harness}}, a reference recipe that substantially lifts frozen VLAs and WAMs across explicit memory, multi-stage reasoning, spatial and goal shifts, Pure-RGB manipulation, and long-horizon mobile tasks while preserving peak in-distribution accuracy. X-Harness generalizes cleanly to unseen test seeds and visual shifts, delivers strong zero-API execution at native control rates, and transfers directly to physical bimanual hardware (Appendix~\ref{supp:real_robot}). By releasing our unified evaluation suite, evolved policy code, and 470,000+ rollout logs at \href{https://github.com/google-research/x-as-policies}{\texttt{github.com/google-research/x-as-policies}}, we establish reproducible baselines, durable design principles, and an empirical foundation for building general-purpose agentic robots.

\FloatBarrier

{\small
\setlength{\bibsep}{1.75pt}
\bibliographystyle{ieeenat_fullname}
\bibliography{main}

\begin{thebibliography}{53}
\providecommand{\natexlab}[1]{#1}
\providecommand{\url}[1]{\texttt{#1}}
\expandafter\ifx\csname urlstyle\endcsname\relax
  \providecommand{\doi}[1]{doi: #1}\else
  \providecommand{\doi}{doi: \begingroup \urlstyle{rm}\Url}\fi

\bibitem[{Awomo-WAM Team}(2026)]{awomo2026}
{Awomo-WAM Team}.
\newblock {Awomo-0.5} technical report.
\newblock \url{https://github.com/Awomo-WestlakeDI/Awomo-0.5}, 2026.
\newblock Official RoboCasa365 and RoboDojo Leaderboards
  (\url{https://robocasa.ai/leaderboard.html},
  \url{https://robodojo-benchmark.com}).

\bibitem[Black et~al.(2025{\natexlab{a}})Black, Brown, Darpinian, Dhabalia,
  Driess, Esmail, Equi, Finn, Fusai, Galliker, et~al.]{intelligence2025pi05}
Kevin Black, Noah Brown, James Darpinian, Karan Dhabalia, Danny Driess, Adnan
  Esmail, Michael~Robert Equi, Chelsea Finn, Niccolo Fusai, Manuel~Y. Galliker,
  et~al.
\newblock $\pi_{0.5}$: A {Vision-Language-Action} model with open-world
  generalization.
\newblock In \emph{Conference on Robot Learning (CoRL)}, pages 17--40,
  2025{\natexlab{a}}.

\bibitem[Black et~al.(2025{\natexlab{b}})Black, Brown, Driess, Esmail, Equi,
  Finn, Fusai, Groom, Hausman, Ichter, et~al.]{black2024pi0}
Kevin Black, Noah Brown, Danny Driess, Adnan Esmail, Michael~Robert Equi,
  Chelsea Finn, Niccolo Fusai, Lachy Groom, Karol Hausman, Brian Ichter, et~al.
\newblock $\pi_0$: A {Vision-Language-Action} flow model for general robot
  control.
\newblock In \emph{Robotics: Science and Systems (RSS)}, 2025{\natexlab{b}}.

\bibitem[Chen et~al.(2026{\natexlab{a}})Chen, Chen, Li, Tang, Su, Lu, Wan,
  Chen, Liu, Yan, et~al.]{robodojo2025}
Tianxing Chen, Yue Chen, Zixuan Li, Junyuan Tang, Kailun Su, Haoran Lu, Weijie
  Wan, Baijun Chen, Songling Liu, Haowen Yan, et~al.
\newblock {RoboDojo}: A unified sim-and-real benchmark for comprehensive
  evaluation of generalist robot manipulation policies.
\newblock \emph{arXiv preprint arXiv:2607.04434}, 2026{\natexlab{a}}.

\bibitem[Chen et~al.(2026{\natexlab{b}})Chen, Chen, Chen, Cai, Liu, Li, Liang,
  Lin, Ge, Gu, et~al.]{mu2025robotwin2}
Tianxing Chen, Zanxin Chen, Baijun Chen, Zijian Cai, Yibin Liu, Zixuan Li,
  Qiwei Liang, Xianliang Lin, Yiheng Ge, Zhenyu Gu, et~al.
\newblock {RoboTwin 2.0}: A scalable data generator and benchmark with strong
  domain randomization for robust bimanual robotic manipulation.
\newblock In \emph{International Conference on Machine Learning (ICML)}, pages
  13673--13699, 2026{\natexlab{b}}.

\bibitem[Chen et~al.(2026{\natexlab{c}})Chen, Mo, Chen, Lu, Kuang, and
  Li]{astraguide2026}
Xingping Chen, Zihao Mo, Yimin Chen, Yu Lu, Zhanghui Kuang, and Guanbin Li.
\newblock Can {Astra} guide a {VLA} to solve {RoboMME}? {A} self-evolving
  harness for memory-dependent manipulation.
\newblock \url{https://legend717.github.io/astra-guide-vla-website/},
  2026{\natexlab{c}}.
\newblock Technical report.

\bibitem[Chu et~al.(2026)Chu, Barr, Le~Goues, O'Hearn, Harman, Sarro, and
  Ye]{openrua2026}
Zhaoyang Chu, Earl~T. Barr, Claire Le~Goues, Peter O'Hearn, Mark Harman,
  Federica Sarro, and He Ye.
\newblock {OpenRUA}: {Robot-Use} agents are zero-shot visuomotor policies.
\newblock \emph{arXiv preprint arXiv:2610.02459}, 2026.

\bibitem[Colledanchise and {\"O}gren(2018)]{colledanchise2018bt}
Michele Colledanchise and Petter {\"O}gren.
\newblock \emph{Behavior Trees in Robotics and {AI}: An Introduction}.
\newblock CRC Press, 2018.

\bibitem[Dai et~al.(2026)Dai, Fu, Lee, Liu, Zhang, Yang, Finn, Fazeli, and
  Chai]{robomme2025}
Yinpei Dai, Hongze Fu, Jayjun Lee, Yuejiang Liu, Haoran Zhang, Jianing Yang,
  Chelsea Finn, Nima Fazeli, and Joyce Chai.
\newblock {RoboMME}: Benchmarking and understanding memory for robotic
  generalist policies.
\newblock In \emph{International Conference on Machine Learning (ICML)}, pages
  22396--22450, 2026.

\bibitem[{Dexmal Team}(2026)]{dm05_2026}
{Dexmal Team}.
\newblock {OpenDM}: An open-world foundation model for general-purpose embodied
  intelligence.
\newblock \url{https://github.com/dexmal/opendm}, 2026.
\newblock DM0.5 release; official RoboDojo leaderboard baseline
  (\url{https://robodojo-benchmark.com}).

\bibitem[Fu et~al.(2026)Fu, Yu, El-Refai, Kou, Xue, Huang, Xiao, Fei-Fei, Shi,
  Wu, et~al.]{capx2026}
Letian Fu, Justin Yu, Karim El-Refai, Ethan Kou, Haoru Xue, Huang Huang, Wenli
  Xiao, Li Fei-Fei, Guanya Shi, Jiajun Wu, et~al.
\newblock {CaP-X}: A framework for benchmarking and improving coding agents for
  robot manipulation.
\newblock In \emph{International Conference on Machine Learning (ICML)}, pages
  32155--32212, 2026.

\bibitem[Gao et~al.(2026)Gao, Zeng, Zhou, Ge, Shangguan, Wang, Wan, Zheng, Zhu,
  Li, et~al.]{hexashell2026}
Hongcheng Gao, Jianshu Zeng, Alex Zhou, Shijia Ge, Xuan Shangguan, Yazhe Wang,
  Yexing Wan, Zelin Zheng, Jay Zhu, Xiping Li, et~al.
\newblock {HexaShell}: A universal harness for robot coding agents in
  simulation and the real world.
\newblock \emph{arXiv preprint arXiv:2610.23479}, 2026.

\bibitem[Hu et~al.(2026)Hu, Shridhar, Lu, Shah, Chiang, Tan, and
  Xie]{hu2026orchestrating}
Jiaheng Hu, Mohit Shridhar, Caden Lu, Dhruv Shah, Hao-Tien~Lewis Chiang, Jie
  Tan, and Annie Xie.
\newblock What matters in orchestrating robot policies: A systematic study of
  hierarchical {VLA} agents.
\newblock \emph{arXiv preprint arXiv:2606.10267}, 2026.

\bibitem[Huang et~al.(2023)Huang, Wang, Zhang, Li, Wu, and
  Fei-Fei]{huang2023voxposer}
Wenlong Huang, Chen Wang, Ruohan Zhang, Yunzhu Li, Jiajun Wu, and Li Fei-Fei.
\newblock {VoxPoser}: Composable {3D} value maps for robotic manipulation with
  language models.
\newblock In \emph{Conference on Robot Learning (CoRL)}, pages 540--562, 2023.

\bibitem[Huang et~al.(2024)Huang, Wang, Li, Zhang, and Fei-Fei]{huang2024rekep}
Wenlong Huang, Chen Wang, Yunzhu Li, Ruohan Zhang, and Li Fei-Fei.
\newblock {ReKep}: Spatio-temporal reasoning of relational keypoint constraints
  for robotic manipulation.
\newblock In \emph{Conference on Robot Learning (CoRL)}, pages 4573--4602,
  2024.

\bibitem[Isola(2026)]{isola2026robotuse}
Phillip Isola.
\newblock {Robot-Use} agents.
\newblock Blog post,
  \url{https://web.mit.edu/phillipi/www/writing/robot-use-agents.html}, 2026.

\bibitem[Karaev et~al.(2025)Karaev, Makarov, Wang, Neverova, Vedaldi, and
  Rupprecht]{karaev2024cotracker3}
Nikita Karaev, Iurii Makarov, Jianyuan Wang, Natalia Neverova, Andrea Vedaldi,
  and Christian Rupprecht.
\newblock {CoTracker3}: Simpler and better point tracking by pseudo-labelling
  real videos.
\newblock In \emph{IEEE/CVF International Conference on Computer Vision
  (ICCV)}, pages 6013--6022, 2025.

\bibitem[Kim et~al.(2026)Kim, Jang, Koo, Jang, Kim, Kim, Yoon, Jang, Choi, Han,
  et~al.]{rldx2026}
Dongyoung Kim, Huiwon Jang, Myungkyu Koo, Suhyeok Jang, Taeyoung Kim, Beomjun
  Kim, Byungjun Yoon, Changsung Jang, Daewon Choi, Dongsu Han, et~al.
\newblock {RLDX-1} technical report.
\newblock \emph{arXiv preprint arXiv:2605.03269}, 2026.

\bibitem[Kim et~al.(2024)Kim, Pertsch, Karamcheti, Xiao, Balakrishna, Nair,
  Rafailov, Foster, Sanketi, Vuong, et~al.]{kim2024openvla}
Moo~Jin Kim, Karl Pertsch, Siddharth Karamcheti, Ted Xiao, Ashwin Balakrishna,
  Suraj Nair, Rafael Rafailov, Ethan~P. Foster, Pannag~R. Sanketi, Quan Vuong,
  et~al.
\newblock {OpenVLA}: An open-source {Vision-Language-Action} model.
\newblock In \emph{Conference on Robot Learning (CoRL)}, pages 2679--2713,
  2024.

\bibitem[Li et~al.(2026{\natexlab{a}})Li, Chi, Zheng, Liu, Wang, Gao, Jiang,
  Hong, Xiao, Yang, et~al.]{embodiedevo2026}
Wenyi Li, Yizhe Chi, Youjie Zheng, Chunyu Liu, Xiaoqiu Wang, Mingju Gao, Zheng
  Jiang, Deyao Hong, Calvin Xiao, Kaisen Yang, et~al.
\newblock {EmbodiedEvo}: Self-evolving coding agent for physical intelligence.
\newblock \emph{arXiv preprint arXiv:2610.18867}, 2026{\natexlab{a}}.

\bibitem[Li et~al.(2024)Li, Hsu, Gu, Mees, Pertsch, Walke, Fu, Lunawat, Sieh,
  Kirmani, et~al.]{li2024simplerenv}
Xuanlin Li, Kyle Hsu, Jiayuan Gu, Oier Mees, Karl Pertsch, Homer~Rich Walke,
  Chuyuan Fu, Ishikaa Lunawat, Isabel Sieh, Sean Kirmani, et~al.
\newblock Evaluating real-world robot manipulation policies in simulation.
\newblock In \emph{Conference on Robot Learning (CoRL)}, pages 3705--3728,
  2024.

\bibitem[Li et~al.(2026{\natexlab{b}})Li, Zhang, Li, Huang, Wang, Zhang, Deng,
  Wei, Zhang, Gao, et~al.]{li2026roboharness}
Zexi Li, Yehang Zhang, Wenqian Li, Haojian Huang, Chenxu Wang, Chunyang Zhang,
  Shiyuan Deng, Yangkai Wei, Tianyi Zhang, Ruize Gao, et~al.
\newblock {Robo-Harness K1}: Harnessing {Robot-Use} agents via perception
  augmentation.
\newblock \emph{arXiv preprint arXiv:2609.29389}, 2026{\natexlab{b}}.

\bibitem[Liang et~al.(2023)Liang, Huang, Xia, Xu, Hausman, Ichter, Florence,
  and Zeng]{liang2023codeaspolicies}
Jacky Liang, Wenlong Huang, Fei Xia, Peng Xu, Karol Hausman, Brian Ichter, Pete
  Florence, and Andy Zeng.
\newblock {Code as Policies}: Language model programs for embodied control.
\newblock In \emph{IEEE International Conference on Robotics and Automation
  (ICRA)}, pages 9493--9500, 2023.

\bibitem[Liu et~al.(2023)Liu, Zhu, Gao, Feng, Liu, Zhu, and
  Stone]{liu2023libero}
Bo Liu, Yifeng Zhu, Chongkai Gao, Yihao Feng, Qiang Liu, Yuke Zhu, and Peter
  Stone.
\newblock {LIBERO}: Benchmarking knowledge transfer for lifelong robot
  learning.
\newblock In \emph{Advances in Neural Information Processing Systems (NeurIPS),
  Datasets and Benchmarks Track}, pages 44776--44791, 2023.

\bibitem[Liu et~al.(2024)Liu, Zeng, Ren, Li, Zhang, Yang, Jiang, Li, Yang, Su,
  et~al.]{liu2023groundingdino}
Shilong Liu, Zhaoyang Zeng, Tianhe Ren, Feng Li, Hao Zhang, Jie Yang, Qing
  Jiang, Chunyuan Li, Jianwei Yang, Hang Su, et~al.
\newblock {Grounding DINO}: Marrying {DINO} with grounded pre-training for
  open-set object detection.
\newblock In \emph{European Conference on Computer Vision (ECCV)}, pages
  38--55, 2024.

\bibitem[Lu et~al.(2026)Lu, Wu, Kou, Lin, Fu, Xiao, Mandlekar, Xu, Shi,
  Goldberg, et~al.]{aspire2026}
Runyu Lu, Yubo Wu, Ethan Kou, Haotian Lin, Letian Fu, Wenli Xiao, Ajay
  Mandlekar, Yinzhen Xu, Guanya Shi, Ken Goldberg, et~al.
\newblock {ASPIRE}: Agentic skills discovery for robotics.
\newblock \emph{arXiv preprint arXiv:2607.00272}, 2026.

\bibitem[McDermott et~al.(1998)McDermott, Ghallab, Howe, Knoblock, Ram, Veloso,
  Weld, and Wilkins]{mcdermott1998pddl}
Drew McDermott, Malik Ghallab, Adele Howe, Craig Knoblock, Ashwin Ram, Manuela
  Veloso, Daniel Weld, and David Wilkins.
\newblock {PDDL}---the planning domain definition language.
\newblock Technical Report CVC TR-98-003/DCS TR-1165, Yale Center for
  Computational Vision and Control, 1998.
\newblock Version 1.2.

\bibitem[Nasiriany et~al.(2026)Nasiriany, Nasiriany, Maddukuri, and
  Zhu]{nasiriany2026robocasa365}
Soroush Nasiriany, Sepehr Nasiriany, Abhiram Maddukuri, and Yuke Zhu.
\newblock {RoboCasa365}: A large-scale simulation framework for training and
  benchmarking generalist robots.
\newblock In \emph{International Conference on Learning Representations
  (ICLR)}, pages 98643--98667, 2026.

\bibitem[Nilsson(1994)]{nilsson1994teleoreactive}
Nils~J. Nilsson.
\newblock Teleo-reactive programs for agent control.
\newblock \emph{Journal of Artificial Intelligence Research}, 1:\penalty0
  139--158, 1994.

\bibitem[{NVIDIA}(2025)]{nvidia2023isaacsim}
{NVIDIA}.
\newblock {NVIDIA Isaac Sim}.
\newblock \url{https://developer.nvidia.com/isaac/sim}, 2025.
\newblock Version 5.1.

\bibitem[{NVIDIA} et~al.(2025){NVIDIA}, Bjorck, Casta{\~n}eda, Cherniadev, Da,
  Ding, Fan, Fang, Fox, Hu, et~al.]{bjorck2025gr00t}
{NVIDIA}, Johan Bjorck, Fernando Casta{\~n}eda, Nikita Cherniadev, Xingye Da,
  Runyu Ding, Linxi Fan, Yu Fang, Dieter Fox, Fengyuan Hu, et~al.
\newblock {GR00T N1}: An open foundation model for generalist humanoid robots.
\newblock \emph{arXiv preprint arXiv:2503.14734}, 2025.

\bibitem[Qin et~al.(2026)Qin, Wu, Feng, Wang, Yu, Sun, Fan, He, Chen, Lian,
  et~al.]{qin2026towardsagenticrobots}
Chonghan Qin, Dongming Wu, Kaituo Feng, Jiaxu Wang, Xinlei Yu, Peiwen Sun,
  Yijia Fan, Xilin He, Bingao Chen, Weitong Lian, et~al.
\newblock Towards agentic robots: Model, data, environment, and harness.
\newblock \emph{arXiv preprint arXiv:2610.20442}, 2026.

\bibitem[Ravi et~al.(2025)Ravi, Gabeur, Hu, Hu, Ryali, Ma, Khedr, R{\"a}dle,
  Rolland, Gustafson, et~al.]{ravi2024sam2}
Nikhila Ravi, Valentin Gabeur, Yuan-Ting Hu, Ronghang Hu, Chaitanya Ryali,
  Tengyu Ma, Haitham Khedr, Roman R{\"a}dle, Chloe Rolland, Laura Gustafson,
  et~al.
\newblock {SAM 2}: Segment anything in images and videos.
\newblock In \emph{International Conference on Learning Representations
  (ICLR)}, pages 28085--28128, 2025.

\bibitem[Shi et~al.(2025)Shi, Ichter, Equi, Ke, Pertsch, Vuong, Tanner,
  Walling, Wang, Fusai, et~al.]{shi2025hirobot}
Lucy~Xiaoyang Shi, Brian Ichter, Michael Equi, Liyiming Ke, Karl Pertsch, Quan
  Vuong, James Tanner, Anna Walling, Haohuan Wang, Niccolo Fusai, et~al.
\newblock {Hi Robot}: Open-ended instruction following with hierarchical
  {Vision-Language-Action} models.
\newblock In \emph{International Conference on Machine Learning (ICML)}, pages
  54919--54933, 2025.

\bibitem[Singh et~al.(2023)Singh, Blukis, Mousavian, Goyal, Xu, Tremblay, Fox,
  Thomason, and Garg]{singh2023progprompt}
Ishika Singh, Valts Blukis, Arsalan Mousavian, Ankit Goyal, Danfei Xu, Jonathan
  Tremblay, Dieter Fox, Jesse Thomason, and Animesh Garg.
\newblock {ProgPrompt}: Generating situated robot task plans using large
  language models.
\newblock In \emph{IEEE International Conference on Robotics and Automation
  (ICRA)}, pages 11523--11530, 2023.

\bibitem[Song et~al.(2026)Song, Liang, Fu, Wang, Yang, and
  Liu]{embodiedrsi2026}
Python Song, Zhixuan Liang, Kelsey Fu, Mengdi Wang, Junfeng Yang, and Shilong
  Liu.
\newblock {EmbodiedRSI}: Active continual robot learning through
  hypothesis-guided co-evolution.
\newblock \emph{arXiv preprint arXiv:2610.10498}, 2026.

\bibitem[Sridhar et~al.(2026)Sridhar, Pan, Sharma, and Finn]{memer2026}
Ajay Sridhar, Jennifer Pan, Satvik Sharma, and Chelsea Finn.
\newblock Scaling up memory for robotic control via experience retrieval.
\newblock In \emph{International Conference on Learning Representations
  (ICLR)}, pages 97142--97166, 2026.

\bibitem[Sundaralingam et~al.(2023)Sundaralingam, Hari, Fishman, Garrett,
  Van~Wyk, Blukis, Millane, Oleynikova, Handa, Ramos,
  et~al.]{sundaralingam2023curobo}
Balakumar Sundaralingam, Siva Kumar~Sastry Hari, Adam Fishman, Caelan Garrett,
  Karl Van~Wyk, Valts Blukis, Alexander Millane, Helen Oleynikova, Ankur Handa,
  Fabio Ramos, et~al.
\newblock {CuRobo}: Parallelized collision-free robot motion generation.
\newblock In \emph{IEEE International Conference on Robotics and Automation
  (ICRA)}, pages 8112--8119, 2023.

\bibitem[Tao et~al.(2025)Tao, Xiang, Shukla, Qin, Hinrichsen, Yuan, Bao, Lin,
  Liu, Chan, et~al.]{tao2024maniskill3}
Stone Tao, Fanbo Xiang, Arth Shukla, Yuzhe Qin, Xander Hinrichsen, Xiaodi Yuan,
  Chen Bao, Xinsong Lin, Yulin Liu, Tse-Kai Chan, et~al.
\newblock Demonstrating {GPU} parallelized robot simulation and rendering for
  generalizable embodied {AI} with {ManiSkill3}.
\newblock In \emph{Robotics: Science and Systems (RSS)}, 2025.

\bibitem[Todorov et~al.(2012)Todorov, Erez, and Tassa]{todorov2012mujoco}
Emanuel Todorov, Tom Erez, and Yuval Tassa.
\newblock {MuJoCo}: A physics engine for model-based control.
\newblock In \emph{IEEE/RSJ International Conference on Intelligent Robots and
  Systems (IROS)}, pages 5026--5033, 2012.

\bibitem[Wang et~al.(2026{\natexlab{a}})Wang, Jiang, Deng, Xue, Ye, Duan,
  Haghtalab, Sastry, Abbeel, and Qi]{rpg2026}
Yen-Jen Wang, Haozhe Jiang, Shuying Deng, Haoru Xue, Weirui Ye, Rocky Duan,
  Nika Haghtalab, S.~Shankar Sastry, Pieter Abbeel, and Haozhi Qi.
\newblock Reconstruct, practice, {Go Real}: Guided self-improvement for
  embodied agents.
\newblock \emph{arXiv preprint arXiv:2610.02204}, 2026{\natexlab{a}}.

\bibitem[Wang et~al.(2026{\natexlab{b}})Wang, Dang, Shi, Gao, Yin, Tam, Liu,
  Wang, Lu, Zhang, et~al.]{wang2026justrua}
Zefan Wang, Yufan Dang, Xuanwei Shi, Mingqi Gao, Cheng Yin, SikYuen Tam, Hanyu
  Liu, Yasheng Wang, Yaxi Lu, Songyang Zhang, et~al.
\newblock {JustRUA}: A controlled study of {GPT-6} {Robot-Use} agents.
\newblock \emph{arXiv preprint arXiv:2611.16009}, 2026{\natexlab{b}}.

\bibitem[{X Square Robot}(2026)]{walloss1lite2026}
{X Square Robot}.
\newblock {WALL-OSS} 1.0 {Lite} -- {RoboDojo}: Model card.
\newblock
  \url{https://huggingface.co/x-square-robot/wall-oss-1.0-lite-robodojo}, 2026.

\bibitem[Xiang et~al.(2020)Xiang, Qin, Mo, Xia, Zhu, Liu, Liu, Jiang, Yuan,
  Wang, et~al.]{xiang2020sapien}
Fanbo Xiang, Yuzhe Qin, Kaichun Mo, Yikuan Xia, Hao Zhu, Fangchen Liu, Minghua
  Liu, Hanxiao Jiang, Yifu Yuan, He Wang, et~al.
\newblock {SAPIEN: A SimulAted Part-Based Interactive ENvironment}.
\newblock In \emph{IEEE/CVF Conference on Computer Vision and Pattern
  Recognition (CVPR)}, pages 11094--11104, 2020.

\bibitem[{Xiaomi Robotics Team} et~al.(2026){Xiaomi Robotics Team}, Guo, Jin,
  Li, Li, Li, Liu, Peng, Qin, Su, et~al.]{xiaomirobotics2026xr1}
{Xiaomi Robotics Team}, Jun Guo, Piaopiao Jin, Jason Li, Peiyan Li, Yingyan Li,
  Futeng Liu, Wanli Peng, Optimus Qin, Yifei Su, et~al.
\newblock {Xiaomi-Robotics-1}: Scaling {Vision-Language-Action} models with
  over {100K} hours of real-world trajectories.
\newblock \emph{arXiv preprint arXiv:2607.15330}, 2026.

\bibitem[Yang et~al.(2024)Yang, Kang, Huang, Zhao, Xu, Feng, and
  Zhao]{yang2024depthanythingv2}
Lihe Yang, Bingyi Kang, Zilong Huang, Zhen Zhao, Xiaogang Xu, Jiashi Feng, and
  Hengshuang Zhao.
\newblock {Depth Anything V2}.
\newblock In \emph{Advances in Neural Information Processing Systems
  (NeurIPS)}, pages 21875--21911, 2024.

\bibitem[Yu et~al.(2026)Yu, Zhang, Liu, Liu, Kang, Li, Shi, Ma, Yang, Pan,
  et~al.]{walloss2026}
Ryan Yu, Pushi Zhang, Starrick Liu, Brae Liu, Miracle Kang, Shalfun Li, Lights
  Shi, Ellie Ma, Ping Yang, Chris Pan, et~al.
\newblock {Wall-OSS-0.5} technical report.
\newblock \emph{arXiv preprint arXiv:2605.30877}, 2026.

\bibitem[Zhang et~al.(2025)Zhang, Xu, Liu, Yu, Li, Gao, Fei, Yin, Wu, Jiang,
  et~al.]{zhang2024vlabench}
Shiduo Zhang, Zhe Xu, Peiju Liu, Xiaopeng Yu, Yuan Li, Qinghui Gao, Zhaoye Fei,
  Zhangyue Yin, Zuxuan Wu, Yu-Gang Jiang, et~al.
\newblock {VLABench}: A large-scale benchmark for language-conditioned robotics
  manipulation with long-horizon reasoning tasks.
\newblock In \emph{IEEE/CVF International Conference on Computer Vision
  (ICCV)}, pages 11142--11152, 2025.

\bibitem[Zhang et~al.(2026{\natexlab{a}})Zhang, Wang, Ouyang, Huang, Li, Su,
  Jin, Chai, Liang, Dou, et~al.]{gpt6astra2026}
Wenbo Zhang, Kaixuan Wang, Yutao Ouyang, Xiaoyu Huang, Liyang Li, Kailun Su,
  Weiyang Jin, Wenhao Chai, Haotian Liang, Zhiyang Dou, et~al.
\newblock An unexpected robot policy: Early evaluations of {GPT-6 Astra} on
  {RoboDojo} and beyond.
\newblock \emph{arXiv preprint arXiv:2609.24170}, 2026{\natexlab{a}}.

\bibitem[Zhang et~al.(2026{\natexlab{b}})Zhang, Zhang, Gao, Li, Liu, Nie, Zhu,
  Qiu, Yan, Liu, et~al.]{harnessvla2026}
Yixian Zhang, Huanming Zhang, Feng Gao, Xiao Li, Zhihao Liu, Yi Nie, Chunyang
  Zhu, Jiaxing Qiu, Yuchen Yan, Jiyuan Liu, et~al.
\newblock {Harness VLA}: Steering frozen {VLAs} into reliable manipulation
  primitives via memory-guided agents.
\newblock \emph{arXiv preprint arXiv:2607.08448}, 2026{\natexlab{b}}.

\bibitem[Zhou et~al.(2025)Zhou, Xu, Tie, Chen, Zhang, Chu, Zhou, and
  Sun]{liberopro2025}
Xueyang Zhou, Yangming Xu, Guiyao Tie, Yongchao Chen, Guowen Zhang, Duanfeng
  Chu, Pan Zhou, and Lichao Sun.
\newblock {LIBERO-PRO}: Towards robust and fair evaluation of
  {Vision-Language-Action} models beyond memorization.
\newblock \emph{arXiv preprint arXiv:2510.03827}, 2025.

\bibitem[Zhu et~al.(2020)Zhu, Wong, Mandlekar, Mart{\'\i}n-Mart{\'\i}n, Joshi,
  Lin, Maddukuri, Nasiriany, and Zhu]{zhu2020robosuite}
Yuke Zhu, Josiah Wong, Ajay Mandlekar, Roberto Mart{\'\i}n-Mart{\'\i}n,
  Abhishek Joshi, Kevin Lin, Abhiram Maddukuri, Soroush Nasiriany, and Yifeng
  Zhu.
\newblock robosuite: A modular simulation framework and benchmark for robot
  learning.
\newblock \emph{arXiv preprint arXiv:2009.12293}, 2020.

\bibitem[Zitkovich et~al.(2023)Zitkovich, Yu, Xu, Xu, Xiao, Xia, Wu, Wohlhart,
  Welker, Wahid, et~al.]{brohan2023rt2}
Brianna Zitkovich, Tianhe Yu, Sichun Xu, Peng Xu, Ted Xiao, Fei Xia, Jialin Wu,
  Paul Wohlhart, Stefan Welker, Ayzaan Wahid, et~al.
\newblock {RT-2}: {Vision-Language-Action} models transfer web knowledge to
  robotic control.
\newblock In \emph{Conference on Robot Learning (CoRL)}, pages 2165--2183,
  2023.

\end{thebibliography}
}

\clearpage
\appendix
\makeatletter\setlength{\@fptop}{0pt}\makeatother
\providecommand{\suppcite}[1]{\cite{#1}}

\section{Overview}
\label{supp:overview}

This appendix gives the details and full numbers behind the main paper. Every value plotted in the seven study figures is reported here with full breakdowns. Values are success rates (\%) on test episodes (\S\ref{sec:protocol}); bold marks the best unless noted. Table~\ref{tab:results} gives two VLA-alone scores: \emph{Official}, the leaderboard score, and \emph{Rerun}, the same VLA under our protocol (\S\ref{supp:leader}).

\begin{itemize}[leftmargin=*,itemsep=0pt,topsep=2pt]
    \item \S\ref{supp:protocol}: episodes, seeds, perception, and statistics.
    \item \S\ref{supp:bench}: the eight benchmarks and the VLAs used on each.
    \item \S\ref{supp:models}: the VLMs, the 15 VLA checkpoints, and the perception models.
    \item \S\ref{supp:scale}: how we count rollouts, configurations, GPU-hours, tokens, and cost.
    \item \S\ref{supp:figs}: setups and numbers behind Figures~\ref{fig:action}--\ref{fig:runtime}.
    \item \S\ref{supp:extended}: per-benchmark, per-VLA, and per-tier breakdowns; candidate pools, multi-turn rescue, unseen-seed, and local-edge tables behind \S\ref{sec:recipe_main}--\ref{sec:edge_gen}; paired qualitative rollout comparisons; SimplerEnv VA/Pure-RGB and domain-randomization transfer; and evolution ablations.
    \item \S\ref{supp:leader}: official and rerun scores of the base VLAs.
    \item \S\ref{supp:program}: example VLM-written policy code.
    \item \S\ref{supp:real_robot}: supplementary physical validation on a bimanual YAM robot.
\end{itemize}

\section{Protocol Details}
\label{supp:protocol}

\lead{Cells and episodes.} Each cell is one task under one configuration, tested on at least 20 episodes. Most studies test 20, with seeds 0--19: initial states 10--29 on LIBERO and LIBERO-PRO, and layouts 0--19 on RoboDojo.

\lead{Training and test episodes are kept apart.} Policy-code evolution runs on \emph{training episodes}: by default 10 per task, seeds 1000--1009 (initial states 30--39 on LIBERO(-PRO); the non-test \texttt{val} split on RoboMME). RoboCasa365's tier and pool studies (\S\ref{supp:fig:runtime}) use 40, seeds 1000--1039. RoboDojo uses its own training layouts, seeds $\geq 100$ (test: official seed 0, layouts 0--19): 5 for the top rows of Table~\ref{tab:results} and 20 in the tier and gate studies. Code is frozen before any test episode.

\lead{Success checks and frozen models.} Each episode is judged by the benchmark's own success check. No model is trained or fine-tuned. All VLMs, VLAs, and perception models keep fixed weights or API versions for the whole study.

\lead{Episode completion.} Episodes interrupted by transient runtime faults are rerun; any incomplete episode is counted as a failure.

\lead{Paired design.} Within a study, all arms (compared designs) share seeds. Differences are paired by task and episode.

\lead{Perception at test time.} On all eight benchmarks, the harness reads no simulator object pose and no simulator object segmentation at test time. Boxes and eroded object masks come from Grounding~DINO and SAM~2 on RGB images (\S\ref{supp:models}). Standard kinematic self-masking zeroes depth on the robot's own links (matching a real robot's kinematic self-filter) and carries no object information. Where depth is given, policy code computes 3D positions from the masks, the depth image, and camera calibration.

\lead{Two depth regimes.} Seven benchmarks provide the depth channel of calibrated RGB-D cameras, rendered by the simulator (\emph{simulator depth}; SimplerEnv in its \texttt{rgbd} observation mode); the VLA keeps its RGB input, and only the 3D scene tools read depth. In the RoboCasa365 tier and pool studies, Depth Anything~V2 fills holes in the depth image. \mbox{RoboDojo} provides no depth (\emph{Pure-RGB}): its code receives 2D detections in the top rows of Table~\ref{tab:results}, and metric depth from RGB images in the tier and pool studies, as described next. The depth study (Figure~\ref{fig:context}d) keeps the same detector masks in all three conditions and varies only the depth input (\S\ref{supp:fig:context}).

\lead{Metric depth on Pure-RGB RoboDojo.} In the tier and pool studies on RoboDojo, depth comes from Depth Anything~V2 Small, which predicts affine-invariant inverse depth $d$ from one RGB image. In every image, the harness fits \mbox{$1/Z = a\,d + b$} (RANSAC followed by Huber reweighted least squares) from four sources of known geometry.
\begin{itemize}[leftmargin=*,itemsep=0pt,topsep=2pt]
    \item \emph{Calibrated cameras.} The head camera has fixed intrinsics and a fixed pose. Each wrist camera takes its pose from the pose of its arm, so every pixel of the three views defines a ray in one metric frame.
    \item \emph{Tabletop plane.} The table is a horizontal plane of known height. The ray through a table pixel, or through the lowest pixel of an object mask, meets this plane at a metric point.
    \item \emph{Forward-kinematics points.} Forward kinematics gives the 3D positions of points on both arms, such as the gripper and the flange. They project into the head image at a known depth.
    \item \emph{Head--wrist triangulation.} A point detected in the head image and in a wrist image is triangulated from its two rays: the rays must pass within 3\,cm of each other, and the point is the midpoint of their closest points. When no matching wrist pixel is given, the tool searches along the ray of the head pixel in the wrist image and keeps the depth whose image patch matches best, with a normalized cross-correlation of at least 0.7.
\end{itemize}
The inputs are the RGB images, the camera calibration, and the robot's joint state. No step reads simulator depth or a simulator object pose.

\lead{First harness version.} The VLM sees the task instruction, initial camera frames, and one detector output before it writes: boxes on those frames and, where depth is given, 3D positions from the depth image. The initial code is checked for errors without being run and goes to eight evolution rounds, with no one-shot evolution. Figure~\ref{fig:improve}a,b and the RoboTwin~2.0 row of Table~\ref{tab:results} use this version; the other top rows of Table~\ref{tab:results} add the inputs below.

\lead{Later harness version.} Most top rows of Table~\ref{tab:results} use a later harness version with three additions: (i)~reusable cross-task \emph{skill cards} (concise helper functions and lessons distilled from training episodes of other tasks in the same benchmark); (ii)~a \emph{VLA-alone wrapper} evaluated on training episodes so evolution initializes from the base VLA whenever the first draft scores lower; and (iii)~a compact demonstration contact summary on RoboDojo tasks with demos. When candidates tie in training successes, the selection gate prefers higher training task progress, and RoboDojo's planner searches nearby grasp poses when a requested pose is unreachable.

\lead{Statistics.} Differences between compared arms are paired by task and episode, with 95\% confidence intervals estimated via a two-level paired bootstrap (resampling tasks or candidates, then episodes, 10,000 times). For multi-arm hypothesis families (such as the template study in Figure~\ref{fig:forms}, the perception and calling-mode studies, and evolution gains), we apply the Holm correction ($\alpha=0.05$); ``significant'' denotes passing Holm, and Figure~\ref{fig:online}b additionally confirms significance via McNemar's test. Reported gains are computed before rounding, so they may differ by $0.1$ from the difference of rounded rates.

\section{Benchmarks}
\label{supp:bench}

Table~\ref{supp:tab:bench} lists the eight benchmarks. Together they hold 267 tasks. They cover memory, distribution shift, contact skills, two arms, a mobile base, and common sense.

\lead{Task sets in the main results.} On RoboMME, 14 of the 16 tasks use the gripper. The other two, PatternLock and RouteStick, use a stick tool. Table~\ref{tab:results} reports all 16, and \S\ref{supp:extended} splits the row; Figures~\ref{fig:context}d, \ref{fig:online}a,c,d, \ref{fig:forms}a--c, \ref{fig:improve}a,b,d run the 14 gripper tasks. On ManiSkill3, Table~\ref{tab:results} reports all 14 tasks. Eight are core tasks: StackCube, PullCubeTool, PushT, PegInsertionSide, PlugCharger, LiftPegUpright, PokeCube, and PlaceSphere. The other 6 are extended tasks; \S\ref{supp:extended} splits the row. LIBERO-PRO applies four shifts to the 40 LIBERO tasks: object, position, language, and task. RoboCasa365 has 37 fixed-base and 13 mobile-base tasks.

\begin{table}[!htbp]
    \centering
    \caption{\textbf{Eight simulation benchmarks with 267 tasks.} Engine, robot, number of tasks, task content, and the VLAs used. ManiSkill3 has no public multi-task, language-conditioned VLA checkpoint, so it tests the harness with perception tools and motion primitives alone.}
    \label{supp:tab:bench}
    \footnotesize\sloppy\setlength{\parskip}{0pt}
    \setlength{\tabcolsep}{3.5pt}
    \begin{adjustbox}{max width=\textwidth, max totalheight=0.86\textheight}
\begin{tabularx}{\textwidth}{@{}l >{\raggedright\arraybackslash}p{2.25cm} >{\raggedright\arraybackslash}p{1.95cm} r >{\raggedright\arraybackslash}X >{\raggedright\arraybackslash}p{2.75cm}@{}}
        \toprule
        \textbf{Benchmark} & \textbf{Engine} & \textbf{Robot} & \textbf{Tasks} & \textbf{What the tasks are} & \textbf{VLA(s)} \\
        \midrule
        RoboMME & SAPIEN~\suppcite{xiang2020sapien} (ManiSkill fork) & Panda & 16 & Four suites of four: Counting, Permanence, Reference, Imitation; 14 gripper tasks & MME-VLA suite, $\pi_{0.5}$ \\
        LIBERO / LIBERO-PRO & robosuite~\suppcite{zhu2020robosuite} + MuJoCo~\suppcite{todorov2012mujoco} & Panda & 40 & Four suites of ten; LIBERO-PRO shifts object, position, language, and task & $\pi_{0.5}$ \\
        ManiSkill3 & SAPIEN & Panda & 14 & Pick, push, pull, roll, stack, and insert; 8 core and 6 extended tasks & none \\
        RoboTwin 2.0 & SAPIEN + cuRobo~\suppcite{sundaralingam2023curobo} & two-arm ALOHA AgileX & 50 & Two-arm manipulation & $\pi_{0.5}$ \\
        RoboCasa365 & robosuite + MuJoCo & mobile Panda & 50 & Kitchen tasks: 37 fixed base, 13 mobile base & $\pi_{0.5}$, RLDX-1, XR-1, Awomo-0.5I \\
        RoboDojo & Isaac Sim 5.1~\suppcite{nvidia2023isaacsim} & two-arm ARX-X5 & 42 & Five skill axes: Memory 6, Precision 8, Long-horizon 8, Open 8, Generalization 12 & $\pi_{0.5}$, DM0.5, WALL-OSS-1.0-lite, Awomo-0.5I \\
        SimplerEnv & SAPIEN & WidowX, Google robot & 6 & Bridge 4, Google robot 2 & $\pi_0$ \\
        VLABench & MuJoCo & Panda & 49 & Five tracks: in-distribution, cross-category, common sense, semantic instruction, unseen texture & $\pi_{0.5}$, $\pi_0$ \\
        \midrule
        \textbf{Total} & & & \textbf{267} & & \\
        \bottomrule
    \end{tabularx}
\end{adjustbox}
\end{table}

\section{Models}
\label{supp:models}

\lead{VLMs.} We run six frontier cloud VLMs across the main studies, plus two locally served FP8 open-weight multimodal models---\texttt{Qwen3.6-35B-A3B-FP8} (Qwen3.6-35B) and \texttt{Qwen3.6-27B-FP8} (Qwen3.6-27B)---for zero-API edge deployment (\S\ref{sec:edge_gen}). Gemini~3.8 Flash (Flash), Claude Sonnet~5.5 (Sonnet), and Claude Opus~5.5 (Opus) write policy code in most studies. For cross-family generality, Gemini~4 Argon (Argon) runs the run-time tier study (\S\ref{sec:runtime}) and GPT-6 Astra (Astra) evolves code in the RoboDojo selection-gate study (\S\ref{sec:improve}); both also write candidates for the pools and sets of \S\ref{supp:extended}. Gemini~3.5 Flash-Lite issues in-loop sub-goals at the lower end on RoboMME (\S\ref{supp:fig:action}). In Figure~\ref{fig:action}a,b, Flash uses default thinking, and Opus low and high reasoning effort.

\lead{VLAs.} We run 15 public VLA checkpoints, frozen (Table~\ref{supp:tab:vla}); six are $\pi_{0.5}$ fine-tunes, one per benchmark.

\lead{Perception models and planners.} The perception tools run four models. Grounding DINO (tiny) detects objects (default threshold 0.1; 0.15 in the scene-graph tool of Figure~\ref{supp:fig:program}; at most 20 candidates). SAM~2 gives one mask per box. CoTracker3 tracks points online. Depth Anything~V2 Small gives monocular depth in the depth study and in the tier study on RoboDojo, and fills depth holes in the tier study on RoboCasa365. Motion primitives use cuRobo~\suppcite{sundaralingam2023curobo} on RoboDojo and RoboTwin, and the ManiSkill motion planner on RoboMME and ManiSkill3. Elsewhere, end-effector targets go to the benchmark's controller or inverse kinematics.

\section{Scale Accounting}
\label{supp:scale}

Table~\ref{supp:tab:scale} reports conservative lower bounds on rollout and cell counts over non-overlapping logged batches, alongside full-study compute and token totals matching Figure~\ref{fig:overview}.

\begin{table}[!htbp]
    \centering
    \begin{minipage}[t]{0.53\textwidth}
        \centering
        \caption{\textbf{Fifteen VLA checkpoints.} Where each is used. Rows 1, 3, and 4 are from the MME-VLA suite. Row 2 alone scores 18.6 on the 16 RoboMME tasks. Row 14 is the public WALL-OSS-1.0-lite checkpoint for RoboDojo, used in the last row of Table~\ref{tab:results}. Rows 10 and 15 each count one model: $\pi_0$ has three public fine-tunes, and Awomo-0.5I its official weights for RoboCasa365 and for RoboDojo.}
        \label{supp:tab:vla}
        \footnotesize
        \setlength{\tabcolsep}{2.8pt}
        \begin{adjustbox}{max width=\linewidth}
\begin{tabular}{@{}r l l@{}}
            \toprule
            & \textbf{Checkpoint} & \textbf{Benchmark and role} \\
            \midrule
            1 & MME-VLA, memory (frame sampling) & RoboMME, VLA alone \\
            2 & $\pi_{0.5}$ RoboMME baseline & RoboMME, reference \\
            3 & MME-VLA, target-marked sub-goal & RoboMME, harness \\
            4 & MME-VLA, language-only sub-goal & RoboMME, Figure~\ref{fig:action}a \\
            5 & $\pi_{0.5}$ LIBERO & LIBERO, LIBERO-PRO \\
            6 & $\pi_{0.5}$ RoboCasa & RoboCasa365 \\
            7 & $\pi_{0.5}$ RoboTwin & RoboTwin 2.0 \\
            8 & $\pi_{0.5}$ RoboDojo & RoboDojo \\
            9 & $\pi_{0.5}$ VLABench & VLABench \\
            10 & $\pi_0$ & SimplerEnv, VLABench \\
            11 & RLDX-1 & RoboCasa365 \\
            12 & Xiaomi-Robotics-1~\suppcite{xiaomirobotics2026xr1} (here: XR-1) & RoboCasa365 \\
            13 & DM0.5 & RoboDojo \\
            14 & WALL-OSS-1.0-lite~\suppcite{walloss2026, walloss1lite2026} & RoboDojo \\
            15 & Awomo-0.5I~\suppcite{awomo2026} & RoboCasa365, RoboDojo \\
            \bottomrule
        \end{tabular}
        \end{adjustbox}
    \end{minipage}%
    \hfill
    \begin{minipage}[t]{0.44\textwidth}
        \centering
        \caption{\textbf{Over 470,000 rollouts across eight benchmarks.} Top: conservative core-study rollout count over non-overlapping logged batches (excluding 14,000+ additional extended-validation rollouts in Tables~\ref{supp:tab:simpler_va}, \ref{supp:tab:fresh_seeds}, and~\ref{supp:tab:local_qwen}). Bottom: compute and cost (\emph{Logged}: summed from job and API logs; \emph{Full study}: total including earlier batches).}
        \label{supp:tab:scale}
        \footnotesize
        \setlength{\tabcolsep}{3.0pt}
        \begin{adjustbox}{max width=\linewidth}
\begin{tabular}{@{}l r r@{}}
            \toprule
            \textbf{Benchmark} & & \textbf{Rollouts} \\
            \midrule
            RoboMME & & 259,630 \\
            LIBERO / LIBERO-PRO & & 81,082 \\
            RoboDojo & & 45,088 \\
            RoboCasa365 & & 38,860 \\
            RoboTwin 2.0 & & 37,740 \\
            ManiSkill3 & & 14,960 \\
            VLABench & & 10,820 \\
            SimplerEnv & & 1,200 \\
            \midrule
            \textbf{Total} & & \textbf{489,380} \\
            Cells in Figure~\ref{supp:fig:settings} & & 4,025 \\
            \midrule
            \textbf{Compute and cost} & \textbf{Logged} & \textbf{Full study} \\
            \midrule
            GPU-hours (H100 80\,GB) & 12,103 & 28,000+ \\
            VLM tokens & 35.8\,B & 42\,B+ \\
            VLM API cost (list price) & \$95.6K & \$120K+ \\
            Task runs (evolution and test) & -- & 55,000+ \\
            \bottomrule
        \end{tabular}
        \end{adjustbox}
    \end{minipage}
\end{table}

\begin{figure}[!htbp]
    \centering
    \includegraphics[width=\textwidth]{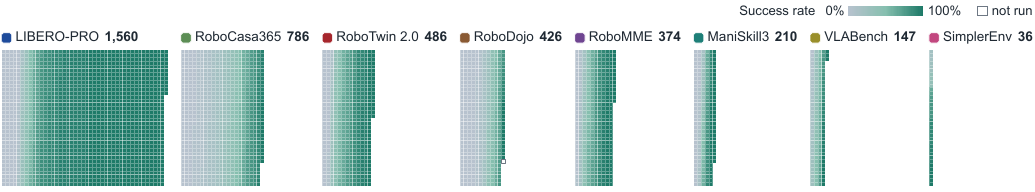}
    \caption{\textbf{4,025 evaluated configurations across eight benchmarks.} Each tile represents one cell (one task under a specific VLM, VLA, and harness configuration), grouped by benchmark and sorted by test success rate from low to high. The LIBERO-PRO block includes standard LIBERO cells.}
    \label{supp:fig:settings}
\end{figure}

\lead{Configurations.} The 4,025 cells in Figure~\ref{supp:fig:settings} span nine disjoint study batches (80,479 test episodes) across all eight benchmarks, covering evolved policy code, the template study, mid-episode object disturbances, multi-VLA extensions, the depth study, and benchmark-specific ablations. Table~\ref{supp:tab:scale} reports the full rollout total across these and all remaining studies.

\lead{Compute, tokens, and cost.} All runs span 50 days (2026-09-24 to 2026-11-12) on single-GPU H100 nodes (up to 64 concurrent jobs). Table~\ref{supp:tab:scale} pairs exact logged totals with full-study totals matching Figure~\ref{fig:overview}: 12,103 logged GPU-hours across recorded jobs (over 28,000\,h including episode-time estimates for earlier batches); 35.8\,B logged VLM tokens across 671,571 recorded calls and batch logs (over 42\,B in total); and \$95.6K in logged VLM API list-price spend with cached input pricing (over \$120K in total, excluding GPU compute). Across all evolution rounds and test evaluations, the study comprises over 55,000 single-task runs.

\section{Setups and Numbers Behind Each Figure}
\label{supp:figs}

The tables below list every value plotted in the study figures. Unless noted, a value is a success rate over all test episodes. Error bars (Figures~\ref{fig:action}, \ref{fig:context}a--c, \ref{fig:online}b, \ref{fig:where}a, \ref{fig:runtime}a,b) are 95\% bootstrap intervals over tasks (candidates in Figure~\ref{fig:context}c), then episodes (Wilson interval for one task). They describe each setting alone; the tests are paired. Violins (Figures~\ref{fig:context}d, \ref{fig:forms}d, \ref{fig:runtime}d) show the distribution of single results; their dark bars span the quartiles and their diamonds mark the means. Table~\ref{supp:tab:panels} gives the settings (cells) and rollouts of each panel; panels can share cells. Figure~\ref{fig:runtime}d's violins omit the 25 RoboCasa365 tasks with no Semi-online call; its diamonds keep them.

\subsection{Action Level of the VLM}
\label{supp:fig:action}

Panels~(a) and~(b) evaluate all 16 RoboMME tasks over 50 episodes each (800 episodes per bar; Table~\ref{supp:tab:action}). In the sub-goal arms, the VLM queries the matching MME-VLA checkpoint (rows~3--4 of Table~\ref{supp:tab:vla}) every 48 control steps ($\approx 11$ calls per episode) with either a language sub-goal or a target-marked sub-goal; at the lighter model scale, Gemini~3.5 Flash-Lite reaches 18.0\% with target-marked sub-goals. Panel~(c) evaluates three RoboDojo tasks over 50 layouts each (150 per bar), comparing direct gripper-pose outputs against calling either cuRobo or a 250-step $\pi_{0.5}$ tool. Panel~(d) evaluates LIBERO-PRO position and task shifts (320 episodes per bar), where per-episode policy code runs either with motion primitives alone or with $\pi_{0.5}$ as an additional tool.

\begin{table}[!htbp]
    \centering
    \begin{minipage}[t]{0.50\textwidth}
        \centering
        \caption{\textbf{Sub-goals carried out by VLAs beat poses that VLMs read from images (Figure~\ref{fig:action}).} Success (\%). (a--c) VLM in the loop; (d) code per episode. VLA alone: dashed lines in the figure.}
        \label{supp:tab:action}
        \footnotesize
        \setlength{\tabcolsep}{3.5pt}
        \begin{adjustbox}{max width=\linewidth}
\begin{tabular}{@{}l c c c@{}}
            \toprule
            \multicolumn{4}{@{}l}{\textit{(a) RoboMME, Flash}} \\
            Raw gripper poses & 13.1 & & \\
            Key poses + motion planner & 17.3 & & \\
            Sub-goals $\rightarrow$ VLA & 34.0 & & \\
            + target point in the image & \best{51.9} & & \\
            VLA alone (MME-VLA) & 46.5 & & \\
            \midrule
            \textit{(b) RoboMME} & Flash & Opus low & Opus high \\
            Key poses & 17.3 & 21.5 & 30.0 \\
            Sub-goals + target point & 51.9 & 50.3 & 53.9 \\
            \midrule
            \textit{(c) RoboDojo} & Flash & Sonnet & Opus \\
            VLM outputs gripper poses & 0.7 & 0.7 & 3.3 \\
            VLM calls VLA tool & 44.7 & 40.0 & 43.3 \\
            VLA alone ($\pi_{0.5}$) & \multicolumn{3}{c}{50.7} \\
            \midrule
            \textit{(d) LIBERO-PRO, pos.\ + task} & Flash & Sonnet & Opus \\
            Motion primitives only & 10.6 & 33.8 & 56.9 \\
            + VLA tool & 36.2 & 39.4 & 48.4 \\
            VLA alone ($\pi_{0.5}$) & \multicolumn{3}{c}{28.4} \\
            \bottomrule
        \end{tabular}
        \end{adjustbox}
    \end{minipage}%
    \hfill
    \begin{minipage}[t]{0.47\textwidth}
        \centering
        \caption{\textbf{Each panel of Figures~\ref{fig:action}--\ref{fig:runtime} draws on 900 to 38,680 rollouts.} Settings\,/\,rollouts per panel.}
        \label{supp:tab:panels}
        \footnotesize
        \setlength{\tabcolsep}{3.2pt}
        \begin{adjustbox}{max width=\linewidth}
\begin{tabular}{@{}l r r r r@{}}
            \toprule
            Figure & (a) & (b) & (c) & (d) \\
            \midrule
            \ref{fig:action} & 64/3,200 & 96/4,800 & 18/900 & 96/1,920 \\
            \ref{fig:context} & 32/1,600 & 32/1,600 & 384/19,200 & 558/11,160 \\
            \ref{fig:online} & 1,934/38,680 & 32/1,600 & 288/5,760 & 384/7,680 \\
            \ref{fig:forms} & \multicolumn{3}{c}{(a--c) 1,206/24,120} & 888/17,760 \\
            \ref{fig:where} & 1,616/32,320 & 701/14,020 & 942/18,840 & -- \\
            \ref{fig:improve} & 920/18,400 & 180/3,600 & 111/2,216$^{\dagger}$ & 612/12,240 \\
            \ref{fig:runtime} & 376/7,503 & 150/3,000 & 119/2,380 & 184/3,670 \\
            \bottomrule
        \end{tabular}
        \end{adjustbox}

        \vspace{2pt}
        {\scriptsize $^{\dagger}$RoboDojo block; RoboCasa365 has 300/9,000 and diagnostics 288/5,760.}
    \end{minipage}
\end{table}

In~(d), adding the VLA tool lifts unevolved Flash significantly ($+25.6$ points, 95\% CI $[+21.2,+30.0]$) and nudges Sonnet ($+5.6$, n.s.), whereas unevolved Opus already reaches 56.9\% using 3D primitives alone (vs.\ 48.4\% with the shifted VLA, $-8.4$, CI $[-14.4,-2.5]$). After eight rounds of pre-test evolution with both tools available (\S\ref{supp:fig:improve}), Opus reaches 96.2\% across the same 320 shifted episodes (155/160 position shift, 153/160 task shift).

\lead{Task-set scope across figures.} Both Figure~\ref{fig:action}a and the \emph{Sub-goals} points in Figure~\ref{fig:online}a use the same in-loop protocol: Figure~\ref{fig:action}a aggregates all 16 RoboMME tasks over 50 episodes (51.9\%), whereas Figure~\ref{fig:online}a evaluates the 14 gripper tasks over 20 episodes (54.3\% for Flash and Opus) to match the other calling modes.

\lead{Four controls on RoboMME.} Four additional controls (16 tasks $\times$ 50 episodes unless noted) validate the findings in panels~(a) and~(b):

\emph{Task notes in the prompt.} Adding the sub-goal prompt's short task notes to the key-pose prompt with Flash across four representative tasks (SwingXtimes, StopCube, PatternLock, RouteStick; 189 paired episodes) changes success by only $+0.1$ points (95\% CI $[-4.2,+4.3]$), confirming that the $+34.6$-point advantage of target-marked sub-goals stems from the action abstraction rather than prompt notes.

\emph{Reasoning effort.} With target-marked sub-goals, Flash scores 47.1\% at low reasoning effort, 51.9\% at default, and 52.4\% at high effort, confirming that default effort suffices. For Opus, high over low effort adds $+8.5$ points with key poses, $+3.6$ with sub-goals (Table~\ref{supp:tab:action}b), and $+8.1$ for unevolved closed-loop policy code (41.9\%$\to$50.0\%), whereas with only the first and current frame (Figure~\ref{fig:context}a) it stays at 32.6\%--33.3\%---confirming that visual context is a stronger lever than extra reasoning compute.

\emph{Reference sub-goals.} Oracle simulator-derived sub-goals establish an upper bound of 83.1\% for target-marked sub-goals versus 49.6\% for language-only sub-goals (Figure~\ref{supp:fig:mme_ablations}a), confirming the substantial headroom unlocked by visually grounded sub-goal conditioning.

\emph{Seeds.} The seven RoboMME arms of Figure~\ref{supp:fig:mme_ablations}a are rerun with two more policy seeds (2,400 episodes per arm). No three-seed mean differs from its one-seed value by over 2.1 points.

\subsection{Context and Perception Tools}
\label{supp:fig:context}

Panel (a) uses Flash with in-loop sub-goals on 16 RoboMME tasks (800 episodes per arm), varying only which frames the VLM sees. In panel (b), a scripted controller executes Flash's decisions to isolate decision quality (800 episodes per arm). Panel (c) evaluates initial unevolved policy candidates (three per task, 2,400 episodes per arm and VLM): showing detector outputs before coding yields a significant $+12.5$-point gain for Opus, and $+5.0$ points for Flash. In that arm, the VLM lists up to 16 object names, and the detector runs on six start frames before coding. Panel (d) evaluates the same frozen evolved policy code across three depth sources (186 cells $\times$ 20 episodes, 3,720 per source) with no test-time VLM calls.

\begin{table}[!htbp]
    \centering
    \begin{minipage}[t]{0.50\textwidth}
        \centering
        \caption{\textbf{Full video, a task note, and tool outputs shown first each help (Figure~\ref{fig:context}a--c).} Success (\%) on RoboMME.}
        \label{supp:tab:context}
        \footnotesize
        \setlength{\tabcolsep}{3.2pt}
        \begin{adjustbox}{max width=\linewidth}
\begin{tabular}{@{}l c c c c c@{}}
            \toprule
            \textit{(a) Flash, in-loop} & Count. & Perm. & Ref. & Imit. & All \\
            \midrule
            Full video & 62.0 & 44.5 & 71.5 & 29.5 & \best{51.9} \\
            First + current frame & 54.5 & 37.5 & 17.5 & 11.5 & 30.3 \\
            \midrule
            \textit{(b) Task note, in-loop} & Swing & Insert & All 16 & & \\
            \midrule
            Without note & 0 & 8 & 59.5 & & \\
            With note & 98 & 64 & \best{72.3} & & \\
            \midrule
            \textit{(c) Tools, unevolved Offline} & Flash & Opus & & & \\
            \midrule
            No perception tools & 37.3 & 53.1 & & & \\
            Detect + track at run time & 36.3 & 55.4 & & & \\
            Track only at run time & 38.4 & 59.5 & & & \\
            Tool outputs shown first & \best{42.3} & \best{65.5} & & & \\
            \bottomrule
        \end{tabular}
        \end{adjustbox}
    \end{minipage}%
    \hfill
    \begin{minipage}[t]{0.47\textwidth}
        \centering
        \caption{\textbf{Accurate depth amplifies 3D perception tools, while VLA-assisted benchmarks remain resilient under monocular depth (Figure~\ref{fig:context}d).} Offline success (\%) across 186 frozen evolved cells evaluated under three depth sources (558 settings).}
        \label{supp:tab:depth}
        \footnotesize
        \setlength{\tabcolsep}{3.0pt}
        \begin{adjustbox}{max width=\linewidth}
\begin{tabular}{@{}l c c c@{}}
            \toprule
            & \makecell{Simulator\\depth} & \makecell{Depth from\\one image} & \makecell{Depth from\\several views} \\
            \midrule
            \textbf{All} & \best{77.0} & 60.6 & 52.1 \\
            LIBERO(-PRO) & 93.4 & 82.7 & 71.0 \\
            \quad standard & 98.1 & 97.5 & 97.1 \\
            \quad position shift & 94.4 & 80.0 & 60.0 \\
            \quad task shift & 87.7 & 70.6 & 55.8 \\
            RoboTwin & 77.7 & 74.0 & 71.0 \\
            RoboMME & 77.5 & 50.0 & 42.1 \\
            ManiSkill3 & 70.0 & 32.3 & 18.5 \\
            RoboCasa365 & 33.1 & 27.5 & 27.3 \\
            \midrule
            Flash policy code & 68.5 & 59.0 & 51.9 \\
            Sonnet policy code & 78.3 & 59.0 & 51.2 \\
            Opus policy code & 84.1 & 63.6 & 53.1 \\
            \bottomrule
        \end{tabular}
        \end{adjustbox}
    \end{minipage}
\end{table}

Monocular depth (\emph{Depth Anything V2} in Figure~\ref{fig:context}d) uses metric-aligned Depth Anything~V2 Small, and multi-view stereo uses a two-view plane sweep with calibrated camera poses. All three conditions use identical RGB boxes and masks from Grounding~DINO and SAM~2, differing only in the depth map provided to the 3D scene tools (150 of 186 cells invoke them). Overall, accurate camera depth outperforms monocular depth by $+16.4$ points (95\% CI $[+11.7,+21.7]$) and multi-view stereo by $+24.9$ points ($[+17.8,+32.7]$), while monocular depth beats multi-view stereo by $+8.5$ points ($[+3.0,+14.8]$), all significant. Table~\ref{supp:tab:depth_grid} breaks down these results and depth errors by benchmark and VLM.

\lead{VLA calls buffer depth estimation noise.} Table~\ref{supp:tab:depth_strata} separates cells that invoke a VLA at run time (123 cells) from primitives-only cells (63 cells). Under monocular depth, VLA-calling code stays resilient across standard LIBERO (97.5\%), RoboTwin (74.0\%, within 3.8 points of camera depth), and RoboCasa365 (within 5.6 points), retaining 66.9\% overall ($-9.6$ vs.\ $-29.8$ points for primitives-only)---a $+20.3$-point robustness advantage ($+37.8$ under multi-view stereo) that holds for every VLM and under a static code-text split as well.

\begin{table}[!htbp]
    \centering
    \caption{\textbf{Per-benchmark depth sensitivity across three VLMs (Figure~\ref{fig:context}d).} Offline success (\%) of frozen evolved policy code with camera depth (Sim), monocular Depth Anything~V2 (One), and multi-view stereo (Sev) over 20 test episodes per task. \emph{Depth error}: mean median absolute relative error against camera depth.}
    \label{supp:tab:depth_grid}
    \footnotesize
    \setlength{\tabcolsep}{4pt}
    \begin{adjustbox}{max width=\textwidth, max totalheight=0.86\textheight}
\begin{tabular}{@{}l r ccc c ccc c ccc c cc@{}}
        \toprule
        & & \multicolumn{3}{c}{\textbf{Flash}} & & \multicolumn{3}{c}{\textbf{Sonnet}} & & \multicolumn{3}{c}{\textbf{Opus}} & & \multicolumn{2}{c}{\textbf{Depth error}} \\
        \cmidrule(lr){3-5}\cmidrule(lr){7-9}\cmidrule(lr){11-13}\cmidrule(lr){15-16}
        Slice & Tasks & Sim & One & Sev & & Sim & One & Sev & & Sim & One & Sev & & One & Sev \\
        \midrule
        LIBERO (std.) & 8 & 98.1 & 95.6 & 95.6 & & 98.1 & 98.1 & 98.8 & & 98.1 & 98.8 & 96.9 & & \multirow{3}{*}{0.067} & \multirow{3}{*}{0.024} \\
        LIBERO-PRO position & 8 & 89.4 & 86.9 & 67.5 & & 95.0 & 73.1 & 57.5 & & 98.8 & 80.0 & 55.0 & & & \\
        LIBERO-PRO task & 8 & 72.5 & 66.2 & 65.0 & & 91.9 & 80.0 & 55.0 & & 98.8 & 65.6 & 47.5 & & & \\
        RoboMME & 14 & 62.9 & 42.9 & 30.7 & & 81.4 & 44.6 & 46.1 & & 88.2 & 62.5 & 49.6 & & 0.048 & 0.026 \\
        RoboTwin & 8 & 79.4 & 74.4 & 71.2 & & 74.4 & 72.5 & 73.1 & & 79.4 & 75.0 & 68.8 & & 0.270 & 0.269 \\
        RoboCasa365 & 8 & 28.1 & 31.9 & 27.5 & & 33.8 & 28.1 & 25.6 & & 37.5 & 22.5 & 28.8 & & 0.485 & 0.315 \\
        ManiSkill3 & 8 & 53.8 & 27.5 & 21.2 & & 71.2 & 27.5 & 6.2 & & 85.0 & 41.9 & 28.1 & & 0.089 & 0.035 \\
        \bottomrule
    \end{tabular}
\end{adjustbox}
\end{table}

\begin{table}[!htbp]
    \centering
    \caption{\textbf{Policy code that calls a VLA is markedly more resilient to estimated depth than primitives-only code (Figure~\ref{fig:context}d).} Success (\%) across the 186 cells of Table~\ref{supp:tab:depth} under camera depth (Sim), monocular depth (One), and multi-view stereo (Sev), split by whether the code invokes a VLA at run time (top) or in static code text (bottom). \emph{Difference}: VLA-calling change minus primitives-only change (points), with 95\% paired bootstrap intervals.}
    \label{supp:tab:depth_strata}
    \footnotesize
    \setlength{\tabcolsep}{6pt}
    \begin{adjustbox}{max width=\textwidth, max totalheight=0.86\textheight}
\begin{tabular}{@{}l l r c c c c c@{}}
        \toprule
        Cells & Policy code & Count & Sim & One & Sev & One $-$ Sim & Sev $-$ Sim \\
        \midrule
        \textbf{All} & Calls VLA & 123 & 76.4 & 66.9 & 64.3 & $-9.6$ $[-14.5,-5.2]$ & $-12.1$ $[-16.8,-7.9]$ \\
        & Primitives only & 63 & 78.1 & 48.3 & 28.2 & $-29.8$ $[-39.7,-21.3]$ & $-49.9$ $[-62.3,-36.1]$ \\
        & \emph{Difference} & & & & & $+20.3$ $[+11.4,+30.0]$ & $+37.8$ $[+23.4,+50.3]$ \\
        \midrule
        Flash policy code & Calls VLA & 42 & 72.1 & 69.5 & 67.4 & $-2.6$ $[-7.0,+1.7]$ & $-4.8$ $[-10.2,+0.1]$ \\
        & Primitives only & 20 & 61.0 & 37.0 & 19.2 & $-24.0$ $[-39.7,-10.5]$ & $-41.8$ $[-58.3,-25.4]$ \\
        & \emph{Difference} & & & & & $+21.4$ $[+7.4,+37.8]$ & $+37.0$ $[+19.7,+54.0]$ \\
        \midrule
        Sonnet policy code & Calls VLA & 38 & 75.8 & 64.6 & 66.2 & $-11.2$ $[-18.1,-4.6]$ & $-9.6$ $[-16.9,-2.8]$ \\
        & Primitives only & 24 & 82.3 & 50.2 & 27.5 & $-32.1$ $[-47.1,-18.6]$ & $-54.8$ $[-70.8,-37.3]$ \\
        & \emph{Difference} & & & & & $+20.9$ $[+6.0,+36.9]$ & $+45.2$ $[+26.3,+62.6]$ \\
        \midrule
        Opus policy code & Calls VLA & 43 & 81.2 & 66.3 & 59.7 & $-14.9$ $[-23.2,-7.7]$ & $-21.5$ $[-30.4,-13.4]$ \\
        & Primitives only & 19 & 90.8 & 57.6 & 38.4 & $-33.2$ $[-47.8,-18.8]$ & $-52.4$ $[-70.6,-33.2]$ \\
        & \emph{Difference} & & & & & $+18.3$ $[+2.5,+33.7]$ & $+30.9$ $[+10.0,+50.6]$ \\
        \midrule
        LIBERO(-PRO) & Calls VLA & 54 & 93.5 & 85.4 & 83.1 & $-8.1$ $[-16.3,-0.8]$ & $-10.5$ $[-16.4,-4.5]$ \\
        & Primitives only & 18 & 93.1 & 74.7 & 34.7 & $-18.3$ $[-37.3,-8.9]$ & $-58.3$ $[-75.7,-22.1]$ \\
        & \emph{Difference} & & & & & $+10.2$ $[+0.7,+28.4]$ & $+47.9$ $[+12.1,+64.0]$ \\
        \midrule
        RoboMME & Calls VLA & 21 & 80.5 & 56.2 & 50.7 & $-24.3$ $[-36.8,-11.8]$ & $-29.8$ $[-43.1,-18.0]$ \\
        & Primitives only & 21 & 74.5 & 43.8 & 33.6 & $-30.7$ $[-51.3,-12.3]$ & $-41.0$ $[-63.7,-17.1]$ \\
        & \emph{Difference} & & & & & $+6.4$ $[-11.7,+25.4]$ & $+11.2$ $[-12.0,+34.3]$ \\
        \midrule
        RoboTwin & Calls VLA & 24 & 77.7 & 74.0 & 71.0 & $-3.8$ $[-11.0,+2.6]$ & $-6.7$ $[-15.0,+0.6]$ \\
        RoboCasa365 & Calls VLA & 24 & 33.1 & 27.5 & 27.3 & $-5.6$ $[-12.6,+1.2]$ & $-5.8$ $[-13.5,+2.4]$ \\
        ManiSkill3 & Primitives only & 24 & 70.0 & 32.3 & 18.5 & $-37.7$ $[-49.0,-25.8]$ & $-51.5$ $[-73.6,-31.7]$ \\
        \midrule
        \multicolumn{8}{@{}l}{\emph{Split by the code text alone}} \\
        \textbf{All} & Code contains a VLA call & 144 & 78.1 & 65.7 & 61.8 & $-12.5$ $[-17.7,-7.8]$ & $-16.3$ $[-22.3,-10.9]$ \\
        & No VLA call in the code & 42 & 73.1 & 43.1 & 18.6 & $-30.0$ $[-42.5,-18.9]$ & $-54.5$ $[-67.7,-40.5]$ \\
        & \emph{Difference} & & & & & $+17.5$ $[+5.7,+30.5]$ & $+38.2$ $[+23.6,+52.1]$ \\
        \bottomrule
    \end{tabular}
\end{adjustbox}
\end{table}

\subsection{Timing of VLM Calls}
\label{supp:fig:online}

Panel~(a) aggregates 1,934 cells (38,680 test episodes) across six benchmarks (with standard LIBERO and LIBERO-PRO plotted separately) into 146 benchmark--VLM--mode points, summarized in Table~\ref{supp:tab:online_hz}. Each shaded ellipse spans 1.3 standard deviations around the episode-weighted mean of its mode family (in-loop VLM calls; per-episode code; unevolved Offline code and VLA alone; and evolved policy code).

\lead{VLM-wait-adjusted step rate.} Panel~(a) plots $\mathrm{Hz_{eff}} = S / (S/f + T_{\mathrm{VLM}})$, where $S$ is mean control steps per episode, $f$ is the native simulator rate (16.7--25\,Hz; gray band), and $T_{\mathrm{VLM}}$ is mean VLM response wait per episode. This metric isolates latency from remote VLM calls independently of local GPU load, while full wall-clock episode times are reported below. Modes with zero test-time VLM calls run at the native rate by definition. When online step counts are not logged separately, $S$ uses the matched Offline step count, and Semi-online guard runs scale the native rate by the batch-level active motion share.

\lead{Success on shared benchmarks.} Across the five benchmarks shared by all unevolved modes (ManiSkill3, RoboCasa365, RoboDojo, RoboMME, and RoboTwin), success and test-time VLM calls per episode progress from 37.8\% at 0 calls (Offline) to 41.7\% at 0.9 calls (guard), 40.1\% at 1.2 calls (code per episode), 45.4\% at 2.3 calls (multi-turn code), and 46.1\% at 6.7 calls (tool agent). By contrast, after eight rounds of pre-test evolution (Tables~\ref{supp:tab:online_hz} and~\ref{supp:tab:tiers}, free-form code), Offline policy code surges to 77.5\% across the 186 shared non-RoboDojo cells with 0 test-time calls (77.7\% with a Semi-online guard). Table~\ref{supp:tab:online_a} reports the matched 14-task RoboMME breakdown.

\begin{table}[!htbp]
    \centering
    \begin{minipage}[t]{0.50\textwidth}
        \centering
        \caption{\textbf{Evolved Offline policy code has no VLM wait and scores far above every mode before evolution (Figure~\ref{fig:online}a).} Dots, cells, and episodes per group; range of the VLM-wait-adjusted step rate (Hz); pooled test success (\%).}
        \label{supp:tab:online_hz}
        \footnotesize
        \setlength{\tabcolsep}{2.2pt}
        \begin{adjustbox}{max width=\linewidth}
\begin{tabular}{@{}l r r r c c@{}}
            \toprule
            Group & Dots & Cells & Episodes & Hz & Succ. \\
            \midrule
            \multicolumn{6}{@{}l}{\textit{Before evolution}} \\
            Tool agent & 21 & 216 & 4,320 & 0.9--7.0 & 51.0 \\
            In-loop sub-goals & 2 & 28 & 560 & 3.0--6.1 & 54.3 \\
            Code per episode & 21 & 216 & 4,320 & 0.5--8.7 & 45.7 \\
            Multi-turn code & 21 & 216 & 4,320 & 0.6--10.3 & 50.3 \\
            Policy code + guard & 15 & 144 & 2,880 & 3.4--19.5 & 41.7 \\
            Offline policy code & 21 & 648 & 12,960 & 16.7--25 & 44.6 \\
            VLA alone & 6 & 64 & 1,280 & 16.7--25 & 47.3 \\
            \midrule
            \multicolumn{6}{@{}l}{\textit{After eight evolution rounds}} \\
            Offline policy code & 21 & 216 & 4,320 & 16.7--25 & 71.5 \\
            Policy code + Semi-online guard & 18 & 186 & 3,720 & 9.0--17.5 & 77.7 \\
            \midrule
            Total & 146 & 1,934 & 38,680 & & \\
            \bottomrule
        \end{tabular}
        \end{adjustbox}
    \end{minipage}%
    \hfill
    \begin{minipage}[t]{0.47\textwidth}
        \centering
        \caption{\textbf{Policy code with a guard needs about one VLM call per episode, while in-loop sub-goals require 11--12 calls yet trail evolved Offline code (RoboMME dots of Figure~\ref{fig:online}a).} RoboMME, 14 tasks $\times$ 20 episodes: VLM calls per episode and success (\%).}
        \label{supp:tab:online_a}
        \footnotesize
        \setlength{\tabcolsep}{3.0pt}
        \begin{adjustbox}{max width=\linewidth}
\begin{tabular}{@{}l c c c c c c@{}}
            \toprule
            & \multicolumn{2}{c}{Flash} & \multicolumn{2}{c}{Sonnet} & \multicolumn{2}{c}{Opus} \\
            \cmidrule(lr){2-3}\cmidrule(lr){4-5}\cmidrule(lr){6-7}
            Calling mode & Calls & Succ. & Calls & Succ. & Calls & Succ. \\
            \midrule
            Offline policy code & 0 & 38.6 & 0 & 46.4 & 0 & 64.5 \\
            Policy code + guard & 0.99 & 52.5 & 0.96 & 46.1 & 0.99 & 73.9 \\
            Code per episode & 1.74 & 38.9 & 1.26 & \best{55.4} & 1.51 & 67.5 \\
            Multi-turn code & 2.32 & 36.4 & 1.46 & 49.6 & 1.36 & \best{75.4} \\
            Tool agent & 6.01 & \best{58.6} & 7.42 & 50.4 & 6.58 & 72.1 \\
            In-loop sub-goals & 10.74 & 54.3 & 12.37 & 46.1 & 11.03 & 54.3 \\
            \bottomrule
        \end{tabular}
        \end{adjustbox}
    \end{minipage}
\end{table}

The six calling modes are:
\begin{itemize}[leftmargin=*,itemsep=0pt,topsep=2pt]
    \item \emph{Offline policy code}: written once per task, as condition--action rules (free-form code on LIBERO), before testing, unevolved; no call at test time. Values average three candidates per task. Round 0 in Figure~\ref{fig:improve}b is free-form code.
    \item \emph{Policy code + guard}: Offline code runs under a guard that only corrects. When it fires, the VLM answers ``continue'' or gives 1--3 corrective actions; it never rewrites the code. On RoboMME, it fires at the start of a demo-video episode, after two actions without progress, or on an empty perception result, up to 3 times per episode.
    \item \emph{Code per episode}: the VLM writes policy code once at the start of each episode, with up to 3 calls.
    \item \emph{Multi-turn code} (\emph{Multi-turn} in the figure): the VLM writes code in up to 5 turns per episode and sees the result between turns.
    \item \emph{Tool agent}: the VLM makes up to 15 decisions per episode, each a single action.
    \item \emph{In-loop sub-goals} (\emph{Sub-goals} in the figure): the VLM issues a target-marked sub-goal every 48 control steps.
\end{itemize}

\lead{Four further calling modes.} The runs of Figure~\ref{fig:online}a include four more modes before evolution, not drawn (Table~\ref{supp:tab:online_more}):
\begin{itemize}[leftmargin=*,itemsep=0pt,topsep=2pt]
    \item \emph{ReKep per episode}: as Code per episode, but the VLM fills the ReKep template (\S\ref{supp:fig:forms}); up to 3 calls.
    \item \emph{Plan per episode}: the VLM lists 1 to 20 actions at the start of the episode, and they run open-loop; up to 3 calls.
    \item \emph{PDDL with VLM predicates}: one unevolved PDDL candidate runs. The VLM states which facts about the scene (the predicates) hold at the start, after a change of the state, and when an operator makes no progress; up to 6 calls.
    \item \emph{Numbered-object decisions}: the VLM makes up to 5 decisions per episode. Each has one to three steps that name detected objects by the numbers drawn on the image.
\end{itemize}
ReKep per episode runs on the five benchmarks of Figure~\ref{fig:online}c,d, and the other three modes on RoboMME and RoboDojo (with $\pi_{0.5}$ completing remaining steps on RoboDojo once the plan ends). Compared against their Offline template counterparts, ReKep per episode improves significantly in 4 of 15 benchmark--VLM pairs ($+10.2$ to $+19.6$ points), while PDDL with VLM predicates stays within $-2.5$ to $+5.2$ points. On RoboMME, all four modes remain well below evolved Offline policy code (70.4--90.4\%, Table~\ref{supp:tab:tiers}) while consuming 14k--83k tokens and 43--327\,s per episode versus 31--35\,s for Offline policy code.

\lead{Prior harness designs under matched tools.} Table~\ref{supp:tab:prior} maps the core paradigms of prior robot-use systems to our study arms under one unified interface sharing identical perception tools, motion primitives, VLAs, and test episodes.

\begin{table}[!htbp]
    \centering
    \caption{\textbf{Prior harness paradigms mapped to controlled study arms under matched tools.} Each row names one paradigm, the study arms that instantiate it, and where results appear.}
    \label{supp:tab:prior}
    \footnotesize\sloppy\setlength{\parskip}{0pt}
    \setlength{\tabcolsep}{4.0pt}
    \begin{adjustbox}{max width=\textwidth, max totalheight=0.86\textheight}
\begin{tabularx}{\textwidth}{@{}>{\raggedright\arraybackslash}p{4.2cm} >{\raggedright\arraybackslash}X >{\raggedright\arraybackslash}p{3.5cm}@{}}
        \toprule
        \textbf{Prior paradigm} & \textbf{Arms in this study} & \textbf{Results} \\
        \midrule
        Open-loop plans and per-episode code~\suppcite{liang2023codeaspolicies, singh2023progprompt} & Plan per episode (open-loop); Code per episode; Multi-turn code; open- vs.\ closed-loop Offline code & \S\ref{sec:online}, Tables~\ref{supp:tab:online_hz}--\ref{supp:tab:online_bcd} \\
        Keypoint constraints~\suppcite{huang2024rekep} & ReKep per episode (five benchmarks); Offline ReKep template & \S\ref{sec:online}--\ref{sec:forms}, Tables~\ref{supp:tab:online_more}, \ref{supp:tab:forms} \\
        Symbolic planning~\suppcite{mcdermott1998pddl} & Offline PDDL template; PDDL with VLM predicates & \S\ref{sec:online}--\ref{sec:forms}, Tables~\ref{supp:tab:online_more}, \ref{supp:tab:forms} \\
        Behavior trees, coroutines, condition--action rules~\suppcite{colledanchise2018bt, nilsson1994teleoreactive} & Three Offline templates; explicit state-machine graph interpreter & \S\ref{sec:forms}, Table~\ref{supp:tab:forms}, Figure~\ref{supp:fig:mme_ablations}b,c \\
        VLM planners sending sub-goals to a VLA~\suppcite{shi2025hirobot, hu2026orchestrating} & In-loop language sub-goals; target-marked (visual-prompted) sub-goals & \S\ref{sec:action}, \S\ref{sec:online}, Tables~\ref{supp:tab:action}, \ref{supp:tab:online_a}, \ref{supp:tab:online_more}, Figure~\ref{fig:validation}c \\
        \bottomrule
    \end{tabularx}
\end{adjustbox}
\end{table}

Across all three VLMs, in-loop sub-goal planners average 51.5\% on RoboMME at 10.7--12.4 VLM calls per episode and 27.7\% on RoboCasa365 at 4.6--7.8 calls (47k--68k input tokens per episode; Tables~\ref{supp:tab:online_a} and~\ref{supp:tab:online_more}). By contrast, evolved Offline policy code reaches 81.2\% pooled on RoboMME and 33.1\% on the same RoboCasa365 tasks with zero test-time VLM calls (Figure~\ref{fig:validation}c).

Panel (b) uses one Flash candidate per task and arm, before evolution, on 16 RoboMME tasks with 50 episodes each (800 per arm). The two variants share one interface and differ in what they see at run time. Closed-loop policy code reads the live frames and the robot state at every step; open-loop code sees only the first frame. Closed-loop policy code gains $+14.5$ points (95\% CI $[+11.3,+17.8]$, McNemar $p<0.0001$), improving 13 of 16 tasks.

Panels~(c) and~(d) evaluate 8 tasks each on ManiSkill3, RoboCasa365, and RoboTwin, 10 on RoboDojo, and 14 on RoboMME (plus 8 LIBERO-PRO tasks across 2 shifts in panel~d; Table~\ref{supp:tab:online_bcd}). In panel~(c), a lightweight corrective guard wraps one unevolved candidate per task (firing 0.22--1.79 times per episode) and reports the direct gain over that wrapped candidate, while Holm-corrected paired tests compare against the three-candidate Offline mean (significant for Flash on RoboTwin and for Flash and Opus on RoboMME). In panel~(d), multi-turn code outperforms single-turn per-episode code in 15 of 18 benchmark--VLM pairs (384 cells, 7,680 episodes).

\begin{table}[!htbp]
    \centering
    \begin{minipage}[t]{0.50\textwidth}
        \centering
        \caption{\textbf{On RoboMME and additional benchmarks, in-loop calling modes stay below evolved Offline policy code (first harness version on RoboMME: 70.4, 82.9, 90.4; Table~\ref{supp:tab:tiers}).} All before evolution. (a) RoboMME, 14 tasks $\times$ 20 episodes: VLM calls per episode and success (\%). (b) Success (\%) on RoboDojo, 10 tasks $\times$ 20 layouts. (c) Success (\%) of ReKep per episode on three more benchmarks (8 tasks $\times$ 20 episodes). (d) RoboCasa365 in-loop sub-goals (8 tasks $\times$ 20 episodes): calls per episode and success (\%).}
        \label{supp:tab:online_more}
        \footnotesize
        \setlength{\tabcolsep}{2.8pt}
        \begin{adjustbox}{max width=\linewidth}
\begin{tabular}{@{}l c c c c c c@{}}
            \toprule
            & \multicolumn{2}{c}{Flash} & \multicolumn{2}{c}{Sonnet} & \multicolumn{2}{c}{Opus} \\
            \cmidrule(lr){2-3}\cmidrule(lr){4-5}\cmidrule(lr){6-7}
            \textit{(a) RoboMME} & Calls & Succ. & Calls & Succ. & Calls & Succ. \\
            \midrule
            ReKep per episode & 1.02 & 40.7 & 1.03 & \best{55.7} & 1.03 & \best{72.5} \\
            Plan per episode & 1.01 & 49.3 & 1.01 & 41.4 & 1.00 & 52.1 \\
            PDDL with VLM predicates & 3.56 & 42.5 & 3.74 & 48.6 & 3.67 & 59.3 \\
            Numbered-object decisions & 3.31 & \best{51.4} & 3.27 & 46.4 & 3.27 & 55.0 \\
            \midrule
            \textit{(b) RoboDojo} & \multicolumn{2}{c}{Flash} & \multicolumn{2}{c}{Sonnet} & \multicolumn{2}{c}{Opus} \\
            \midrule
            ReKep per episode & \multicolumn{2}{c}{36.0} & \multicolumn{2}{c}{26.0} & \multicolumn{2}{c}{\best{37.0}} \\
            Plan per episode & \multicolumn{2}{c}{\best{36.5}} & \multicolumn{2}{c}{\best{35.0}} & \multicolumn{2}{c}{35.0} \\
            PDDL with VLM predicates & \multicolumn{2}{c}{29.0} & \multicolumn{2}{c}{23.0} & \multicolumn{2}{c}{32.5} \\
            Numbered-object decisions & \multicolumn{2}{c}{32.0} & \multicolumn{2}{c}{25.0} & \multicolumn{2}{c}{31.0} \\
            \midrule
            \multicolumn{7}{@{}l}{\textit{(c) ReKep per episode on three more benchmarks}} \\
            ManiSkill3 & \multicolumn{2}{c}{23.1} & \multicolumn{2}{c}{46.2} & \multicolumn{2}{c}{60.0} \\
            RoboTwin & \multicolumn{2}{c}{36.2} & \multicolumn{2}{c}{23.8} & \multicolumn{2}{c}{58.1} \\
            RoboCasa365 & \multicolumn{2}{c}{26.2} & \multicolumn{2}{c}{15.6} & \multicolumn{2}{c}{31.2} \\
            \midrule
            \textit{(d) RoboCasa365 sub-goals} & Calls & Succ. & Calls & Succ. & Calls & Succ. \\
            \midrule
            In-loop sub-goals & 4.59 & 36.2 & 7.81 & 17.5 & 5.46 & 29.4 \\
            \bottomrule
        \end{tabular}
        \end{adjustbox}
    \end{minipage}%
    \hfill
    \begin{minipage}[t]{0.47\textwidth}
        \centering
        \caption{\textbf{Closed-loop feedback lifts unevolved policy code, and corrective guards and multi-turn refinement broadly add further gains (Figure~\ref{fig:online}b--d).} All before evolution. (b) Offline success (\%). (c, d) Gain (points); in (c), bold: significant.}
        \label{supp:tab:online_bcd}
        \footnotesize
        \setlength{\tabcolsep}{3.0pt}
        \begin{adjustbox}{max width=\linewidth}
\begin{tabular}{@{}l c c c c c@{}}
            \toprule
            \textit{(b) Flash} & Count. & Perm. & Ref. & Imit. & All \\
            \midrule
            Open-loop & 38.5 & 38.5 & 16.0 & 14.0 & 26.8 \\
            Closed-loop & 41.0 & 56.5 & 41.5 & 26.0 & \best{41.3} \\
            \midrule
            \textit{(c) Guard gain} & Flash & Sonnet & Opus & & \\
            \midrule
            RoboTwin & \textbf{19.4} & 12.5 & 7.5 & & \\
            RoboMME & \textbf{13.9} & 0.4 & \textbf{8.9} & & \\
            RoboCasa365 & 3.8 & 1.3 & $-$4.4 & & \\
            ManiSkill3 & 0.6 & 3.1 & 3.1 & & \\
            RoboDojo & 1.5 & 0.0 & $-$1.0 & & \\
            \midrule
            \textit{(d) Multi-turn gain} & Flash & Sonnet & Opus & & \\
            \midrule
            RoboTwin & 6.2 & 21.9 & 15.0 & & \\
            RoboMME & $-$2.5 & $-$5.7 & 7.9 & & \\
            RoboCasa365 & 3.1 & 1.9 & 11.2 & & \\
            ManiSkill3 & 13.8 & 11.9 & 7.5 & & \\
            RoboDojo & 7.5 & $-$7.0 & 1.5 & & \\
            LIBERO-PRO & 3.4 & 5.0 & 2.8 & & \\
            \bottomrule
        \end{tabular}
        \end{adjustbox}
    \end{minipage}
\end{table}

\subsection{Policy-Code Form: Templates vs.\ Free-form}
\label{supp:fig:forms}

The five templates are:
\emph{Coroutine}, a generator that yields one action per step;
\emph{Rules}, an ordered list of condition--action rules where the first rule whose condition holds fires;
\emph{Behavior tree}, a tree of predefined nodes run every step;
\emph{PDDL}, a goal and operators, from which a uniform-cost search plans and replans after each step;
\emph{ReKep}, a set of keypoints and a list of stages.

The template study evaluates unevolved Offline candidates across seven benchmark slices (3 candidates per task on ManiSkill3, RoboCasa365, and RoboTwin; 1 candidate per task on the three LIBERO slices and RoboMME) using each benchmark's standard VLA from Table~\ref{supp:tab:bench}. LIBERO slices evaluate all five templates, and the remaining slices evaluate Behavior Tree and PDDL; Tables~\ref{supp:tab:forms} and~\ref{supp:tab:forms_base} report template gains and free-form baselines. Companion runs covering Coroutine, Rules, and ReKep on the four non-LIBERO slices---plus all five templates on 10 RoboDojo tasks (free-form baselines: 27.0\% Flash, 23.7\% Sonnet, 36.5\% Opus)---are shown in gray in Table~\ref{supp:tab:forms} with their own Holm families and display the same pattern.

\begin{table}[!htbp]
    \centering
    \caption{\textbf{Structured templates help lighter VLMs (Flash), whereas free-form policy code gives stronger VLMs (Sonnet, Opus) higher flexibility and accuracy (Figure~\ref{fig:forms}).} Change in success (points) when the VLM fills a template instead of free-form code (unevolved, Offline tier). Bold: significant after Holm correction. Gray cells come from companion template runs with their own Holm families (\S\ref{supp:fig:forms}); the bottom rows and Figure~\ref{fig:forms} count the black cells. Co: coroutine; Ru: condition--action rules; BT: behavior tree; PD: PDDL; RK: ReKep.}
    \label{supp:tab:forms}
    \footnotesize
    \providecommand{\ear}[1]{\textcolor{black!55}{#1}}
    \setlength{\tabcolsep}{4.0pt}
    \begin{adjustbox}{max width=\textwidth, max totalheight=0.86\textheight}
\begin{tabular}{@{}l ccccc c ccccc c ccccc@{}}
        \toprule
        & \multicolumn{5}{c}{\textbf{Flash}} & & \multicolumn{5}{c}{\textbf{Sonnet}} & & \multicolumn{5}{c}{\textbf{Opus}} \\
        \cmidrule(lr){2-6}\cmidrule(lr){8-12}\cmidrule(lr){14-18}
        Slice & Co & Ru & BT & PD & RK & & Co & Ru & BT & PD & RK & & Co & Ru & BT & PD & RK \\
        \midrule
        LIBERO (std.) & \textbf{16.9} & \textbf{13.1} & \textbf{26.2} & \textbf{10.0} & \textbf{33.8} & & \textbf{$-$11.9} & \textbf{$-$28.8} & \textbf{$-$20.0} & $-$3.1 & \textbf{$-$14.4} & & 0.0 & 0.6 & $-$1.2 & 0.6 & 0.6 \\
        LIBERO-PRO position & 5.0 & \textbf{5.0} & 4.4 & 3.1 & \textbf{8.8} & & \textbf{$-$11.2} & \textbf{$-$12.5} & \textbf{$-$24.4} & \textbf{$-$9.4} & \textbf{$-$10.6} & & 5.0 & 3.1 & \textbf{6.2} & 2.5 & 3.1 \\
        LIBERO-PRO task & \textbf{10.0} & \textbf{6.9} & $-$2.5 & $-$0.6 & $-$3.1 & & $-$3.1 & \textbf{11.9} & \textbf{$-$7.5} & \textbf{8.1} & \textbf{$-$12.5} & & \textbf{$-$20.0} & \textbf{$-$17.5} & \textbf{$-$9.4} & \textbf{$-$21.2} & \textbf{$-$21.9} \\
        RoboMME & \ear{6.4} & \ear{3.6} & \textbf{7.5} & \textbf{7.1} & \ear{4.3} & & \ear{\textbf{6.4}} & \ear{\textbf{17.9}} & \textbf{13.6} & 1.1 & \ear{4.3} & & \ear{$-$1.4} & \ear{$-$3.9} & $-$0.4 & $-$3.2 & \ear{\textbf{$-$14.6}} \\
        RoboTwin & \ear{\textbf{13.3}} & \ear{10.4} & \textbf{19.2} & $-$1.9 & \ear{$-$4.4} & & \ear{1.9} & \ear{$-$9.8} & $-$8.1 & $-$3.8 & \ear{$-$12.3} & & \ear{$-$0.6} & \ear{$-$0.8} & 1.9 & $-$2.9 & \ear{$-$1.9} \\
        RoboCasa365 & \ear{9.0} & \ear{6.2} & \textbf{11.0} & 8.5 & \ear{0.6} & & \ear{$-$4.4} & \ear{$-$1.9} & 0.8 & $-$2.7 & \ear{$-$2.1} & & \ear{$-$1.9} & \ear{$-$0.8} & 4.2 & 2.1 & \ear{$-$0.4} \\
        ManiSkill3 & \ear{$-$0.6} & \ear{$-$7.1} & $-$9.0 & $-$4.8 & \ear{$-$6.5} & & \ear{10.4} & \ear{5.4} & 9.6 & 2.7 & \ear{0.6} & & \ear{7.9} & \ear{1.5} & $-$2.9 & 1.2 & \ear{5.4} \\
        \ear{RoboDojo} & \ear{3.2} & \ear{6.3} & \ear{$-$0.8} & \ear{2.3} & \ear{$-$4.0} & & \ear{$-$0.5} & \ear{7.5} & \ear{0.7} & \ear{1.8} & \ear{$-$5.5} & & \ear{$-$3.2} & \ear{$-$7.0} & \ear{$-$3.2} & \ear{$-$6.3} & \ear{$-$7.5} \\
        \midrule
        Signif.\ (+\,/\,$-$) & \multicolumn{5}{c}{13 / 0} & & \multicolumn{5}{c}{3 / 11} & & \multicolumn{5}{c}{1 / 5} \\
        Mean change & \multicolumn{5}{c}{$+$7.6} & & \multicolumn{5}{c}{$-$5.9} & & \multicolumn{5}{c}{$-$3.0} \\
        \bottomrule
    \end{tabular}
\end{adjustbox}
\end{table}

Figure~\ref{fig:forms}d plots per-candidate template gains over free-form code (296 candidates per VLM, 888 total, including the two RoboMME stick-tool tasks), yielding means of $+6.2$ (Flash), $-3.7$ (Sonnet), and $-1.7$ (Opus), compared with slice-averaged means of $+7.6$, $-5.9$, and $-3.0$ in panels~(a--c). Across all VLMs the median shift is 0, with wide per-task spread in both directions.

\FloatBarrier
\begin{wraptable}{r}{0.42\textwidth}
    \vspace{-10pt}
    \centering
    \caption{\textbf{Free-form code baseline of the template study.} Unevolved Offline success (\%). On RoboTwin, RoboCasa365, and ManiSkill3, values average three candidates per task (round~0 of Table~\ref{supp:tab:improve_ab} evaluates one).}
    \label{supp:tab:forms_base}
    \footnotesize
    \setlength{\tabcolsep}{3.5pt}
    \begin{adjustbox}{max width=\linewidth}
\begin{tabular}{@{}l c c c@{}}
        \toprule
        Slice & Flash & Sonnet & Opus \\
        \midrule
        LIBERO (std.) & 48.1 & 81.9 & 98.1 \\
        LIBERO-PRO position & 30.6 & 58.8 & 33.8 \\
        LIBERO-PRO task & 6.9 & 29.4 & 63.1 \\
        RoboMME & 29.3 & 52.5 & 74.6 \\
        RoboTwin & 21.0 & 33.8 & 62.7 \\
        RoboCasa365 & 17.7 & 16.2 & 24.8 \\
        ManiSkill3 & 26.9 & 41.0 & 54.6 \\
        \bottomrule
    \end{tabular}
    \end{adjustbox}
    \vspace{-6pt}
\end{wraptable}

Before evolution, strong VLMs already recognize when to delegate to an in-distribution VLA. On standard LIBERO (8 tasks), unevolved Opus sends 95\% of its control calls to the VLA and matches it within 0.6 points. Flash and Sonnet initially delegate far less (18\% and 9\%), before converging to 96.6\%--97.5\% across all 40 standard tasks after evolution (Figure~\ref{supp:fig:breakdowns}b).

\begin{figure}[!htbp]
    \centering
    \includegraphics[width=\textwidth]{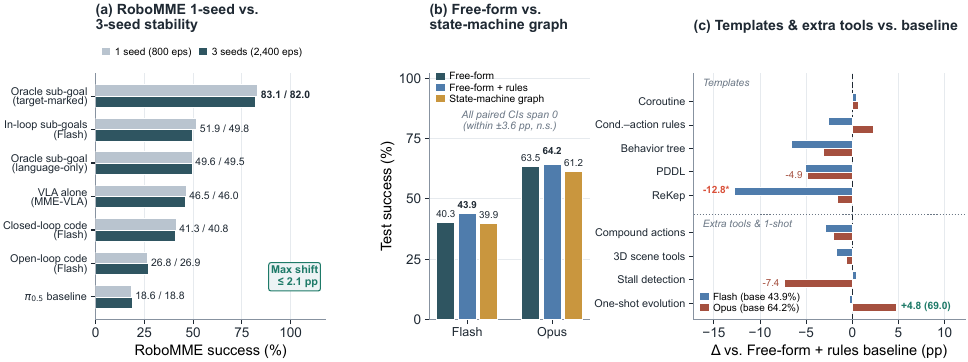}
    \caption{\textbf{RoboMME seed stability and code-representation ablations.} \textbf{(a)} One-seed (800 episodes) vs.\ three-seed (2,400 episodes) success (\%) across seven arms on all 16 RoboMME tasks: no arm shifts by more than 2.1 points, and oracle target-marked sub-goals reach 83.1\% / 82.0\%. \textbf{(b)} Unevolved Offline code across 14 RoboMME gripper tasks (2,100 episodes/arm): free-form Python matches explicit state-machine graphs within $\pm$3.6 points (all paired 95\% CIs span 0: Flash $+3.6$ $[-2.4,+9.4]$ and $-0.5$ $[-6.2,+5.5]$; Opus $+0.7$ $[-5.5,+6.1]$ and $-2.3$ $[-9.2,+5.6]$). \textbf{(c)} Point change over the \emph{Free-form + rules} baseline (43.9\% Flash, 64.2\% Opus) across five templates, three hand-crafted tools, and one-shot evolution ($^*$significant after Holm correction).}
    \label{supp:fig:mme_ablations}
\end{figure}

\lead{State-machine graphs.} On the 14 RoboMME gripper tasks (2,100 test episodes per arm and VLM; Figure~\ref{supp:fig:mme_ablations}b), we also compare free-form Python against an explicit \emph{State-machine graph} interpreter and a text-guided \emph{Free-form code + rules} arm. All three variants stay within 3.6 points for both Flash and Opus with no significant difference, confirming that strong VLMs synthesize clean state-machine logic directly in free-form Python (Figure~\ref{supp:fig:program}).

\lead{Templates and added tools on RoboMME.} A companion Offline study on the same 14 tasks compares \emph{Free-form code + rules} against all five structured templates, three added hand-crafted tools (\emph{Compound actions}, \emph{3D scene tools}, and \emph{Stall detection}), and \emph{One-shot evolution} (Figure~\ref{supp:fig:mme_ablations}c). Neither rigid templates nor extra hand-crafted tools surpass free-form code before evolution, whereas one-shot evolution already lifts Opus by $+4.8$ points to 69.0\%.

\subsection{Where Harnesses Benefit VLAs}
\label{supp:fig:where}

Table~\ref{supp:tab:where} details all 18 slices in Figure~\ref{fig:where}a using evolved Offline free-form code across six VLA families ($\pi_{0.5}$, RLDX-1, XR-1, DM0.5, MME-VLA, and $\pi_0$). The LIBERO-PRO position and task slices evaluate 32 tasks each (the remaining 8 per shift from Figure~\ref{fig:improve}a join the full 160-task LIBERO-PRO row in Table~\ref{tab:results}); on those 32-task slices, evolved Opus code invokes perception-driven motion primitives in 29 and 28 tasks (and the VLA in 9 each). On SimplerEnv Bridge (4 tasks), the harness lifts $\pi_0$ from 42.5\% to 81.2\%--97.5\%.

Table~\ref{supp:tab:vla_use} measures how often evolved Offline policy code invokes its base VLA across benchmarks. On in-distribution and multi-stage manipulation suites (standard LIBERO, RoboTwin~2.0, and fixed-base RoboCasa365), the VLA acts in $92.5\%$ to $100\%$ of solved episodes. Under LIBERO-PRO spatial/task shifts ($66.3\%$) and RoboMME ($11.3\%$), evolved code automatically shifts toward 3D perception and motion primitives where spatial shifts or memory demands exceed the base VLA's training distribution.

\lead{VLA use in the pool rows.} Table~\ref{supp:tab:vla_use_pool} reports the same step-level breakdown for the RoboCasa365 and Pure-RGB RoboDojo pool rows of Table~\ref{tab:results}. On RoboCasa365 (Awomo-0.5I), the VLA executes in $98.5\%$ of solved Offline episodes and $96.2\%$ of solved rescue episodes ($75.2\%$--$81.3\%$ of all environment steps), showing that X-Harness orchestrates rather than replaces the base VLA. On Pure-RGB RoboDojo (WALL-OSS-1.0-lite), VLA usage adapts cleanly by skill axis: high on generalization ($68$--$70\%$ of solved episodes) and long-horizon tasks ($60$--$62\%$), while shifting to 3D primitives on precision ($17$--$19\%$) and open-world tasks ($0$--$2\%$).

\begin{table}[!htbp]
    \centering
    \begin{minipage}[t]{0.51\textwidth}
        \centering
        \caption{\textbf{Harnesses help most on tasks that need what the VLA lacks (Figure~\ref{fig:where}a).} Success (\%), Offline tier, on 18 slices ($\pi_{0.5}$ unless named), sorted by VLA alone; bold: best VLM. $^\ast$: row of Table~\ref{tab:results}.}
        \label{supp:tab:where}
        \footnotesize
        \renewcommand{\arraystretch}{0.94}
        \setlength{\tabcolsep}{3.0pt}
        \begin{adjustbox}{max width=\linewidth}
\begin{tabular}{@{}l c c c c@{}}
            \toprule
            Slice (VLA) & Alone & Flash & Sonnet & Opus \\
            \midrule
            RoboCasa mobile ($\pi_{0.5}$)$^\ast$ & 6.2 & 8.8 & 20.0 & \best{28.8} \\
            RoboDojo ($\pi_{0.5}$)$^\ast$ & 9.6 & 17.1 & 18.8 & \best{28.1} \\
            RoboCasa mobile (RLDX-1) & 19.2 & 17.3 & 21.2 & \best{33.5} \\
            RoboDojo (DM0.5)$^\ast$ & 19.6 & 21.1 & 21.6 & \best{31.1} \\
            LIBERO-PRO position & 23.6 & 69.1 & 87.5 & \best{97.5} \\
            VLABench$^\ast$ & 25.5 & 36.3 & 52.1 & \best{68.3} \\
            LIBERO-PRO task & 28.9 & 77.8 & 84.5 & \best{97.8} \\
            RoboCasa fixed ($\pi_{0.5}$)$^\ast$ & 29.1 & 31.1 & 33.4 & \best{37.6} \\
            RoboCasa fixed (RLDX-1) & 36.2 & 36.5 & 38.0 & \best{45.8} \\
            RoboCasa mobile (XR-1) & 42.3 & 47.7 & 44.6 & \best{54.6} \\
            SimplerEnv Bridge ($\pi_0$) & 42.5 & 85.0 & \best{97.5} & 81.2 \\
            RoboMME (MME-VLA)$^\ast$ & 50.0 & 64.7 & 77.8 & \best{89.7} \\
            RoboCasa fixed (XR-1) & 54.2 & 53.2 & 55.5 & \best{56.6} \\
            RoboTwin 2.0 & 76.6 & 78.2 & 81.2 & \best{83.3} \\
            LIBERO-PRO object & 85.2 & 84.2 & 86.5 & \best{88.4} \\
            SimplerEnv Google ($\pi_0$) & 87.5 & 80.0 & \best{85.0} & \best{85.0} \\
            LIBERO-PRO language & 96.1 & 94.0 & 96.4 & \best{97.2} \\
            LIBERO standard$^\ast$ & 97.2 & 96.6 & \best{97.5} & 96.1 \\
            \bottomrule
        \end{tabular}
        \end{adjustbox}
    \end{minipage}%
    \hfill
    \begin{minipage}[t]{0.46\textwidth}
        \centering
        \caption{\textbf{Weaker base VLAs benefit most from a harness, and stronger VLMs consistently unlock larger gains (Figure~\ref{fig:where}b,c).} Gain over the VLA alone (points), Offline tier. In (b), the gain is the mean of the three VLMs' gains, each on the tasks that VLM ran (with DM0.5, Sonnet ran 41 of 42 tasks); (c) uses $\pi_{0.5}$ as the VLA.}
        \label{supp:tab:where_bc}
        \footnotesize
        \setlength{\tabcolsep}{3.0pt}
        \begin{adjustbox}{max width=\linewidth}
\begin{tabular}{@{}l l c c@{}}
            \toprule
            \multicolumn{2}{@{}l}{\textit{(b) Same tasks, three VLAs}} & Alone & Gain \\
            \midrule
            RoboCasa fixed & $\pi_{0.5}$ & 29.1 & 5.0 \\
            & RLDX-1 & 36.2 & 3.9 \\
            & XR-1 & 54.2 & 0.9 \\
            RoboCasa mobile & $\pi_{0.5}$ & 6.2 & 13.1 \\
            & RLDX-1 & 19.2 & 4.7 \\
            & XR-1 & 42.3 & 6.7 \\
            RoboDojo & $\pi_{0.5}$ & 9.6 & 11.7 \\
            & DM0.5 & 19.6 & 4.8 \\
            \bottomrule
        \end{tabular}
        \end{adjustbox}

        \vspace{5pt}
        \begin{adjustbox}{max width=\linewidth}
\begin{tabular}{@{}l c c c@{}}
            \toprule
            \textit{(c) Gain per VLM} & Flash & Sonnet & Opus \\
            \midrule
            VLABench & 10.8 & 26.6 & 42.8 \\
            LIBERO-PRO (4 shifts) & 23.6 & 30.7 & 36.8 \\
            RoboCasa mobile & 2.7 & 13.8 & 22.7 \\
            RoboDojo ($\pi_{0.5}$) & 7.5 & 9.2 & 18.5 \\
            RoboTwin 2.0 & 1.6 & 4.6 & 6.7 \\
            \bottomrule
        \end{tabular}
        \end{adjustbox}
    \end{minipage}
\end{table}

\begin{table}[!htbp]
    \centering
    \caption{\textbf{The share of solved episodes in which evolved policy code calls the VLA ranges from 100\% on standard LIBERO to 11.3\% on RoboMME (Figure~\ref{fig:where}a).} Offline policy code of the top rows of Table~\ref{tab:results} (RoboTwin~2.0: later harness version); a cell is one task and VLM. \emph{Can call}: cells whose code contains a VLA call. \emph{Do call}: cells whose VLA controls at least one step of a test episode. \emph{Solved with VLA}: share (\%) of solved test episodes with at least one VLA step. \emph{VLA steps}: share (\%) of all environment steps controlled by the VLA. \emph{All}: the three VLMs together. --: no step records.}
    \label{supp:tab:vla_use}
    \footnotesize
    \setlength{\tabcolsep}{3.0pt}
    \begin{adjustbox}{max width=\textwidth, max totalheight=0.86\textheight}
\begin{tabular}{@{}l r r r c cccc c cccc@{}}
        \toprule
        & \multicolumn{3}{c}{Cells} & & \multicolumn{4}{c}{Solved with VLA} & & \multicolumn{4}{c}{VLA steps} \\
        \cmidrule(lr){2-4}\cmidrule(lr){6-9}\cmidrule(lr){11-14}
        Benchmark (tasks) & Total & Can call & Do call & & Flash & Sonnet & Opus & All & & Flash & Sonnet & Opus & All \\
        \midrule
        LIBERO standard (40) & 120 & 120 & 120 & & 100.0 & 100.0 & 100.0 & 100.0 & & 98.4 & 99.4 & 99.3 & 99.0 \\
        RoboTwin 2.0 (50) & 150 & 150 & 145 & & 98.1 & 90.6 & 93.4 & 94.0 & & 88.1 & 88.3 & 83.9 & 86.8 \\
        RoboCasa fixed (37) & 111 & 109 & 108 & & 100.0 & 95.1 & 83.8 & 92.5 & & 77.8 & 76.5 & 65.6 & 73.3 \\
        LIBERO-PRO shifts (160) & 480 & 425 & 340 & & 74.6 & 66.8 & 58.7 & 66.3 & & 61.4 & 58.9 & 52.5 & 57.8 \\
        SimplerEnv (6) & 18 & 18 & 12 & & 80.0 & 46.4 & 56.6 & 60.5 & & 51.9 & 58.6 & 46.8 & 52.4 \\
        RoboCasa mobile (13) & 39 & 35 & 32 & & 95.7 & 19.2 & 46.7 & 44.7 & & 65.5 & 59.5 & 50.2 & 58.8 \\
        RoboMME (16) & 48 & 38 & 13 & & 38.6 & 1.6 & 0.0 & 11.3 & & 41.6 & 1.6 & 0.1 & 16.5 \\
        VLABench (49) & 147 & 134 & -- & & -- & -- & -- & -- & & -- & -- & -- & -- \\
        \bottomrule
    \end{tabular}
\end{adjustbox}
\end{table}

\begin{table}[!htbp]
    \centering
    \caption{\textbf{The pooled policy code calls the VLA in over 96\% of solved episodes on RoboCasa365 and adapts by skill axis on Pure-RGB RoboDojo (Table~\ref{tab:results}, pool rows).} Step records across all test episodes (20 per task and tier). \emph{Call}: tasks whose code invokes the VLA. \emph{With VLA}: solved episodes containing at least one VLA step and their share (\%). \emph{VLA steps}: share (\%) of environment steps controlled by the VLA (overall and within solved episodes).}
    \label{supp:tab:vla_use_pool}
    \footnotesize
    \setlength{\tabcolsep}{5pt}
    \begin{adjustbox}{max width=\textwidth, max totalheight=0.86\textheight}
\begin{tabular}{@{}l r r r r r r c r r r r r r@{}}
        \toprule
        & \multicolumn{6}{c}{\textbf{Offline}} & & \multicolumn{6}{c}{\textbf{Semi-online rescue}} \\
        \cmidrule(lr){2-7}\cmidrule(lr){9-14}
        Scope (tasks) & Call & Solved & \multicolumn{2}{c}{With VLA} & VLA steps & In solved & & Call & Solved & \multicolumn{2}{c}{With VLA} & VLA steps & In solved \\
        \midrule
        \multicolumn{14}{@{}l}{\emph{RoboCasa365, Awomo-0.5I (1,000 test episodes per tier)}} \\
        \textbf{All 50 tasks} & 50 & 656 & 646 & 98.5 & 81.3 & 91.4 & & 49 & 773 & 744 & 96.2 & 75.2 & 84.6 \\
        Fixed base (37) & 37 & 521 & 511 & 98.1 & 79.1 & 90.4 & & 36 & 580 & 564 & 97.2 & 76.7 & 86.9 \\
        Mobile base (13) & 13 & 135 & 135 & 100.0 & 84.5 & 93.2 & & 13 & 193 & 180 & 93.3 & 72.7 & 81.1 \\
        \midrule
        \multicolumn{14}{@{}l}{\emph{Pure-RGB RoboDojo, WALL-OSS-1.0-lite (840 test episodes per tier)}} \\
        \textbf{All 42 tasks} & 22 & 479 & 181 & 37.8 & 39.3 & 34.5 & & 23 & 524 & 197 & 37.6 & 41.0 & 33.7 \\
        Generalization (12) & 9 & 152 & 106 & 69.7 & 62.3 & 66.8 & & 9 & 167 & 113 & 67.7 & 68.5 & 62.4 \\
        Long-horizon (8) & 5 & 70 & 42 & 60.0 & 50.5 & 39.9 & & 5 & 74 & 46 & 62.2 & 48.0 & 41.8 \\
        Memory (6) & 1 & 84 & 18 & 21.4 & 12.8 & 17.4 & & 2 & 88 & 18 & 20.5 & 13.2 & 16.6 \\
        Precision (8) & 4 & 90 & 15 & 16.7 & 21.0 & 24.6 & & 4 & 97 & 18 & 18.6 & 23.3 & 23.1 \\
        Open (8) & 3 & 83 & 0 & 0.0 & 21.9 & 0.0 & & 3 & 98 & 2 & 2.0 & 22.9 & 0.1 \\
        \bottomrule
    \end{tabular}
\end{adjustbox}
\end{table}

\subsection{Policy-Code Evolution and Revision Selection}
\label{supp:fig:improve}

\lead{One evolution round.} In each round, the VLM receives a text summary of prior rounds, the current policy code and per-episode training outcomes, up to three failure reports (action logs and outcomes), and its own working notes ($\le$3,000 characters). Except on RoboDojo, training failure reports optionally include simulator diagnostics (ground-truth object poses/motions, detection matching, and an oracle-mask rerun), used strictly during evolution and never at test time; \S\ref{supp:extended} confirms strong gains even without them. Each proposed revision is evaluated on training episodes and accepted by the selection gate only if it strictly increases training successes while regressing on at most two previously solved episodes (ties break by training task progress in the later harness version). When a rejected revision solves a previously failing episode, the next round prompts the VLM to merge that fix. Eight rounds require $\sim$8 VLM calls, 350k--440k tokens, and 1.1--1.5 GPU-hours per task.

\lead{Gate rules.} On ManiSkill3 and RoboTwin (Figure~\ref{fig:improve}a), our simple default selection gate achieves 75.4\% and 81.7\% at round~8, consistently outperforming stricter zero-regression (72.0\% / 72.1\%), Bayesian (68.0\% / 78.1\%), and merge-free (75.0\% / 78.8\%) alternatives.

\lead{Panels.} Panel~(a) plots 21 benchmark--VLM evolution curves from round-0 initial code (Table~\ref{supp:tab:improve_ab}); extending evolution to rounds~16 and~32 across 143 cells confirms that gains plateau after round~8 (76.4\%, 77.6\%, and 77.1\%). Panel~(b) compares round~0$\to$8 gains per VLM across three benchmarks, and panel~(d) evaluates one-shot evolution from a single training episode. Panel~(c) isolates three protocol mechanisms: gated acceptance beats ungated edits by $+5.8$ points on RoboDojo DM0.5 (GPT-6 Astra); training-only candidate selection lifts blind RoboCasa365 tests by $+9.2$ points over VLA alone (Opus); and diagnostics-free evolution preserves most gains across 144 cells (Figure~\ref{supp:fig:ablations}b).

\begin{table}[!htbp]
    \centering
    \begin{minipage}[t]{0.50\textwidth}
        \centering
        \caption{\textbf{Four rounds bring most of the gain (Figure~\ref{fig:improve}a,b).} Offline test success (\%) after each number of evolution rounds. Round 0 is the VLM's initial free-form policy code. The last row of (a) pools 143 cells under the later version of the harness (\S\ref{supp:protocol}).}
        \label{supp:tab:improve_ab}
        \footnotesize
        \setlength{\tabcolsep}{3.0pt}
        \begin{adjustbox}{max width=\linewidth}
\begin{tabular}{@{}l r c c c c c@{}}
            \toprule
            \textit{(a) Rounds} & Cells & 0 & 1 & 2 & 4 & 8 \\
            \midrule
            LIBERO(-PRO) & 72 & 50.3 & 73.5 & 83.3 & 88.9 & 91.7 \\
            \quad standard & 24 & 77.3 & 87.5 & 91.5 & 96.2 & 96.2 \\
            \quad position shift & 24 & 40.6 & 75.0 & 89.0 & 92.1 & 94.8 \\
            \quad task shift & 24 & 32.9 & 57.9 & 69.6 & 78.3 & 84.0 \\
            RoboMME & 42 & 52.1 & 65.6 & 75.6 & 79.5 & 81.2 \\
            RoboTwin & 24 & 42.7 & 63.5 & 71.9 & 80.6 & 81.7 \\
            ManiSkill3 & 23 & 44.1 & 67.0 & 69.6 & 75.7 & 75.4 \\
            RoboCasa365 & 23 & 23.0 & 25.7 & 27.2 & 31.5 & 31.1 \\
            \midrule
            \multicolumn{7}{@{}l@{}}{\textit{Later version of the harness, after 8 / 16 / 32 rounds}} \\
            All cells & 143 & \multicolumn{5}{l}{76.4 / 77.6 / 77.1} \\
            \bottomrule
        \end{tabular}
        \end{adjustbox}

        \vspace{5pt}
        \begin{adjustbox}{max width=\linewidth}
\begin{tabular}{@{}l c c c@{}}
            \toprule
            \textit{(b) 0 $\rightarrow$ 8 rounds} & Flash & Sonnet & Opus \\
            \midrule
            RoboMME (14 tasks) & 29.3$\rightarrow$70.4 & 52.5$\rightarrow$82.9 & 74.6$\rightarrow$90.4 \\
            LIBERO-PRO task (8 tasks) & 7.5$\rightarrow$73.1 & 29.4$\rightarrow$83.1 & 61.9$\rightarrow$95.6 \\
            RoboTwin (8 tasks) & 35.6$\rightarrow$73.1 & 28.8$\rightarrow$86.2 & 63.8$\rightarrow$85.6 \\
            \bottomrule
        \end{tabular}
        \end{adjustbox}
    \end{minipage}%
    \hfill
    \begin{minipage}[t]{0.47\textwidth}
        \centering
        \caption{\textbf{The selection gate and one-shot evolution both raise test success (Figure~\ref{fig:improve}c,d).} (c) Success (\%). (d) Gain (points); bold: significant. Tier: Offline; (c) top also Semi-online.}
        \label{supp:tab:improve_cd}
        \footnotesize
        \setlength{\tabcolsep}{3.0pt}
        \begin{adjustbox}{max width=\linewidth}
\begin{tabular}{@{}l c c c@{}}
            \toprule
            \textit{(c) top: RoboDojo, gate} & Test & Change & \\
            \midrule
            Starting candidate & 30.9 & & \\
            Revisions kept with the gate & 34.9 & \up{3.9} & \\
            Every revision kept & 29.1 & \dn{1.9} & \\
            \midrule
            \textit{(c) middle: RoboCasa365, Opus} & Training & Test & \\
            \midrule
            Selected policy code & 67.1 & 48.5 & \\
            VLA alone & 39.7 & 39.3 & \\
            \midrule
            \textit{(c) bottom: Sim.\ diagnostics} & Test & Change & \\
            \midrule
            With diagnostics (144 cells) & 76.4 & & \\
            Without diagnostics & 70.8 & \dn{5.6} & \\
            \midrule
            \textit{(d) One-shot evolution gain} & Flash & Sonnet & Opus \\
            \midrule
            RoboTwin & \textbf{23.8} & 11.3 & 5.8 \\
            ManiSkill3 & 9.6 & 10.6 & \textbf{17.3} \\
            RoboMME & \textbf{12.1} & \textbf{4.3} & 0.7 \\
            RoboCasa365 & \textbf{10.4} & $-$1.7 & 2.5 \\
            LIBERO-PRO & \textbf{4.1} & \textbf{3.8} & \textbf{5.6} \\
            \bottomrule
        \end{tabular}
        \end{adjustbox}
    \end{minipage}
\end{table}

The gains of eight rounds over round~0 across all seven slices of panel~(a) are significant, e.g., $+31.3$ points on ManiSkill3, $+39.0$ on RoboTwin, and on LIBERO-PRO $+54.2$ (position shift) and $+51.0$ (task shift).

\subsection{VLM Involvement at Run Time}
\label{supp:fig:runtime}

Panels (a), (b), and (d) use Gemini~4 Argon with the Awomo-0.5I VLA on 50 RoboCasa365 cells (1,000 episodes) and 42 RoboDojo cells (840 episodes). The tiers work as follows.
\begin{itemize}[leftmargin=*,itemsep=0pt,topsep=2pt]
    \item \emph{Offline}: the evolved policy code runs with no VLM. If it does not beat the VLA-alone wrapper on the training episodes, the VLA-alone wrapper runs instead.
    \item \emph{Semi-online}: the same code runs with a guard that may take over. The guard calls the VLM when the code reports done too early, raises an error, has an action refused, stalls, or receives an empty perception result. It calls at most 3 times per episode on RoboCasa365 and 5 on RoboDojo. The VLM either declines or takes over with new code.
    \item \emph{Full-online}: from the evolved code, the VLM writes one code cell per turn and sees its result, for about ten turns per episode.
\end{itemize}
The pool bars in panel~(a) come from Table~\ref{supp:tab:pool}; the two hollow bars show the preceding milestone's Full-online score (768/1,000, 76.8\%) and the strongest of three earlier pool one-shot takeover runs (669/1,000, 66.9\%, vs.\ 651/1,000 in the preceding milestone row), providing a conservative baseline against multi-turn rescue (77.3\%). Panel~(b) bins RoboCasa365 tasks by Offline training success: below 50\% (12 tasks), 50--84\% (19 tasks), and 85\% or higher (19 tasks). Panel~(c) evaluates one-shot takeovers on LIBERO(-PRO), ManiSkill3, and RoboTwin across Flash, Sonnet, and Opus (119 cells, 2,380 episodes), replacing the running code for the remainder of the episode when triggered. Across guard triggers, Flash takes over 95\% of the time, Sonnet 35\%, and Opus 32\%, breaking more evolved episodes than they rescue (Table~\ref{supp:tab:runtime}c).

\section{Extended Results}
\label{supp:extended}

\lead{Semi-online next to Offline.} Table~\ref{supp:tab:tiers} reruns policy code in the Semi-online tier (\S\ref{sec:runtime}): eight rows of Table~\ref{tab:results} and the later-version RoboTwin~2.0 code, where Table~\ref{tab:results} reports the first version.

\begin{table}[!htbp]
    \centering
    \begin{minipage}[t]{0.46\textwidth}
        \centering
        \caption{\textbf{Runtime VLM calls help most on harder low-training-score tasks, while evolved Offline code should avoid one-shot takeovers (Figure~\ref{fig:runtime}).} (a, b) Success (\%) per run-time tier. (c) Episodes rescued / broken. (d) VLM tokens per episode, over all test episodes.}
        \label{supp:tab:runtime}
        \footnotesize
        \renewcommand{\arraystretch}{0.94}
        \setlength{\tabcolsep}{2.8pt}
        \begin{adjustbox}{max width=\linewidth}
\begin{tabular}{@{}l c c c@{}}
            \toprule
            \textit{(a) Success} & Offline & Semi-online & Full-online \\
            \midrule
            RoboCasa365 & 65.0 & 65.7 & \best{75.2} \\
            RoboDojo & 40.0 & 43.9 & \best{49.2} \\
            \textit{Calls per episode} & & & \\
            RoboCasa365 & 0 & 0.3 & 9.6 \\
            RoboDojo & 0 & 2.4 & 11.9 \\
            \midrule
            \textit{(b) RoboCasa365, training score} & & & \\
            Below 50\% (12 tasks) & 29.2 & 29.2 & \best{60.4} \\
            50--84\% (19 tasks) & 62.1 & 63.4 & \best{66.6} \\
            85\% or more (19 tasks) & 90.5 & 91.1 & \best{93.2} \\
            \midrule
            \textit{(c) Takeovers} & Flash & Sonnet & Opus \\
            Evolution kept a revision & 4 / 30 & 4 / 5 & 1 / 10 \\
            No revision kept & 7 / 2 & 1 / 3 & 11 / 2 \\
            \midrule
            \textit{(d) Tokens / episode} & Offline & Semi-online & Full-online \\
            RoboCasa365 & 0 & 33.5k & 561.5k \\
            RoboDojo & 0 & 177.4k & 748.6k \\
            \bottomrule
        \end{tabular}
        \end{adjustbox}
    \end{minipage}%
    \hfill
    \begin{minipage}[t]{0.51\textwidth}
        \centering
        \caption{\textbf{Evolved Offline policy code already captures nearly the full gain across benchmarks, with a one-shot guard adding up to +4.4 points.} Success (\%) after eight evolution rounds in the Offline tier and with a one-shot takeover guard (Semi-online tier).}
        \label{supp:tab:tiers}
        \footnotesize
        \setlength{\tabcolsep}{2.6pt}
        \begin{adjustbox}{max width=\linewidth}
\begin{tabular}{@{}l c c c c c c@{}}
            \toprule
            & \multicolumn{3}{c}{Offline} & \multicolumn{3}{c}{Semi-online (guard)} \\
            \cmidrule(lr){2-4}\cmidrule(lr){5-7}
            Benchmark & Flash & Sonnet & Opus & Flash & Sonnet & Opus \\
            \midrule
            RoboMME (16) & 64.7 & 77.8 & 89.7 & 64.1 & 77.2 & 89.7 \\
            \quad first harness version (14) & 70.4 & 82.9 & 90.4 & 72.1 & 84.6 & 90.4 \\
            LIBERO-PRO shifts & 82.2 & 89.3 & 95.4 & 81.5 & 89.2 & 95.0 \\
            LIBERO standard & 96.6 & 97.5 & 96.1 & 96.6 & 96.6 & 97.1 \\
            RoboTwin 2.0 & 78.2 & 81.2 & 83.3 & 80.1 & 80.9 & 84.4 \\
            RoboCasa fixed & 31.1 & 33.4 & 37.6 & 29.2 & 31.6 & 42.0 \\
            RoboCasa mobile & 8.8 & 20.0 & 28.8 & 6.2 & 19.6 & 30.0 \\
            RoboDojo ($\pi_{0.5}$) & 17.1 & 18.8 & 28.1 & 18.2 & 18.7 & 29.3 \\
            VLABench & 36.3 & 52.1 & 68.3 & 39.4 & 54.2 & 70.6 \\
            ManiSkill3 & 69.6 & 81.8 & 89.3 & 69.6 & 81.4 & 87.5 \\
            \bottomrule
        \end{tabular}
        \end{adjustbox}
    \end{minipage}
\end{table}

\lead{First and later harness versions on shared benchmarks.} Table~\ref{supp:tab:versions} compares the first and later harness versions across five shared task sets (\S\ref{supp:protocol}). Both versions deliver large, consistent Offline gains over the base VLA---with Opus, the first version already gains $+32.7$ points on LIBERO-PRO, $+12.0$ on RoboTwin~2.0, and $+42.1$ on RoboMME, while the later version adds further gains over the first on LIBERO-PRO ($+4.1$ with Opus) and fixed-base RoboCasa365 ($+5.7$ across VLMs; $+5.0$ over VLA alone). Table~\ref{supp:tab:later_ci} reports paired 95\% bootstrap intervals across all remaining rows of Table~\ref{tab:results}, confirming positive Opus gains on every benchmark.

\begin{table}[!htbp]
    \centering
    \caption{\textbf{Both the first and later harness versions deliver large gains over the base VLA across shared benchmarks.} Offline success (\%) after eight evolution rounds, with paired differences (points) and 95\% bootstrap intervals.}
    \label{supp:tab:versions}
    \footnotesize
    \setlength{\tabcolsep}{5pt}
    \begin{adjustbox}{max width=\textwidth, max totalheight=0.86\textheight}
\begin{tabular}{@{}l l c c c c c c@{}}
        \toprule
        Benchmark (tasks) & VLM & VLA alone & First & Later & First $-$ VLA alone & Later $-$ VLA alone & Later $-$ First \\
        \midrule
        LIBERO-PRO shifts (160) & Flash & 58.6 & 82.2 & 82.2 & $+23.6$ $[+17.0,+30.4]$ & $+23.6$ $[+17.3,+30.1]$ & $0.0$ $[-3.7,+3.8]$ \\
        & Sonnet & 58.6 & 90.8 & 89.3 & $+32.2$ $[+25.6,+39.1]$ & $+30.7$ $[+24.0,+37.5]$ & $-1.5$ $[-4.7,+1.7]$ \\
        & Opus & 58.6 & 91.3 & 95.4 & $+32.7$ $[+26.2,+39.2]$ & $+36.8$ $[+29.9,+43.9]$ & $+4.1$ $[+1.2,+7.3]$ \\
        & All & 58.6 & 88.1 & 89.0 & $+29.5$ $[+23.2,+36.0]$ & $+30.3$ $[+24.2,+36.8]$ & $+0.9$ $[-1.1,+2.8]$ \\
        \midrule
        LIBERO standard (40) & Flash & 97.2 & 94.5 & 96.6 & $-2.8$ $[-8.5,+1.0]$ & $-0.6$ $[-2.5,+1.4]$ & $+2.1$ $[-1.6,+7.7]$ \\
        & Sonnet & 97.2 & 96.5 & 97.5 & $-0.8$ $[-3.7,+2.5]$ & $+0.2$ $[-1.6,+2.4]$ & $+1.0$ $[-1.3,+3.4]$ \\
        & Opus & 97.2 & 97.2 & 96.1 & $0.0$ $[-1.6,+1.6]$ & $-1.1$ $[-3.1,+0.8]$ & $-1.1$ $[-3.1,+0.7]$ \\
        & All & 97.2 & 96.1 & 96.8 & $-1.2$ $[-3.8,+1.2]$ & $-0.5$ $[-1.8,+1.0]$ & $+0.7$ $[-1.1,+2.9]$ \\
        \midrule
        RoboTwin 2.0 (50) & Flash & 76.6 & 75.9 & 78.2 & $-0.7$ $[-7.1,+5.6]$ & $+1.6$ $[-2.8,+6.1]$ & $+2.3$ $[-3.3,+8.1]$ \\
        & Sonnet & 76.6 & 83.4 & 81.2 & $+6.8$ $[+0.1,+13.6]$ & $+4.6$ $[-0.8,+10.5]$ & $-2.2$ $[-7.3,+3.0]$ \\
        & Opus & 76.6 & 88.6 & 83.3 & $+12.0$ $[+6.7,+17.6]$ & $+6.7$ $[+1.4,+12.4]$ & $-5.3$ $[-9.7,-1.3]$ \\
        & All & 76.6 & 82.6 & 80.9 & $+6.0$ $[+1.1,+11.3]$ & $+4.3$ $[+0.1,+8.9]$ & $-1.7$ $[-4.6,+1.1]$ \\
        \midrule
        RoboMME (14) & Flash & 48.2 & 70.4 & 62.9 & $+22.1$ $[+0.4,+43.9]$ & $+14.6$ $[-3.2,+32.9]$ & $-7.5$ $[-22.9,+6.4]$ \\
        & Sonnet & 48.2 & 82.9 & 81.4 & $+34.6$ $[+15.0,+53.2]$ & $+33.2$ $[+14.6,+51.1]$ & $-1.4$ $[-8.9,+6.4]$ \\
        & Opus & 48.2 & 90.4 & 88.2 & $+42.1$ $[+25.0,+58.2]$ & $+40.0$ $[+23.9,+55.4]$ & $-2.1$ $[-8.9,+3.9]$ \\
        & All & 48.2 & 81.2 & 77.5 & $+33.0$ $[+14.9,+50.6]$ & $+29.3$ $[+13.3,+44.6]$ & $-3.7$ $[-10.6,+2.7]$ \\
        \midrule
        RoboCasa fixed (37) & Flash & 29.1 & 26.8 & 31.1 & $-2.3$ $[-9.1,+3.6]$ & $+2.0$ $[-2.4,+6.5]$ & $+4.3$ $[-1.8,+11.4]$ \\
        & Sonnet & 29.1 & 26.5 & 33.4 & $-2.6$ $[-8.9,+3.8]$ & $+4.3$ $[-0.9,+9.5]$ & $+6.9$ $[+0.5,+13.5]$ \\
        & Opus & 29.1 & 31.6 & 37.6 & $+2.6$ $[-4.6,+9.9]$ & $+8.5$ $[+1.5,+15.7]$ & $+5.9$ $[-0.3,+12.4]$ \\
        & All & 29.1 & 28.3 & 34.0 & $-0.8$ $[-5.6,+4.1]$ & $+5.0$ $[+0.5,+9.4]$ & $+5.7$ $[+2.3,+9.4]$ \\
        \bottomrule
    \end{tabular}
\end{adjustbox}
\end{table}

\begin{table}[!htbp]
    \centering
    \caption{\textbf{Paired 95\% bootstrap intervals confirm consistent Offline gains across the remaining benchmarks of Table~\ref{tab:results}.} Offline success (\%) after eight evolution rounds. \emph{Pooled}: over all test episodes. \emph{Five-axis mean}: macro average over the five RoboDojo skill axes. On RoboDojo DM0.5, Sonnet covers 41 of 42 tasks (its VLA-alone entry reflects the matched subset). Paired differences are computed before rounding.}
    \label{supp:tab:later_ci}
    \footnotesize
    \setlength{\tabcolsep}{5pt}
    \begin{adjustbox}{max width=\textwidth, max totalheight=0.86\textheight}
\begin{tabular}{@{}l l c c c c c c c@{}}
        \toprule
        & & \multicolumn{3}{c}{\textbf{Pooled}} & & \multicolumn{3}{c}{\textbf{Five-axis mean}} \\
        \cmidrule(lr){3-5}\cmidrule(lr){7-9}
        Benchmark (tasks) & VLM & VLA alone & Later & Later $-$ VLA alone & & VLA alone & Later & Later $-$ VLA alone \\
        \midrule
        RoboMME (16) & Flash & 50.0 & 64.7 & $+14.7$ $[-2.2,+31.6]$ & & & & \\
        & Sonnet & 50.0 & 77.8 & $+27.8$ $[+7.2,+46.9]$ & & & & \\
        & Opus & 50.0 & 89.7 & $+39.7$ $[+25.9,+53.4]$ & & & & \\
        & All & 50.0 & 77.4 & $+27.4$ $[+12.6,+42.1]$ & & & & \\
        \midrule
        RoboCasa mobile (13) & Flash & 6.2 & 8.8 & $+2.7$ $[-2.7,+8.5]$ & & & & \\
        & Sonnet & 6.2 & 20.0 & $+13.8$ $[-0.8,+31.5]$ & & & & \\
        & Opus & 6.2 & 28.8 & $+22.7$ $[+9.2,+38.1]$ & & & & \\
        & All & 6.2 & 19.2 & $+13.1$ $[+4.1,+23.7]$ & & & & \\
        \midrule
        RoboDojo, $\pi_{0.5}$ (42) & Flash & 9.6 & 17.1 & $+7.5$ $[+0.4,+16.0]$ & & 8.7 & 17.3 & $+8.5$ $[+1.2,+17.0]$ \\
        & Sonnet & 9.6 & 18.8 & $+9.2$ $[+0.6,+19.2]$ & & 8.7 & 19.4 & $+10.7$ $[+1.7,+20.8]$ \\
        & Opus & 9.6 & 28.1 & $+18.5$ $[+8.5,+29.0]$ & & 8.7 & 29.1 & $+20.4$ $[+10.0,+31.4]$ \\
        & All & 9.6 & 21.3 & $+11.7$ $[+4.0,+20.4]$ & & 8.7 & 21.9 & $+13.2$ $[+5.1,+22.2]$ \\
        \midrule
        RoboDojo, DM0.5 (42) & Flash & 19.6 & 21.1 & $+1.4$ $[-6.7,+9.5]$ & & 20.5 & 21.7 & $+1.2$ $[-7.6,+9.7]$ \\
        & Sonnet & 20.1 & 21.6 & $+1.5$ $[-4.2,+7.9]$ & & 20.5 & 22.6 & $+2.1$ $[-3.8,+8.8]$ \\
        & Opus & 19.6 & 31.1 & $+11.4$ $[+4.2,+19.6]$ & & 20.5 & 32.8 & $+12.3$ $[+4.7,+20.8]$ \\
        & All & 19.8 & 24.6 & $+4.8$ $[-1.1,+11.4]$ & & 20.5 & 25.7 & $+5.2$ $[-1.1,+12.0]$ \\
        \midrule
        VLABench (49) & Flash & 25.5 & 36.3 & $+10.8$ $[+3.0,+19.2]$ & & & & \\
        & Sonnet & 25.5 & 52.1 & $+26.6$ $[+17.7,+35.6]$ & & & & \\
        & Opus & 25.5 & 68.3 & $+42.8$ $[+34.0,+51.4]$ & & & & \\
        & All & 25.5 & 52.2 & $+26.7$ $[+19.6,+33.9]$ & & & & \\
        \midrule
        SimplerEnv (6) & Flash & 57.5 & 83.3 & $+25.8$ $[-4.2,+56.7]$ & & & & \\
        & Sonnet & 57.5 & 93.3 & $+35.8$ $[+5.0,+66.7]$ & & & & \\
        & Opus & 57.5 & 82.5 & $+25.0$ $[-3.3,+56.7]$ & & & & \\
        & All & 57.5 & 86.4 & $+28.9$ $[+0.3,+58.9]$ & & & & \\
        \bottomrule
    \end{tabular}
\end{adjustbox}
\end{table}

\begin{figure}[!htbp]
    \centering
    \includegraphics[width=\textwidth]{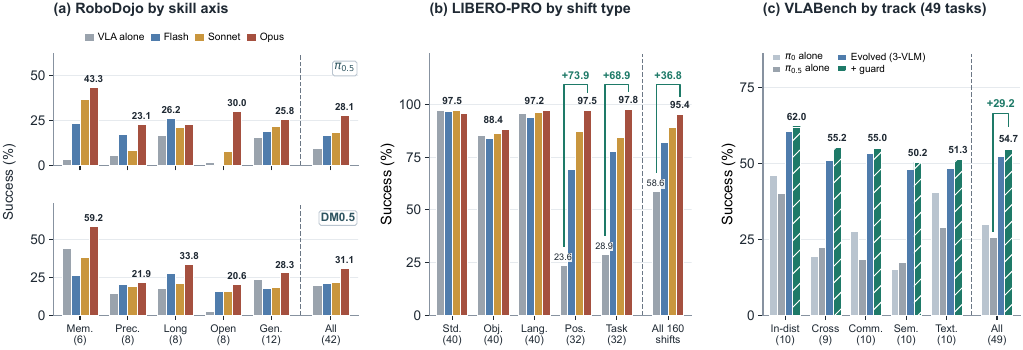}
    \caption{\textbf{Extended breakdowns across RoboDojo skill axes, LIBERO-PRO shift types, and VLABench tracks (Offline tier after eight evolution rounds).} \textbf{(a)} RoboDojo five skill axes + All (42 tasks) for $\pi_{0.5}$ (top) and DM0.5 (bottom; Sonnet covers 7 of 8 Open tasks, 41 of 42 overall). \textbf{(b)} LIBERO-PRO shift breakdown (Position and Task rows cover 32 tasks; \emph{All 160 shifts} adds the 8 tasks per shift of Figure~\ref{fig:improve}a). \textbf{(c)} VLABench across five tracks (49 tasks, 20 episodes/task), comparing $\pi_0$ alone, $\pi_{0.5}$ alone, 3-VLM evolved Offline code, and Semi-online \emph{+ guard}.}
    \label{supp:fig:breakdowns}
\end{figure}

\lead{RoboDojo by skill axis.} Figure~\ref{supp:fig:breakdowns}a splits RoboDojo across its five skill axes for both $\pi_{0.5}$ and DM0.5. With $\pi_{0.5}$ (pooled over three VLMs), the harness raises memory from 3.3\% to 34.4\% (43.3\% Opus), precision from 5.6\% to 16.5\%, open tasks from 1.9\% to 12.7\% (30.0\% Opus), long-horizon from 16.9\% to 23.5\%, and generalization from 15.8\% to 22.4\%. On DM0.5, which starts stronger on memory (44.2\%), Opus adds $+15.0$ points (59.2\%) and lifts all five axes.

\lead{LIBERO-PRO by shift type.} Figure~\ref{supp:fig:breakdowns}b splits LIBERO-PRO by shift type. With Opus, X-Harness preserves and lifts high base-VLA accuracy under object (88.4\%, $+3.2$) and language (97.2\%, $+1.1$) shifts while delivering large gains under position (23.6\%$\to$97.5\%, $+73.9$) and task (28.9\%$\to$97.8\%, $+68.9$) shifts.

\lead{VLABench by track.} Figure~\ref{supp:fig:breakdowns}c splits VLABench across its five tracks (49 tasks). Evolved Offline policy code (0 test-time VLM calls), pooled over three VLMs, lifts $\pi_{0.5}$ by 19 to 35 points on every track, with the largest gains on common sense ($+34.7$) and semantic instruction ($+30.5$), and adding a guard raises all-track success further to 54.7\%. As a second reference, $\pi_0$ alone scores 29.9\% versus 25.5\% for $\pi_{0.5}$.

\lead{SimplerEnv visual-matching, variant aggregation, and Pure-RGB transfer.} Table~\ref{supp:tab:simpler} evaluates the six rasterized non-drawer SimplerEnv visual-matching tasks using public $\pi_0$ checkpoints. On WidowX Bridge, Offline evolved policy code more than doubles base VLA success (42.5\%$\to$87.9\%, including 100.0\% on carrot on plate and 96.7\% on block stacking), while preserving strong performance on Google robot where $\pi_0$ alone already approaches ceiling (83.3\% evolved, reaching 85.0\% with Sonnet and Opus).

To evaluate zero-shot transfer under broader visual shifts, Table~\ref{supp:tab:simpler_va} runs the same frozen visual-matching Offline code unchanged across all 21 official Google-robot \emph{Variant Aggregation (VA)} conditions (1,425 episodes per arm) and under monocular \emph{Pure-RGB} depth (Depth Anything~V2 Small). Frozen Offline code matches the specialist $\pi_0$ VLA across all 1,425 VA episodes (85.2\% vs.\ 86.4\% on coke can; 70.4\% vs.\ 70.8\% on move near; both 95\% CIs spanning zero) with zero test-time VLM calls. Under Pure-RGB, it reaches 85.8\% on Google robot (95.0\% on coke can) and outperforms $\pi_0$ on Bridge (48.3\% vs.\ 42.5\%, with $+45.0$ points on carrot on plate).

\begin{table}[!htbp]
    \centering
    \begin{minipage}[t]{0.47\textwidth}
        \centering
        \caption{\textbf{On SimplerEnv, evolved policy code adds most where the base VLA falls short.} Success (\%). \emph{Alone}: $\pi_0$ on the 20 test episodes. \emph{Evolved}: Offline policy code after eight rounds, pooled over three VLMs (60 episodes). \emph{Official grid}: $\pi_0$ alone on all 636 official episodes, next to the published score of the checkpoint.}
        \label{supp:tab:simpler}
        \footnotesize
        \setlength{\tabcolsep}{3.0pt}
        \begin{adjustbox}{max width=\linewidth}
\begin{tabular}{@{}l c c c c@{}}
            \toprule
            & \multicolumn{2}{c}{Test episodes} & \multicolumn{2}{c}{Official grid} \\
            \cmidrule(lr){2-3}\cmidrule(lr){4-5}
            Task & Alone & Evolved & Rerun & Published \\
            \midrule
            \multicolumn{5}{@{}l}{\textit{Bridge (WidowX)}} \\
            Spoon on towel & 35.0 & \best{68.3} & 41.7 & 45.8 \\
            Carrot on plate & 25.0 & \best{100.0} & 25.0 & 36.1 \\
            Green block on yellow block & 20.0 & \best{96.7} & 20.8 & 26.4 \\
            Eggplant in basket & \best{90.0} & 86.7 & 87.5 & 81.9 \\
            \midrule
            \multicolumn{5}{@{}l}{\textit{Google robot}} \\
            Pick coke can & \best{95.0} & 93.3 & 95.7 & 97.9 \\
            Move near & \best{80.0} & 73.3 & 77.9 & 78.7 \\
            \bottomrule
        \end{tabular}
        \end{adjustbox}
    \end{minipage}%
    \hfill
    \begin{minipage}[t]{0.50\textwidth}
        \centering
        \caption{\textbf{Zero-shot transfer of frozen visual-matching SimplerEnv policy code to official Variant Aggregation (VA) and monocular Pure-RGB.} Top: full official VA grids (1,425 episodes per arm; $\Delta$ vs.\ Alone with 95\% layout-bootstrap CI). Bottom: Pure-RGB (Depth Anything~V2 Small, 60 episodes/task) vs.\ camera-depth Evolved and Alone (bold: best per suite).}
        \label{supp:tab:simpler_va}
        \footnotesize
        \setlength{\tabcolsep}{2.2pt}
        \begin{adjustbox}{max width=\linewidth}
\begin{tabular}{@{}l c c c c@{}}
            \toprule
            \multicolumn{5}{@{}l}{\textit{(a) Official Variant Aggregation (0 VLM calls)}} \\
            Task ($n$) & Alone & \makecell{Flash / Sonnet\\/ Opus} & Evol. & $\Delta$ [95\% CI] \\
            \midrule
            Coke can (825) & 86.4 & 85.3/85.5/84.8 & 85.2 & $-1.2$ $[-2.9,+0.5]$ \\
            Move near (600) & 70.8 & 70.3/70.7/70.2 & 70.4 & $-0.4$ $[-3.6,+2.9]$ \\
            \midrule
            \multicolumn{5}{@{}l}{\textit{(b) Monocular Pure-RGB (Depth Anything~V2 Small)}} \\
            Suite / Task & Alone & Pure-RGB & Cam.\ depth & \\
            \midrule
            Google robot (2) & \best{87.5} & 85.8 & 83.3 & \\
            \quad Coke / Move near & 95.0/80.0 & 95.0/76.7 & 93.3/73.3 & \\
            Bridge (4) & 42.5 & 48.3 & \best{87.9} & \\
            \quad Carrot / Eggplant & 25.0/90.0 & 70.0/80.0 & 100.0/86.7 & \\
            \quad Spoon / Stack & 35.0/20.0 & 23.3/20.0 & 68.3/96.7 & \\
            \bottomrule
        \end{tabular}
        \end{adjustbox}
    \end{minipage}
\end{table}

\lead{RoboCasa365 with three VLAs.} Table~\ref{supp:tab:where} details both fixed-base (37 tasks) and mobile-base (13 tasks) RoboCasa365 splits across $\pi_{0.5}$, RLDX-1, and XR-1; on the mobile base, Opus adds over 12 points to every VLA.

\lead{ManiSkill3 and RoboMME subgroup breakdowns.} On ManiSkill3 (Table~\ref{tab:results}), evolved Offline code reaches 53.8\% / 71.2\% / 85.0\% (Flash / Sonnet / Opus) on the 8 core tasks and 90.8\% / 95.8\% / 95.0\% on the 6 extended tasks. On RoboMME, it reaches 62.9\% / 81.4\% / 88.2\% on the 14 gripper tasks and 77.5\% / 52.5\% / 100.0\% on the two stick-tool tasks (Opus solving 40/40).

\begin{figure}[!htbp]
    \centering
    \includegraphics[width=\textwidth]{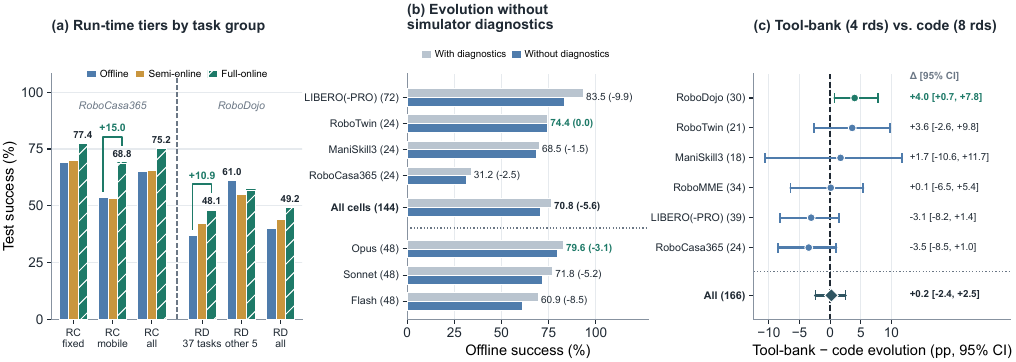}
    \caption{\textbf{Group-level tier progression, diagnostics-free evolution, and tool-bank parity.} \textbf{(a)} Run-time tier success (\%) with Gemini~4 Argon and Awomo-0.5I by task group: RoboCasa365 fixed (510/519/573 of 740), mobile (140/138/179 of 260), all (650/657/752 of 1,000); RoboDojo 37 tasks (275/314/356 of 740), remaining 5 tasks already solved at $\ge 85\%$ by DM0.5 alone (61/55/57 of 100), all (336/369/413 of 840). \textbf{(b)} Eight-round Offline success without vs.\ with simulator diagnostics across 144 cells (2,880 test episodes/arm; label shows score without diagnostics and change). \textbf{(c)} Four-round tool-bank evolution around frozen policy code minus eight-round full policy-code evolution across 166 non-saturated cells (95\% paired bootstrap CIs; RoboDojo $+4.0$ strictly above zero).}
    \label{supp:fig:ablations}
\end{figure}

\lead{Three run-time tiers by group.} Figure~\ref{supp:fig:ablations}a splits the \S\ref{sec:runtime} tier study (Gemini~4 Argon + Awomo-0.5I) by task group. Full-online adds $+102$ net solved episodes on RoboCasa365 (improving 30 of 50 tasks; $+15.0$ points on mobile base) and $+77$ on RoboDojo (improving 20 of 42 tasks; $+10.9$ points on the 37 harder tasks).

\lead{Candidate pools and multi-turn rescue stage.} The \emph{pool} rows of Table~\ref{tab:results} scale X-Harness with stronger base policies (Awomo-0.5I and WALL-OSS-1.0-lite). Inspecting failed training episodes on Awomo-0.5I revealed sub-millimeter arm tracking (0.8\,mm at the 90th percentile on RoboDojo), pointing to grasp verification and post-move stall detection as the main leverage points for shared 3D geometry, memory, and verification tools. Per task, we select the highest-scoring training candidate from a three-VLM pool (Opus, GPT-6 Astra, and Gemini~4 Argon); Table~\ref{supp:tab:pool} tracks each architectural milestone.

\lead{Multi-turn rescue matches Full-online at 42\% lower token cost.} Instead of querying the VLM from step~1, Semi-online multi-turn rescue lets frozen Offline policy code run first and invokes the VLM (up to 24 turns) only on physical failure triggers or brief progress checks. On RoboCasa365 (Awomo-0.5I), this lifts success to 77.3\% ($+12.9$ over official VLA alone; $+8.0$ on fixed-base and $+22.3$ on mobile-base tasks over Offline), with 51.4\% of test episodes needing zero rescue turns. Compared with Full-online (76.8\%), Semi-online rescue matches accuracy while cutting VLM usage by 42\% (5.3 vs.\ 8.6 turns; 207.3k vs.\ 356.9k tokens/episode), and immediate physical triggers outperform VLM-vetoed triggers (Table~\ref{supp:tab:pool}).

\lead{WALL-OSS-1.0-lite on Pure-RGB RoboDojo and broader pool scaling.} Applying the same three-VLM pool and Semi-online rescue stage to WALL-OSS-1.0-lite lifts Offline success to 57.0\% and Semi-online rescue to 62.2\% five-axis macro mean ($+27.2$ over the 35.0\% official baseline; 62.4\% pooled), scaling cleanly from 6-turn to 24-turn rescue (Table~\ref{supp:tab:pool}). Expanding an Opus-led pool with GPT-6 Astra and Gemini~4 Argon candidates (selected strictly on training episodes) likewise yields steady Offline gains on other VLAs: 64.3\%$\to$67.3\% on 50 RoboCasa365 XR-1 tasks, and 38.4\%$\to$41.9\% (with Astra) and 45.0\% (with Argon) on 42 RoboDojo DM0.5 tasks.

\lead{Task-selective Semi-online rescue.} Because Offline policy code already solves easier tasks reliably, Semi-online rescue can be enabled selectively on harder tasks using training criteria alone (Table~\ref{supp:tab:selective_rescue}). Running Offline on high-confidence tasks retains 75.8\% on RoboCasa365 ($+10.2$ over Offline) and 61.5\% five-axis mean on RoboDojo ($+4.5$ over Offline), staying within 0.7--1.5 points of all-task rescue while saving substantial VLM compute.

\lead{Unseen test seeds and local edge deployment.} Figure~\ref{fig:validation} and Tables~\ref{supp:tab:local_qwen} and~\ref{supp:tab:fresh_seeds} provide the complete visual and statistical breakdown behind \S\ref{sec:edge_gen} across 2,080 fresh unseen test episodes per arm and 1,140 zero-API local Qwen3.6 episodes.

\begin{table}[!htbp]
    \centering
    \begin{minipage}[t]{0.49\textwidth}
        \centering
        \caption{\textbf{Architectural milestones for candidate pools and multi-turn rescue on RoboCasa365 and RoboDojo.} Solved test episodes per tier (--: tier not run). On RoboCasa365 (50 tasks, 1,000 episodes), combining a three-VLM pool with immediate physical rescue triggers matches Full-online at 42\% fewer tokens (Figure~\ref{fig:runtime}a's hollow takeover bar shows 66.9\%, the strongest of three earlier pool takeover variants spanning 651--669/1,000); on Pure-RGB RoboDojo (42 tasks, 840 episodes), expanding the rescue budget from 6 to 24 turns raises WALL-OSS-1.0-lite to 524/840 (62.2\% five-axis mean).}
        \label{supp:tab:pool}
        \footnotesize
        \setlength{\tabcolsep}{3.0pt}
        \begin{adjustbox}{max width=\linewidth}
\begin{tabular}{@{}l c c c@{}}
            \toprule
            Milestone & Offline & \makecell{Semi-\\online} & \makecell{Full-\\online} \\
            \midrule
            \multicolumn{4}{@{}l}{\textit{RoboCasa365, Awomo-0.5I (50 tasks, 1,000 episodes)}} \\
            Single-VLM Argon (Figure~\ref{fig:runtime}a) & 650 & 657 & 752 \\
            Three-VLM pool + run-time tools & 657 & 651 & 768 \\
            Pool, multi-turn rescue (Table~\ref{tab:results}) & 656 & \best{773} & -- \\
            Pool, VLM-vetoed rescue triggers & 659 & 731 & -- \\
            \midrule
            \multicolumn{4}{@{}l}{\textit{RoboDojo, Awomo-0.5I (42 tasks, 840 episodes)}} \\
            Single-VLM Argon (Figure~\ref{fig:runtime}a) & 336 & 369 & 413 \\
            Three-VLM pool + run-time tools & 377 & 433 & 434 \\
            Pool, revised tools (Table~\ref{tab:results}) & 379 & 432 & \best{447} \\
            \midrule
            \multicolumn{4}{@{}l}{\textit{RoboDojo, WALL-OSS-1.0-lite (42 tasks, 840 episodes)}} \\
            Pool, 6-turn rescue stage & 479 & 503 & -- \\
            Pool, 24-turn rescue stage (Table~\ref{tab:results}) & 479 & \best{524} & -- \\
            \bottomrule
        \end{tabular}
        \end{adjustbox}
    \end{minipage}%
    \hfill
    \begin{minipage}[t]{0.48\textwidth}
        \centering
        \caption{\textbf{Zero-API local open-weight rescue on RoboCasa365 (Awomo-0.5I): selective gating adds +22.5 points, and compiled Offline policy code is optimal at zero cloud API cost (Figure~\ref{fig:validation}b).} Paired 95\% bootstrap CIs (1,140 local episodes, 0 cloud calls; 12-task slice: 8 fixed-base, 4 mobile-base).}
        \label{supp:tab:local_qwen}
        \footnotesize
        \setlength{\tabcolsep}{2.4pt}
        \begin{adjustbox}{max width=\linewidth}
\begin{tabular}{@{}l c c c@{}}
            \toprule
            Comparison (RoboCasa365) & Tasks & Success (\%) & $\Delta$ [95\% CI] \\
            \midrule
            \multicolumn{4}{@{}l}{\textit{Selective gating ablation (stratified 12-task slice)}} \\
            Gated vs.\ forced (Qwen3.6-35B) & 12 & \best{38.3}\,/\,15.8 & $+22.5$ $[+7.9,+39.2]$ \\
            \makecell[l]{Gated Qwen3.6-27B vs.\\Qwen3.6-35B} & 12 & \best{41.7}\,/\,38.3 & $+3.3$ $[-1.7,+8.3]$ \\
            \midrule
            \multicolumn{4}{@{}l}{\textit{Zero-API Offline vs.\ local and cloud Semi-online (all 50 tasks)}} \\
            \makecell[l]{Offline (0 calls) vs.\\Local Qwen3.6-35B} & 50 & \best{65.6}\,/\,57.3 & $+8.3$ $[+4.2,+12.9]$ \\
            \makecell[l]{Cloud Semi-online (selective)\\vs.\ Local Qwen3.6-35B} & 50 & \best{75.8}\,/\,57.3 & $+18.5$ $[+11.8,+26.0]$ \\
            \bottomrule
        \end{tabular}
        \end{adjustbox}
    \end{minipage}
\end{table}

\begin{table}[!htbp]
    \centering
    \caption{\textbf{Task-selective Semi-online rescue retains nearly the full gain of all-task rescue while skipping VLM calls on easy tasks.} Offline policy code runs on the training-selected subset and Semi-online multi-turn rescue runs on the remaining tasks. Paired differences (points) and 95\% bootstrap intervals are computed by task and episode.}
    \label{supp:tab:selective_rescue}
    \footnotesize
    \setlength{\tabcolsep}{4.5pt}
    \begin{adjustbox}{max width=\textwidth, max totalheight=0.86\textheight}
\begin{tabular}{@{}l l r c c c@{}}
        \toprule
        Counted result & Tasks counted Offline & Solved & Success & Row $-$ Offline & All-task $-$ row \\
        \midrule
        \multicolumn{6}{@{}l}{\emph{RoboCasa365, Awomo-0.5I (50 tasks, 1,000 test episodes)}} \\
        Offline & all 50 & 656 & 65.6 $[57.8,72.9]$ & & $+11.7$ $[+6.6,+17.5]$ \\
        Semi-online (selective) & 17 with Offline training SR $\geq 85\%$ & 758 & 75.8 $[69.4,81.7]$ & $+10.2$ $[+5.1,+16.0]$ & $+1.5$ $[-0.1,+3.6]$ \\
        & 18 with Offline training SR $\geq$ rescue SR & 758 & 75.8 $[68.8,82.2]$ & $+10.2$ $[+5.1,+16.0]$ & $+1.5$ $[0.0,+3.3]$ \\
        & 22 with converged training candidates & 739 & 73.9 $[67.3,80.1]$ & $+8.3$ $[+3.5,+13.9]$ & $+3.4$ $[+0.9,+6.4]$ \\
        Semi-online (all tasks) & none & 773 & 77.3 $[70.7,83.3]$ & $+11.7$ $[+6.6,+17.5]$ & \\
        \midrule
        \multicolumn{6}{@{}l}{\emph{RoboDojo, WALL-OSS-1.0-lite (42 tasks, 840 test episodes), pooled over episodes}} \\
        Offline & all 42 & 479 & 57.0 $[46.4,67.1]$ & & $+5.4$ $[+1.2,+9.9]$ \\
        Semi-online (selective) & 18 with low-intervention rescue in training & 517 & 61.5 $[51.2,71.4]$ & $+4.5$ $[+0.6,+8.9]$ & $+0.8$ $[-0.6,+2.5]$ \\
        & 11 with converged training candidates & 519 & 61.8 $[51.7,71.4]$ & $+4.8$ $[+0.8,+9.2]$ & $+0.6$ $[-0.7,+2.0]$ \\
        Semi-online (all tasks) & none & 524 & 62.4 $[52.0,72.4]$ & $+5.4$ $[+1.2,+9.9]$ & \\
        \midrule
        \multicolumn{6}{@{}l}{\emph{RoboDojo, WALL-OSS-1.0-lite, mean of the five skill axes}} \\
        Offline & all 42 & & 57.0 $[46.0,67.5]$ & & $+5.2$ $[+1.2,+9.3]$ \\
        Semi-online (selective) & 18 with low-intervention rescue in training & & 61.5 $[50.5,71.6]$ & $+4.5$ $[+0.8,+8.4]$ & $+0.7$ $[-0.8,+2.3]$ \\
        & 11 with converged training candidates & & 61.5 $[50.9,71.6]$ & $+4.5$ $[+0.8,+8.3]$ & $+0.7$ $[-0.8,+2.3]$ \\
        Semi-online (all tasks) & none & & 62.2 $[51.4,72.5]$ & $+5.2$ $[+1.2,+9.3]$ & \\
        \bottomrule
    \end{tabular}
\end{adjustbox}
\end{table}

\begin{table}[!htbp]
    \centering
    \caption{\textbf{Frozen X-Harness policy code reproduces large, statistically confirmed gains on brand-new unseen test seeds and layouts (Figure~\ref{fig:validation}a).} Paired evaluation against VLA alone on identical unseen episodes (2,080 fresh episodes per arm) with 95\% two-level paired bootstrap CIs. Runtime rescue calls on RoboCasa365 and RoboDojo are served by Gemini~4 Argon and Claude Opus~5.5 (which also serves GPT-6 Astra-authored candidates).}
    \label{supp:tab:fresh_seeds}
    \footnotesize
    \setlength{\tabcolsep}{4.0pt}
    \begin{adjustbox}{max width=\textwidth, max totalheight=0.86\textheight}
\begin{tabular}{@{}l l l c c c@{}}
        \toprule
        Benchmark (fresh split) & Metric & Arm & Alone & X-Harness & $\Delta$ pp [95\% CI] \\
        \midrule
        LIBERO-PRO ($\pi_{0.5}$, states 40--49) & 24 cells (240 eps) & Offline (Opus) & 49.6 & \best{97.1} & $+47.5$ $[+36.3,+58.8]$ \\
        \midrule
        RoboDojo (WALL-OSS-1.0-lite, 20 new layouts) & 42 tasks, 5-axis macro & Offline & 29.5 & 57.8 & $+28.3$ $[+17.2,+40.1]$ \\
        & 42 tasks, 5-axis macro & Semi-online (all tasks) & 29.5 & \best{58.9} & $+29.4$ $[+17.4,+42.0]$ \\
        & 42 tasks, 5-axis macro & Semi-online (selective) & 29.5 & 57.9 & $+28.4$ $[+16.8,+40.7]$ \\
        \midrule
        RoboCasa365 (Awomo-0.5I, seeds $\geq 2000$) & 50 tasks (1,000 eps) & Offline & 64.2 & 65.8 & $+1.6$ $[-1.3,+4.6]$ \\
        & 50 tasks (1,000 eps) & Semi-online (all tasks) & 64.2 & \best{72.6} & $+8.4$ $[+3.4,+13.8]$ \\
        & 50 tasks (1,000 eps) & Semi-online (selective) & 64.2 & 71.8 & $+7.6$ $[+2.7,+13.0]$ \\
        \bottomrule
    \end{tabular}
\end{adjustbox}
\end{table}

\begin{figure}[p]
    \centering
    \includegraphics[width=\textwidth,height=0.82\textheight,keepaspectratio]{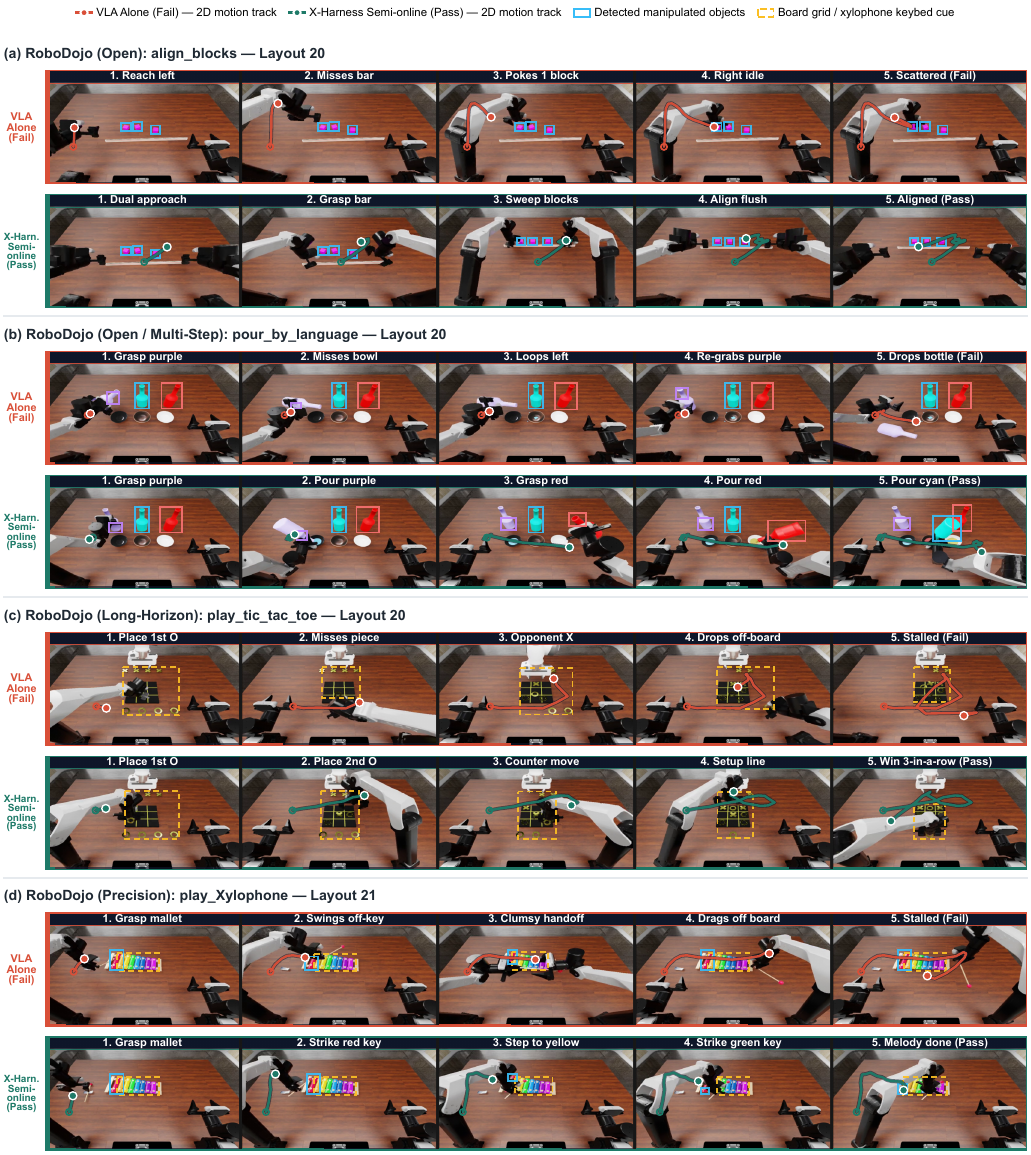}
    \caption{\textbf{Paired qualitative rollouts on Pure-RGB RoboDojo (WALL-OSS-1.0-lite): VLA Alone (Fail) vs.\ X-Harness Semi-online (Pass) at five key action milestones.} Identical unseen test layouts overlaid with cumulative 2D motion tracks (coral: VLA alone; emerald: X-Harness Semi-online), detected object/target bounding boxes, and episode-progress bars. \textbf{(a)}~\emph{align\_blocks} (Open): dual-arm bar grasp aligns all blocks flush, whereas VLA alone pokes one block with an idle right arm. \textbf{(b)}~\emph{pour\_by\_language} (Open / Multi-Step): sequential purple$\to$red$\to$cyan pouring vs.\ looping and dropping the first bottle. \textbf{(c)}~\emph{play\_tic\_tac\_toe} (Long-Horizon): multi-turn board tracking achieves a three-in-a-row win vs.\ off-board drops. \textbf{(d)}~\emph{play\_Xylophone} (Precision): grasps the mallet and steps accurately across the target keys to finish the melody vs.\ clumsy handoff and dragging off-board.}
    \label{supp:fig:qualitative_rd}
\end{figure}

\begin{figure}[p]
    \centering
    \includegraphics[width=\textwidth,height=0.82\textheight,keepaspectratio]{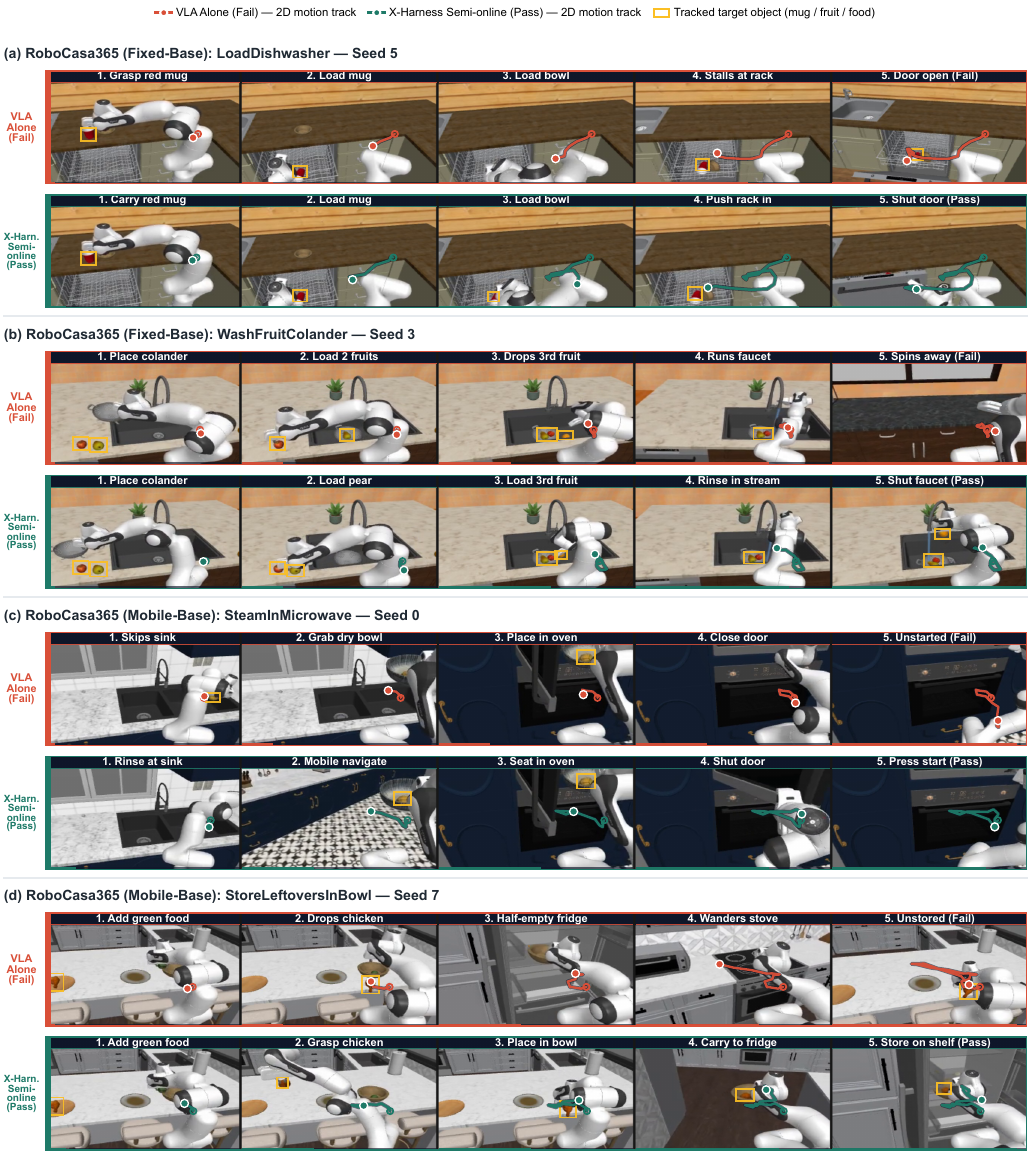}
    \caption{\textbf{Paired qualitative rollouts on RoboCasa365 (Awomo-0.5I): VLA Alone (Fail) vs.\ X-Harness Semi-online (Pass) at five key action milestones.} Identical test seeds overlaid with cumulative 2D motion tracks (coral: VLA alone; emerald: X-Harness Semi-online), detected target-object bounding boxes, and episode-progress bars. \textbf{(a)}~\emph{LoadDishwasher} (Fixed-Base): loads mug and bowl, pushes the rack in, and shuts the door flush vs.\ stalling over the open rack. \textbf{(b)}~\emph{WashFruitColander} (Fixed-Base): loads all three fruits into the colander, rinses under the faucet stream, and shuts the water off vs.\ dropping the third fruit and spinning away. \textbf{(c)}~\emph{SteamInMicrowave} (Mobile-Base): wets the food at the sink, navigates across the kitchen, seats the bowl in the microwave, closes the door, and presses start vs.\ skipping the sink and stalling unstarted. \textbf{(d)}~\emph{StoreLeftoversInBowl} (Mobile-Base): places both food items into the bowl, carries the full bowl to the refrigerator, and stores it cleanly on the shelf vs.\ dropping the chicken drumstick and wandering around the kitchen.}
    \label{supp:fig:qualitative_rc}
\end{figure}

\lead{Paired qualitative rollout comparisons.} Figures~\ref{supp:fig:qualitative_rd} and~\ref{supp:fig:qualitative_rc} compare VLA alone against X-Harness Semi-online on identical unseen test layouts and seeds across Pure-RGB RoboDojo (WALL-OSS-1.0-lite) and RoboCasa365 (Awomo-0.5I). Across coordinated dual-arm manipulation, multi-step state tracking, precise tool use, and mobile-base kitchen navigation, X-Harness Semi-online converts stalled or drifting base-VLA episodes into clean end-to-end completions.

\lead{Mid-episode object disturbances.} Shifting or dropping target objects mid-episode across 40 tasks of four benchmarks (746 cells) confirms strong closed-loop robustness. Offline evolved code reaches 57.3\% undisturbed, 52.5\% after a shift, and 45.6\% after a drop, substantially outperforming the base VLA (30.1\%, 28.4\%, and 20.9\%).

\FloatBarrier
\begin{wraptable}{r}{0.45\textwidth}
    \vspace{-10pt}
    \centering
    \caption{\textbf{Under domain randomization, evolved policy code transfers cleanly and beats the base VLA across all 50 tasks.} Success (\%) on RoboTwin~2.0 for code evolved only in the clean setting, pooled over three VLMs (Offline unless a guard is noted).}
    \label{supp:tab:rand}
    \footnotesize
    \renewcommand{\arraystretch}{0.94}
    \setlength{\tabcolsep}{3.0pt}
    \begin{adjustbox}{max width=\linewidth}
\begin{tabular}{@{}l c c@{}}
        \toprule
        Arm & Randomized & Clean \\
        \midrule
        \multicolumn{3}{@{}l}{\textit{(a) 50 tasks}} \\
        VLA alone ($\pi_{0.5}$) & 75.2 & 76.6 \\
        Evolved policy code & \best{78.4} & \best{80.9} \\
        \quad Flash & 75.9 & 78.2 \\
        \quad Sonnet & 78.4 & 81.2 \\
        \quad Opus & 80.9 & 83.3 \\
        \midrule
        \multicolumn{3}{@{}l}{\textit{(b) 8 tasks, before evolution}} \\
        Initial policy code & 35.6 & 42.7 \\
        Initial policy code (re-evaluation repeat) & 35.0 & 41.0 \\
        One-shot evolution & \best{46.0} & \best{57.1} \\
        Coroutine template & 41.5 & 50.8 \\
        Rules template & 25.6 & 32.7 \\
        ReKep template & 34.0 & 35.6 \\
        \midrule
        \multicolumn{3}{@{}l}{\textit{(c) 8 tasks, after eight evolution rounds}} \\
        Evolved policy code (selection gate) & 59.8 & \best{81.7} \\
        \quad + guard that may take over & \best{72.7} & 79.4 \\
        Without simulator diagnostics & 58.3 & 79.0 \\
        Without VLA calls & 45.2 & 74.4 \\
        Gate with no newly failed episode & 53.5 & 72.1 \\
        Bayesian gate & 54.4 & 78.1 \\
        Gate without merging rejected fixes & 62.5 & 78.8 \\
        \bottomrule
    \end{tabular}
    \end{adjustbox}
    \vspace{-6pt}
\end{wraptable}

\lead{RoboTwin~2.0 under domain randomization.} Deploying policy code evolved solely in clean RoboTwin scenes unchanged under domain randomization (cluttered tables, random backgrounds, lighting, and table heights; Table~\ref{supp:tab:rand}) demonstrates strong visual transfer. Across all 50 tasks, Offline evolved code retains 78.4\% (vs.\ 80.9\% clean) and stays $+3.2$ points above $\pi_{0.5}$ alone, with Opus reaching 80.9\%. On the 8-task ablation, eight evolution rounds still lift initial code by $+24.2$ points under randomization, and adding a lightweight guard ($\le 2$ VLM checks per episode) raises randomized success further from 59.8\% to 72.7\% (90.0\% with Opus).

\lead{Cross-task strategy rules and cold-start inputs.} Distilling $\sim$10 short cross-task strategy rules from other tasks' evolution lifts continued evolution (rounds 9--12 across 89 non-saturated pairs) by $+3.0$ points (43.5\% vs.\ 40.4\%, 95\% CI $[+0.7,+6.1]$; $+5.7$ for Flash). Isolating the two cold-start additions of the later harness version (\S\ref{supp:protocol}) confirms complementary gains. Cross-task \emph{skill cards} lift zero-rollout round-0 code by $+7.3$ points (43.5\%$\to$50.8\% over 143 cells), while initializing from the \emph{VLA-alone wrapper} when stronger boosts starting candidates by $+18.4$ points (42.6\%$\to$61.0\% over 120 cells) and preserves a $+2.1$-point edge after eight rounds (75.9\% vs.\ 73.8\%).

\lead{Evolution without simulator diagnostics.} Removing simulator diagnostics from failure reports and progress tie-breaking from the selection gate tests 144 cells across four benchmarks (2,880 episodes per arm; Figure~\ref{supp:fig:ablations}b). Using only action logs and binary success, eight evolution rounds still reach 70.8\% overall (within 5.6 points of full diagnostics), with RoboTwin unchanged at 74.4\% and Opus retaining 79.6\% (vs.\ 82.7\%), showing that text-only failure logs drive most evolution gains.

\lead{Complementary role of VLA calls.} Ablating VLA calls during evolution highlights their complementary strength under visual and contact complexity. While strong coders such as Opus can solve clean RoboTwin scenes via geometric primitives, retaining VLA calls boosts clean RoboTwin success by $+7.3$ points ($+28.7$ for Flash), adds $+14.6$ points under domain randomization (Table~\ref{supp:tab:rand}), and adds $+9.8$ points on RoboDojo (34.2\% vs.\ 24.4\% across 10 tasks).

\lead{Evolving a tool bank instead of the policy code.} Rather than revising policy code directly, we can hold the starting policy code frozen and evolve a four-hook shared tool bank around it (detection refinement, action expansion, per-step hook, and episode initialization). Across 166 non-saturated cells of six benchmarks, four rounds of tool-bank evolution match eight rounds of full policy-code evolution (63.5\% vs.\ 63.4\%, 95\% CI $[-2.4,+2.5]$; Figure~\ref{supp:fig:ablations}c), including a $+4.0$-point gain on RoboDojo ($[+0.7,+7.8]$).

\FloatBarrier
\Needspace*{14\baselineskip}
\section{Leaderboard References}
\label{supp:leader}

\begin{wraptable}{r}{0.55\textwidth}
    \vspace{-10pt}
    \centering
    \caption{\textbf{Wherever Table~\ref{tab:results} reports both scores, our Rerun meets or exceeds the official leaderboard score.} Official leaderboard and Rerun VLA-alone success (\%; --: not rerun). Top: standard interface used in Table~\ref{tab:results} and Figure~\ref{fig:where}. Bottom: expanded full-body RoboCasa365 interface (all 50 tasks, 1,000 episodes per VLA) used in Figure~\ref{fig:improve}c.}
    \label{supp:tab:leader}
    \footnotesize
    \setlength{\tabcolsep}{2.4pt}
    \begin{adjustbox}{max width=\linewidth}
\begin{tabular}{@{}l l c c c l@{}}
        \toprule
        & & & \multicolumn{2}{c}{Rerun} & \\
        \cmidrule(lr){4-5}
        Benchmark & VLA & Official & All & Fixed / mobile & Board date \\
        \midrule
        RoboMME & MME-VLA & 44.5 & 50.0 & & 2026-10-30 \\
        RoboTwin 2.0 & $\pi_{0.5}$ & 70.7 & 76.6 & & 2026-10-30 \\
        RoboCasa365 & $\pi_{0.5}$ & 16.9 & 23.1 & 29.1 / 6.2 & 2026-11-01 \\
        RoboCasa365 & RLDX-1 & 36.0 & 31.8 & 36.2 / 19.2 & 2026-11-01 \\
        RoboCasa365 & XR-1 & 57.4 & 51.1 & 54.2 / 42.3 & 2026-11-01 \\
        RoboDojo & $\pi_{0.5}$ & 6.9 & 9.6 & & 2026-11-03 \\
        RoboDojo & DM0.5 & 19.3 & 19.6 & & 2026-11-03 \\
        RoboDojo & WALL-OSS-1.0-lite & 35.0 & -- & & 2026-11-11 \\
        \midrule
        \multicolumn{6}{@{}l}{\textit{RoboCasa365, expanded full-body interface (all 50 tasks)}} \\
        RoboCasa365 & $\pi_{0.5}$ & 16.9 & 26.7 & 32.7 / 9.6 & 2026-11-01 \\
        RoboCasa365 & RLDX-1 & 36.0 & 35.0 & 40.1 / 20.4 & 2026-11-01 \\
        RoboCasa365 & XR-1 & 57.4 & 56.2 & 59.7 / 46.2 & 2026-11-01 \\
        \bottomrule
    \end{tabular}
    \end{adjustbox}
    \vspace{-6pt}
\end{wraptable}

Table~\ref{supp:tab:leader} compares each base VLA's published leaderboard score (\emph{Official}) with our evaluation under our protocol (\emph{Rerun}). Across every row of Table~\ref{tab:results} where both are reported, our Rerun meets or exceeds the Official score, ensuring all X-Harness gains are measured against a strong baseline. On RoboDojo, policies are evolved on dedicated training layouts (seeds $\geq 100$) and evaluated frozen on the 20 public test layouts (seed~0), with official leaderboard scores shown as reference.

\lead{Notes to Table~\ref{tab:results}.} On RoboMME, Official and Rerun evaluate the strongest memory MME-VLA checkpoint (row~1 of Table~\ref{supp:tab:vla}), while the harness calls the target-marked sub-goal variant (row~3). On RoboTwin~2.0, Official uses the clean leaderboard reference, and on Awomo-0.5I and WALL-OSS-1.0-lite we load the official weights and report matching official leaderboard scores (pooled WALL-OSS-1.0-lite success is 57.0\% Offline and 62.4\% Semi-online, Table~\ref{supp:tab:pool}).

\lead{Full-body interface verification on RoboCasa365.} Across all 50 RoboCasa365 tasks (1,000 episodes per VLA) under an expanded full-body interface supporting arm, base, and torso control, $\pi_{0.5}$, RLDX-1, and XR-1 match or approach their official leaderboard scores at a 39.3\% mean (Table~\ref{supp:tab:leader}, bottom). Evolved Opus Offline policy code maintains consistent gains across all three models, averaging 47.4\% and reaching 48.5\% with candidate selection (Figure~\ref{fig:improve}c).

\clearpage
\section{Example Policy Code}
\label{supp:program}

Figure~\ref{supp:fig:program} shows evolved Offline policy code (0 test-time VLM calls) written by Flash for RoboMME \emph{InsertPeg} (\emph{Insert} in Figure~\ref{fig:context}b) after eight rounds, where the robot must match the peg, grasp end, and insertion side shown in a demo video. Written as a compact free-form Python state machine (\texttt{PICK} $\to$ \texttt{INSERT}), it queries a 3D scene-graph tool to emit target-marked VLA sub-goals and checks gripper state after each step to retry grasps in a closed loop.

\begin{figure}[!htbp]
    \centering
    \begin{tcolorbox}[
        enhanced,
        colback=findingbg,
        colframe=cambrianframe!75,
        boxrule=0.6pt,
        arc=2.5pt,
        left=8pt,
        right=8pt,
        top=2pt,
        bottom=2pt,
        title={\small\textbf{Example Evolved Offline Policy Code} \hfill \normalfont\scriptsize RoboMME \emph{InsertPeg} (Flash, round 8, 0 test-time VLM calls)},
        coltitle=white,
        colbacktitle=cambrianframe,
        fonttitle=\sffamily
    ]
\begin{lstlisting}[style=xhpython,basicstyle=\ttfamily\scriptsize\linespread{0.96}\selectfont]
class Program:
  def set_task(self, instruction, obs, vision):
    front_cam = obs["cameras"]["front"]
    # 1. Detect box and pegs via 3D perception tool (Grounding DINO + SAM 2 + depth)
    sg = vision.get_scene_graph(
        front_cam["rgb"], front_cam["depth"], front_cam["K"], front_cam["T_w2c"],
        ["box", "wooden box", "peg", "cylinder", "block"], threshold=0.15)
    # 2. Read demo video to identify target peg, grasp end, and insertion side ...
    self.subgoal_pick = f"pick up the peg by grasping the {grasp_end} end at <{pick_r}, {pick_c}>"
    self.subgoal_insert = f"insert the peg from the {insert_side} side at <{box_r}, {box_c}>"
    self.phase = "PICK"

  def get_action(self):
    if self.phase == "PICK":
      self.pick_actions += 1
      return {"control": "policy", "steps": 110, "instruction": self.subgoal_pick}
    elif self.phase == "INSERT":
      return {"control": "policy", "steps": 180, "instruction": self.subgoal_insert}

  def update_step(self, outcome, obs, vision):
    tcp = obs.get("tcp", {})
    opening = tcp.get("gripper_opening_m", 0.08)
    if tcp.get("holding", False) or (0.005 <= opening <= 0.05):
      self.held_confirmed = True
    if self.phase == "PICK":
      if self.held_confirmed:        # grasp confirmed -> advance phase
        self.phase = "INSERT"
      elif self.pick_actions < 2:    # retry grasp once before advancing
        pass
      else:
        self.phase = "INSERT"
\end{lstlisting}
    \end{tcolorbox}
    \vspace{-4pt}
    \caption{\textbf{Example closed-loop policy code written by Flash for RoboMME \emph{InsertPeg} after eight evolution rounds.} Shortened from 261 lines (``\texttt{...}'' marks omitted helper logic). Even without a rigid template, the VLM emits a clean finite-state machine that pairs 3D perception with target-marked VLA sub-goals and closes the loop via live gripper feedback.}
    \label{supp:fig:program}
\end{figure}

\clearpage
\section{Real-Robot Validation on a Bimanual YAM Platform}
\label{supp:real_robot}

Because X-Harness decouples high-level policy code from the base VLA, evolved Offline code transfers directly to physical hardware without modifying VLA weights or native safety controllers. As a supplementary physical transfer check beyond our eight simulation benchmarks, we validate this pipeline on a bimanual YAM robot equipped with three uncalibrated RGB cameras and 14-DoF joint encoders---requiring no depth sensor, camera calibration, or inverse kinematics---paired with frozen task-adapted $\pi_{0.5}$ servers driven solely by language sub-goals across two memory- and state-dependent tasks: \emph{Two-Bin Object Memory} (70 training / 30 held-out test episodes) and \emph{LED-Counted Bean Scooping} (89 training / 44 held-out test episodes).

\begin{figure}[!htbp]
    \centering
    \includegraphics[width=\textwidth]{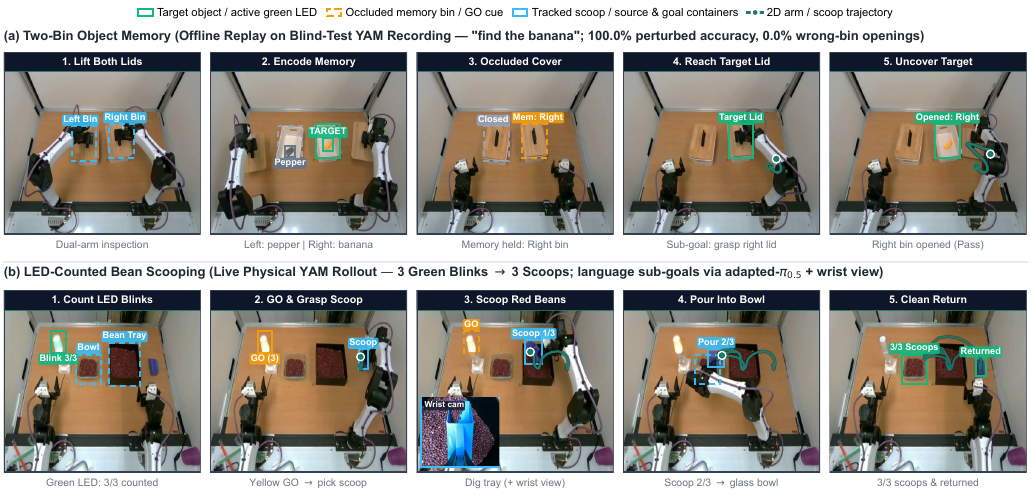}
    \caption{\textbf{Real-robot visual grounding, memory tracking, and sub-goal execution on a bimanual YAM platform.} \textbf{(a)}~Offline replay on a blind-test Two-Bin Object Memory recording (\emph{``find the banana''}): policy code lifts both lids, remembers the occluded target bin, and uncovers the banana ($100.0\%$ perturbed accuracy under RGB + DINO/SAM, $0.0\%$ wrong-bin openings; Table~\ref{supp:tab:real_yam}). \textbf{(b)}~Live physical LED-Counted Bean Scooping rollout (3 green blinks $\to$ 3 scoops) driving the unmodified adapted-$\pi_{0.5}$ server via language sub-goals across all three scoop-and-pour cycles.}
    \label{supp:fig:real_yam}
\end{figure}

Evolved Offline policy code (Gemini~4 Argon, 0 test-time VLM calls, $<$1\,ms decision latency) generalizes cleanly across spatial, photometric, wording, and timing perturbations as well as live hardware deployment (Figure~\ref{supp:fig:real_yam}; full metrics in Table~\ref{supp:tab:real_yam}). Across both lightweight CPU RGB and Grounding~DINO + SAM~2 modes, it achieves $100.0\%$ clean accuracy, $99.3\%$--$100.0\%$ perturbed accuracy with $0.0\%$ wrong-bin openings, and $100.0\%$ safe refusal with zero VLA calls when required memory cues are absent. Against live unmodified $\pi_{0.5}$ WebSocket servers, all 226 generated $30{\times}14$ joint-target chunks pass validation at 72--76\,ms per chunk and cut action MAE by $74\%$--$84\%$ over a static-hold baseline within $11.8$\,GiB peak VRAM on a single RTX~4090 GPU, executing physical multi-scoop trials end to end (Figure~\ref{supp:fig:real_yam}b).

\begin{table}[!htbp]
    \centering
    \caption{\textbf{Real-robot validation on held-out bimanual YAM demonstrations and live adapted-$\pi_{0.5}$ server integration (Figure~\ref{supp:fig:real_yam}).} Evolved Offline policy code (Gemini~4 Argon, 0 VLM calls, $<$1\,ms latency) over three perturbation seeds on held-out recordings (90 Two-Bin replays/condition; 132 Bean Scooping replays/condition; 318 synthetic fault replays; values in \%). \emph{Safety / Faults}: uncovered-bin refusal (Two-Bin) and fault detection (Bean Scooping); \emph{No-cue ref.}: safe refusal with zero VLA calls when memory cues are removed. Bold marks strictly higher score per task.}
    \label{supp:tab:real_yam}
    \footnotesize
    \setlength{\tabcolsep}{5.5pt}
    \begin{adjustbox}{max width=\textwidth}
\begin{tabular}{@{}l l c c c c c@{}}
        \toprule
        Task (held-out split) & Vision & Clean & Perturbed & Color & Safety / Faults & No-cue ref.\ (0 VLA) \\
        \midrule
        Two-Bin Object Memory (30 eps) & CPU RGB & 100.0 & 99.3 & 100.0 & 100.0 & 100.0 \\
        & RGB + DINO/SAM & 100.0 & \best{100.0} & 100.0 & 100.0 & 100.0 \\
        \midrule
        LED-Counted Bean Scooping (44 eps) & CPU RGB & 100.0 & 99.7 & 99.2 & 96.9 & 100.0 \\
        & RGB + DINO/SAM & 100.0 & \best{100.0} & \best{100.0} & \best{100.0} & 100.0 \\
        \midrule
        \multicolumn{7}{@{}l}{\textit{Live adapted-$\pi_{0.5}$ WebSocket integration (10 held-out replays across all four task--vision settings)}} \\
        \multicolumn{7}{@{}l}{\scriptsize Valid $30{\times}14$ chunks: \textbf{226/226} (0 rejected) \;\textbar\; MAE vs.\ static-hold: \textbf{$-$74\% to $-$84\%} (ratio 0.16--0.26) \;\textbar\; Latency: \textbf{72--76\,ms/chunk} \;\textbar\; Peak VRAM: \textbf{11.8\,GiB}} \\
        \bottomrule
    \end{tabular}
\end{adjustbox}
\end{table}

\end{document}